\documentclass[letterpaper, 10 pt, conference]{ieeeconf}

\IEEEoverridecommandlockouts                             

\usepackage[dvips]{graphicx}
\usepackage{orcidlink}
\usepackage{multirow}
\usepackage{amsmath}
\usepackage{dsfont}
\usepackage{tabularx}
\usepackage{comment}

\usepackage{pstool}
\usepackage{mathtools}
\usepackage{color}
\usepackage{psfrag}
\usepackage{caption}

\usepackage{pdfpages}
\usepackage{soul}

\usepackage{pifont}
\usepackage{url}

\usepackage{graphicx}
\usepackage{pstricks}
\usepackage{verbatim}
\usepackage{psfrag}
\usepackage{pstricks}

\usepackage{graphicx}
\usepackage{longtable}
\usepackage{booktabs}
\usepackage{lscape}
\usepackage{array}
\usepackage{multirow}

\csname PSTplotLoaded\endcsname
\let\PSTplotLoaded 
\ifx\PSTricksLoaded \else\input pstricks.tex\fi
\ifx\PSTXKeyLoaded \else \input pst-xkey.tex\fi
\ifx\PSTtoolsloaded \else\input pst-tools.tex\fi
\ifx\PSTtoolsloaded \else\input pstricks-add.tex\fi
\ifx\PSTFPloaded \else   \input pst-fp.tex  \fi
\ifx\MultidoLoaded \else \input multido.tex \fi
\def\fileversion{1.92}
\def\filedate{2019/05/16}
\edef\TheAtCode{\the\catcode`\@}
\catcode`\@=11
\pst@addfams{pst-plot}
\def\pst@linetype{2}%
\SpecialCoor

\newdimen\pstRadUnit
\newdimen\pstRadUnitInv
\begingroup
\catcode`\{=13
\catcode`\}=13
\catcode`\(=13
\catcode`\)=13
\catcode`\,=13
\catcode`\!=1
\catcode`\*=2
\catcode`\ =13
\catcode`\_=13
\catcode`\^^M=13
\gdef\pst@datadelimiters!
\catcode`\{=13%
\catcode`\}=13%
\catcode`\(=13%
\catcode`\)=13%
\catcode`\,=13%
\catcode`\ =13%
\catcode`\^^M=13%
\def,##1!%
\ifcat\noexpand,\noexpand##1%
\expandafter##1%
\else\space%
D\space##1%
\fi*%
\let(,\let),\let{,\let},\let ,\let^^M,\let_\@empty*
\endgroup
\define@key[psset]{pst-plot}{ignoreLines}[0]{\def\psk@ignoreLines{#1}}
\psset[pst-plot]{ignoreLines=0}
\newcount\pst@linecnt
\begingroup
\catcode`\,=13
\catcode`\_=13
\gdef\savedata@#1[#2]{%
  \xdef\pst@tempg{#2_}%
  \endgroup
  \let#1\pst@tempg
  \global\let\pst@tempg\relax
  \ignorespaces}
\gdef\readdata@{%
  \read1 to \pst@tempA
  \ifnum\pst@linecnt=\psk@nStep
    \global\pst@linecnt=0
    \expandafter\readdata@@\pst@tempA_\@nil
  \fi
  \global\advance\pst@linecnt by 1
  \ifeof1\else\expandafter\readdata@\fi}
\gdef\pst@@readfile#1#2\@nil{\addto@pscode{,#1#2}}%
\gdef\readdata@@#1#2\@nil{\xdef\pst@tempg{\pst@tempg,#1#2}}%
\endgroup
\def\readdata{\@ifnextchar[{\readdata@i}{\readdata@i[]}}
\def\readdata@i[#1]#2#3{%
  \openin1=#3
  \begingroup
  \ifx#1\relax#1\else\psset{#1}\fi
  \def\pst@tempg{}%
  \ifeof1
    \@pstrickserr{Data file `#3' not found.}\@ehpa
  \else
    \pst@datadelimiters
    \catcode`\[=1
    \catcode`\]=2
    \pst@cnta=0
    \loop \ifnum\the\pst@cnta<\psk@ignoreLines
      \advance\pst@cnta by 1\relax
      \read1 to \pst@tempA
    \repeat
    \psDEBUG[pst-plot]{>>> ignored \the\pst@cnta\space data lines}%
    \global\pst@linecnt=\psk@nStep
    \readdata@
  \fi
  \endgroup
  \global\let#2\pst@tempg%
  \global\let\pst@tempg\relax%
\ignorespaces}
\def\pst@readfile#1{{\let\readdata@@\pst@@readfile\readdata\pst@tempg{#1}}}
\def\pst@altreadfile#1{%
  \openin1=#1
  \ifeof1
    \@pstrickserr{Data file `#1' not found.}\@ehpa
  \else
    \catcode`\{=10
    \catcode`\}=10
    \catcode`\(=10
    \catcode`\)=10
    \catcode`\,=10
    \catcode`\^^M=10
    \catcode`\[=1
    \catcode`\]=2
    \pst@@altreadfile
  \fi}
\def\pst@@altreadfile{%
  \read1 to \pst@tempg
  \expandafter\pst@@@altreadfile\pst@tempg\@empty\@nil
  \ifeof1\else\expandafter\pst@@@altreadfile\fi}
\def\pst@@@altreadfile#1#2\@nil{\addto@pscode{#1#2}}%
\def\savedata#1{\begingroup\pst@datadelimiters\savedata@{#1}}
\newread\RCD@file
\def\psreadDataColumn{\@ifnextchar[\psreadDataColumn@i{\psreadDataColumn@i[]}}
\def\psreadDataColumn@i[#1]{%
  \psset{#1}%
  \psreadDataColumn@ii
}
\def\psreadDataColumn@ii#1#2#3#4{
  \immediate\openin\RCD@file=#4\relax
  \global\let#3=\@empty
  \pst@cnta=0
  \loop \ifnum\the\pst@cnta<\psk@ignoreLines
      \advance\pst@cnta by 1\relax
      \read\RCD@file to \@tempa
  \repeat
  \loop
    \read\RCD@file to \@tempa
    \ifeof\RCD@file\else
      \edef\@tempa{\@tempa#2}%
      \def\reserved@b{}%
      \@tempswafalse
      \@tempcnta=#1\relax
    \expandafter\@tfor\expandafter\reserved@a
      \expandafter:\expandafter=\@tempa\do{
      \if\reserved@a#2\relax
        \advance\@tempcnta \m@ne
        \ifnum \@tempcnta=\z@
          \expandafter\g@addto@macro\expandafter#3\expandafter{\reserved@b\space}%
          \@tempswatrue
        \fi
        \def\reserved@b{}
      \else
        \edef\reserved@b{\reserved@b\reserved@a}
      \fi
      \if@tempswa\@break@tfor\fi
    }%
  \repeat
  \immediate\closein\RCD@file
}

\def\beginplot@line{\begin@OpenObj}
\def\endplot@line{\psline@ii}
\def\beginplot@polygon{\begin@ClosedObj}
\def\endplot@polygon{\pspolygon@ii}
\def\beginplot@curve{\begin@OpenObj}
\def\endplot@curve{\pscurve@ii}
\def\beginplot@ecurve{\begin@OpenObj}
\def\endplot@ecurve{\psecurve@ii}
\def\beginplot@ccurve{\begin@ClosedObj}
\def\endplot@ccurve{\psccurve@ii}
\def\beginplot@dots{\begin@SpecialObj}
\def\endplot@dots{\psdots@ii}
\define@key[psset]{pst-plot}{Hue}[180]{\pst@getint{#1}\pst@HueAngle}
\psset[pst-plot]{Hue=180}
\def\beginplot@colordots{\begin@SpecialObj}
\def\endplot@colordots{%
  \addto@pscode{%
    \psk@dotsize
    \@nameuse{psds@\psk@dotstyle}
    newpath
    /MaxValue 0 def
    /m n 2 mul def
    n { 
      dup MaxValue gt { dup /MaxValue ED } if
      m 2 roll
    } repeat
    n { dup MaxValue div  
      \pst@number\psyunit div abs 
      \pst@HueAngle\space 360 div exch dup sethsbcolor 
      transform floor .5 add exch floor
      .5 add exch itransform Dot stroke } repeat }%
  \end@SpecialObj%
}
\def\beginplot@bubble{\begin@SpecialObj}
\def\endplot@bubble{%
  \addto@pscode{%
    newpath
    n { dup 
      \pst@number\psyunit div abs 
      transform floor .5 add exch floor
      .5 add exch itransform 
      0 360 arc \psk@fill 
      stroke } repeat }%
  \end@SpecialObj%
}
\def\beginplot@bezier{\begin@OpenObj}
\def\endplot@bezier{\psbezier@ii}
\def\beginplot@cbezier{\begin@ClosedObj}
\def\endplot@cbezier{\pscbezier@ii}
\def\beginplot@cspline{\begin@OpenObj}
\def\endplot@cspline{\pscspline@ii}
\let\beginplot@LineToYAxis\beginplot@line  
\def\endplot@LineToYAxis{\psLineToYAxis@ii}
\let\beginqp@LineToYAxis\beginqp@line
\let\doqp@LineToYAxis\doqp@line
\let\endqp@LineToYAxis\endqp@line
\let\testqp@LineToYAxis\testqp@line
\let\beginplot@LineToXAxis\beginplot@line
\def\endplot@LineToXAxis{\psLineToXAxis@ii}
\let\beginqp@LineToXAxis\beginqp@line
\let\doqp@LineToXAxis\doqp@line
\let\endqp@LineToXAxis\endqp@line
\let\testqp@LineToXAxis\testqp@line
\newif\ifPst@interrupt \Pst@interruptfalse
\define@key[psset]{pst-plot}{barwidth}[0.25cm]{\pst@getlength{#1}\Add@barwidth}
\psset[pst-plot]{barwidth=0.25cm}
\define@key[psset]{pst-plot}{interrupt}[]{\expandafter\pst@interrupt#1,,,\@nil}
\def\pst@interrupt#1,#2,#3,#4\@nil{%
  \ifx\relax#1\relax \Pst@interruptfalse
  \else
    \Pst@interrupttrue
    \def\pst@interrupt@YMax{#1 }%
    \def\pst@interrupt@YMaxSep{#2 }%
    \def\pst@interrupt@YMaxDiff{#3 }%
  \fi
}
\def\psbar@ii{\addto@pscode{false \tx@NArray \psbar@iii}}

\def\psbar@iii{%
  \ifPst@interrupt
    /YMax \pst@interrupt@YMax \strip@pt\psyunit\space mul def
    /YMaxSep \pst@interrupt@YMaxSep \strip@pt\psyunit\space mul def
    /YMaxDiff \pst@interrupt@YMaxDiff \strip@pt\psyunit\space mul def
    /Tilde { 
      /Op ED 
      /DX ED
      currentpoint 2 copy
      /Y ED /X ED   
      X DX add Y YMaxSep 2 div Op   
      X DX dup add add Y           
      curveto
      currentpoint 2 copy pop /X ED 
      X DX add Y YMaxSep 2 div neg Op  
      X DX dup add add Y    
      curveto      
    } def  
    newpath
    n { 
      /Yval exch def /Xval exch def 
      Xval \number\Add@barwidth 0.5 mul sub 0 moveto 
      Yval YMax le {  
        0 Yval rlineto \number\Add@barwidth 0 rlineto 
        0 Yval neg rlineto \number\Add@barwidth neg 0 rlineto
      }{
        0 YMax rlineto 
        \number\Add@barwidth 4 div 
        { add } Tilde
        0 YMax neg rlineto 
        \number\Add@barwidth neg 0 rlineto
        closepath
        Xval \number\Add@barwidth 0.5 mul sub YMax YMaxSep add moveto 
        0 Yval YMax sub YMaxSep sub YMaxDiff sub rlineto 
        \number\Add@barwidth 0 rlineto 
        0 Yval YMax YMaxSep add sub YMaxDiff sub neg rlineto 
        \number\Add@barwidth 4 div neg 
        { sub } Tilde
      } ifelse
    } repeat
  \else
    newpath
    n { 
      /Yval exch def /Xval exch def 
      Xval \number\Add@barwidth 0.5 mul sub 0 moveto 
      0 Yval rlineto \number\Add@barwidth 0 rlineto 
      0 Yval neg rlineto \number\Add@barwidth neg 0 rlineto
    } repeat
  \fi
}%
\def\beginplot@bar{\begin@SpecialObj}
\def\endplot@bar{%
  \psbar@ii\psk@fillstyle\ifpsshadow\pst@closedshadow\fi%
  \pst@stroke
  \end@SpecialObj}
\def\psybar@ii{\addto@pscode{false \tx@NArray \psybar@iii}}
\def\psybar@iii{%
  newpath
  n { 
    /Yval exch def /Xval exch def 
    0 Yval \number\Add@barwidth 0.5 mul sub moveto 
    Xval 0 rlineto 0 \number\Add@barwidth rlineto 
    Xval neg 0 rlineto 0 \number\Add@barwidth neg rlineto
  } repeat
}%
\def\beginplot@ybar{\begin@SpecialObj}
\def\endplot@ybar{%
  \psybar@ii\psk@fillstyle\ifpsshadow\pst@closedshadow\fi%
  \pst@stroke
  \end@SpecialObj}
\def\psLSM@ii{\addto@pscode{ false \tx@NArray \psLSM@iii }}
\def\psLSM@iii{%
  /xiSquare 0 def				
  /xi 0 def					
  /fi 0 def					
  /xifi 0 def					
  exch dup dup /xEnd ED /xStart ED exch
  n { 						
    /Yval ED /Xval ED 				
    /xi xi Xval add def				
    /xiSquare xiSquare Xval dup mul add def	
    /xifi xifi Xval Yval mul add def		
    /fi fi Yval add def				
    Xval xStart lt { /xStart Xval def } if	
    Xval xEnd gt { /xEnd Xval def } if		
  } repeat
  /u xiSquare fi mul xi xifi mul sub n xiSquare mul xi dup mul sub div def
  /v n xifi mul xi fi mul sub n xiSquare mul xi dup mul sub div def
  \Pst@Debug\space 0 gt { 			
    /NimbusSanL-Regu findfont 12 scalefont setfont	
    0 -50 moveto (y=) show 			
    v \pst@number\psyunit \pst@number\psxunit div div 20 string cvs show ( x+) show		
    u \pst@number\psyunit div 20 string cvs show } if
  newpath
  (\psk@xStart) length 0 gt 			
    { \psk@xStart\space \pst@number\psxunit mul }
    { xStart } ifelse 
  dup v mul u add 				
  moveto		 			
  (\psk@xEnd) length 0 gt 			
    { \psk@xEnd\space \pst@number\psxunit mul }
    { xEnd } ifelse 
  dup v mul u add 				
  lineto					
}%
\def\beginplot@LSM{\begin@SpecialObj}
\def\endplot@LSM{%
  \psLSM@ii\psk@fillstyle\ifpsshadow\pst@closedshadow\fi%
  \pst@stroke
  \end@SpecialObj}
\define@key[psset]{pst-plot}{IQLfactor}{\pst@checknum{#1}\pst@IQLfactor}
\psset[pst-plot]{IQLfactor=1.5}
\define@key[psset]{pst-plot}{postAction}[]{\def\psk@postAction{%
  \ifx\relax#1\relax\else\pst@number\psyunit div #1 \pst@number\psyunit mul \fi }}
\psset[pst-plot]{postAction=}
\define@key[psset]{pst-plot}{mediancolor}[black]{\pst@getcolor{#1}\median@linecolor}
\psset[pst-plot]{mediancolor=black}
\define@boolkey[psset]{pst-plot}[Pst@]{markMedian}[true]{}
\psset[pst-plot]{markMedian=false}

\def\psBoxplot@ii{%
  \addto@pscode{
    /Barwidth \number\Add@barwidth 2 div def  
    /Endwidth Barwidth \psk@arrowlength\space mul def  
   NArray bubblesort
   /NArray ED 				
   [ NArray { yUnit mul } forall ] /NArray ED 
   NArray 0 get /MinVal ED		
   NArray m 1 sub get /MaxVal ED	
   m 2 div cvi /M ED 			
   NArray length 2 mod 0 eq {		
     M 1 sub NArray exch get 		
     NArray M get          		
     add 2 div /Median ED  		
   }{
     NArray M get /Median ED  		
   } ifelse
   m 4 mod 0 eq {	  		
     m 4 div cvi dup 1 sub NArray exch get
     exch NArray exch get
     add 2 div floor /LowerQuartil ED
   }{ 
     NArray M 2 div cvi get /LowerQuartil ED 
   } ifelse				
   m 0.75 mul dup dup cvi sub 0 eq {	
     cvi dup 1 sub NArray exch get exch NArray exch get
     add 2 div floor /UpperQuartil ED
   }{					
     NArray m 0.75 mul floor cvi get /UpperQuartil ED
   } ifelse 
   /IQL UpperQuartil LowerQuartil sub \pst@IQLfactor\space mul def
   0 1 m 1 sub { 
     dup /Index ED
     NArray exch get LowerQuartil sub abs IQL sub 0 gt { 
       \psk@dotsize
       \@nameuse{psds@\psk@dotstyle}
       0 NArray Index get \psk@postAction
       Dot
       NArray Index LowerQuartil UpperQuartil LowerQuartil sub \pst@IQLfactor\space mul sub 
       dup /MinVal ED put 
       NArray Index 1 add get /MinVal ED 
    } { exit } ifelse
   } for
   m 1 sub -1 0 {	
     dup /Index ED
     NArray exch get UpperQuartil sub abs IQL sub 0 gt { 
       \psk@dotsize
       \@nameuse{psds@\psk@dotstyle}
       0 NArray Index get \psk@postAction\space
       Dot
       NArray Index UpperQuartil LowerQuartil sub \pst@IQLfactor\space mul UpperQuartil add 
       dup /MaxVal ED put 
       NArray Index 1 sub get /MaxVal ED 
     }{ exit } ifelse
   } for
   Endwidth neg MaxVal \psk@postAction moveto			
   Endwidth dup add 0 rlineto 
   0 MaxVal \psk@postAction moveto 
   0 UpperQuartil \psk@postAction lineto			
   MinVal \psk@postAction MaxVal \psk@postAction lt {
     0 LowerQuartil \psk@postAction moveto			
     0 MinVal \psk@postAction lineto 
     Endwidth neg MinVal \psk@postAction moveto 
     Endwidth dup add 0 rlineto 
   } if
   gsave
   \pst@number\pslinewidth SLW
   \pst@usecolor\pslinecolor
   \tx@setStrokeTransparency 
   \@nameuse{psls@\pslinestyle}
   stroke
   grestore
   newpath
   Barwidth neg LowerQuartil \psk@postAction moveto	
   Barwidth neg UpperQuartil \psk@postAction lineto
   Barwidth dup add 0 rlineto
   Barwidth LowerQuartil \psk@postAction lineto
   closepath
   \pst@usecolor\psfillcolor
   gsave \pst@usecolor\psfillcolor \tx@setTransparency fill grestore
   \@nameuse{psls@solid}
   \ifPst@markMedian
     \pst@number\pslabelsep neg Median moveto currentpoint 
     /YMedian ED /XMedian ED 
      Barwidth neg Median \psk@postAction lineto  
   \else
      Barwidth neg Median \psk@postAction moveto  
   \fi
   Barwidth dup add 0 rlineto 
   \pst@number\pslinewidth SLW
   \pst@usecolor\median@linecolor
   \tx@setStrokeTransparency
   stroke
  }
}%
\def\beginplot@Boxplot{\init@pscode}
\def\endplot@Boxplot{%
  \psBoxplot@ii\psk@fillstyle\ifpsshadow\pst@closedshadow\fi%
  \pst@stroke
  \end@SpecialObj}
\def\psBoxplot{\def\pst@par{}\pst@object{psBoxplot}}
\def\psBoxplot@i#1{%
  \leavevmode
  \pst@killglue
  \begingroup
  \addbefore@par{barwidth=40pt,arrowlength=0.75}%
  \addto@par{plotstyle=Boxplot}%
  \use@par
  \@nameuse{beginplot@\psplotstyle}%
  \addto@pscode{
    /D {} def
    [ #1 ] /NArray ED 
    NArray aload length /m ED
    /xUnit \pst@number\psxunit def
    /yUnit \pst@number\psyunit def
  }%
  \@nameuse{endplot@\psplotstyle}%
  \ignorespaces%
}
\define@key[psset]{pst-plot}{plotstyle}[line]{%
  \@ifundefined{beginplot@#1}%
    {\@pstrickserr{Plot style `#1' not defined}\@eha}%
    {\def\psplotstyle{#1}}}
\psset[pst-plot]{plotstyle=line}
\define@key[psset]{pst-plot}{plotpoints}[50]{%
  \pst@cntg=#1\relax
  \ifnum\pst@cntg<2
    \@pstrickserr{plotpoints parameter must be at least 2}\@ehpa
  \else
    \advance\pst@cntg-1
    \edef\psk@plotpoints{\the\pst@cntg\space}%
  \fi}
\psset[pst-plot]{plotpoints=50}
\define@key[psset]{pst-plot}{yMaxValue}[1.e30]{\def\psk@yMaxValue{#1 }\def\psk@yMinValue{#1 neg }}
\psset{yMaxValue=1.e30}
\define@key[psset]{pst-plot}{yMinValue}[-1.e30]{\def\psk@yMinValue{#1 }}
\psset{yMinValue=-1.e30}
\def\beginqp@line{\pst@oplineto}
\def\doqp@line{ 
  dup
  \psk@yMaxValue \pst@number\psyunit mul gt 
    { moveto }
    { dup \psk@yMinValue \pst@number\psyunit mul lt 
      { moveto }
      { L } ifelse 
    } ifelse
}
\def\endqp@line{%
  \ifPst@variableLW \addto@pscode{ \pst@flattenpath }\fi%
  \end@OpenObj}%

\def\testqp@line{%
  \ifdim\pslinearc>\z@\else
    \ifshowpoints\else
      \ifx\psk@arrowA\@empty
        \ifx\psk@arrowB\@empty
          \@psttrue
        \fi
      \fi
    \fi
  \fi}
\def\beginqp@polygon{moveto }
\def\doqp@polygon{ 
      dup
      \psk@yMaxValue \pst@number\psyunit mul gt 
      { moveto }{ 
          dup
          \psk@yMinValue \pst@number\psyunit mul lt 
          { moveto }{ L } ifelse 
      } ifelse
}
\def\endqp@polygon{%
  \addto@pscode{closepath}%
  \end@ClosedObj}
\def\testqp@polygon{%
  \ifdim\pslinearc>\z@\else
    \ifshowpoints\else
      \@psttrue
    \fi
  \fi}
\def\beginqp@dots{%
  \psk@dotsize
  \@nameuse{psds@\psk@dotstyle}
  Dot }
\def\doqp@dots{Dot }
\def\endqp@dots{\end@SpecialObj}
\def\testqp@dots{\@psttrue}
\def\beginqp@bezier{/n 0 def \pst@oplineto}
\def\doqp@bezier{/n n 1 add def n 3 mod 0 eq { 
    dup \psk@yMaxValue\space \pst@number\psyunit mul gt 
    { moveto pop pop pop pop}
    { dup \psk@yMinValue\space \pst@number\psyunit mul lt 
      { moveto pop pop pop pop}{ curveto } ifelse 
    } ifelse 
  } if
}
\def\endqp@bezier{%
  \addto@pscode{n 3 mod { pop pop } repeat}
  \end@OpenObj}%
\def\testqp@bezier{%
  \ifshowpoints\else
    \ifx\psk@arrowA\@empty
      \ifx\psk@arrowB\@empty
        \@psttrue
      \fi
    \fi
  \fi}
\def\beginqp@cbezier{/n 0 def moveto }
\def\doqp@cbezier{\doqp@bezier}
\def\endqp@cbezier{%
  \addto@pscode{n 3 mod { pop pop } repeat closepath}
  \end@ClosedObj}%
\def\testqp@cbezier{\ifshowpoints\else\@psttrue\fi}
\def\tx@LineToYAxis{LineToYAxis }
\def\psLineToYAxis@ii{%
\addto@pscode{\pst@cp \psline@iii \psk@Ox\space \pst@number\psxunit mul \tx@LineToYAxis}%
\end@OpenObj}
\def\tx@LineToXAxis{LineToXAxis }
\def\psLineToXAxis@ii{%
\addto@pscode{\pst@cp \psline@iii \psk@Oy\space \pst@number\psyunit mul \tx@LineToXAxis}%
\end@OpenObj}
\define@key[psset]{pst-plot}{PSfont}[NimbusRomNo9L-Regu]{\def\psk@PSfont{/#1 }}
\define@key[psset]{pst-plot}{valuewidth}[10]{\pst@getint{#1}\psk@valuewidth }
\define@key[psset]{pst-plot}{fontscale}[10]{\pst@checknum{#1}\psk@fontscale }
\define@key[psset]{pst-plot}{decimals}[-1]{\pst@getint{#1}\psk@decimals }
\psset[pst-plot]{PSfont=NimbusRomNo9L-Regu,fontscale=10,valuewidth=10,decimals=-1}
\newdimen\psxlabelsep
\newdimen\psylabelsep
\define@key[psset]{pst-plot}{xlabelsep}[5pt]{\pssetlength\psxlabelsep{#1}}
\define@key[psset]{pst-plot}{ylabelsep}[5pt]{\pssetlength\psylabelsep{#1}}
\psset[pst-plot]{xlabelsep=5pt,ylabelsep=5pt}

\newif\ifPst@valuesStar\Pst@valuesStarfalse
\newif\ifPst@xvalues\Pst@xvaluesfalse
\def\psvalues@ii{\addto@pscode{ false \tx@NArray \psvalues@iii }}
\def\psvalues@iii{
  \psk@PSfont findfont \psk@fontscale scalefont setfont 
  newpath 
  n { /yO ED /xO ED
      gsave
      \ifPst@xvalues
        xO \pst@number\psxunit div
      \else
        yO \pst@number\psyunit div
      \fi
      \psk@decimals 0 eq { cvi } if
      \psk@decimals 0 gt { 10 \psk@decimals exp dup 3 1 roll mul cvi exch div } if
      \psk@valuewidth string cvs /Str ED
      \ifPst@valuesStar
      Str stringwidth pop /yS \psk@fontscale def /xS ED 
      gsave newpath 
        xO \ifPst@xvalues \pst@number\pslabelsep add \fi 
        yO \ifPst@xvalues \psk@fontscale 4 div sub \else \pst@number\pslabelsep add \fi 
        moveto \ifx\psk@rot\@empty\else\psk@rot rotate \fi
        xS 0 rlineto 0 yS rlineto xS neg 0 rlineto 0 yS neg rlineto 
        closepath  1 setgray fill stroke 
      grestore 
      \fi
      xO \ifPst@xvalues \pst@number\pslabelsep add \fi
      yO \ifPst@xvalues \psk@fontscale 4 div sub \else \pst@number\pslabelsep add \fi 
      moveto \ifx\psk@rot\@empty\else\psk@rot rotate \fi 
      Str show 
      grestore } repeat 
}%
\def\beginplot@values{\Pst@valuesStarfalse\begin@SpecialObj}
\expandafter\def\csname beginplot@values*\endcsname{\Pst@valuesStartrue\begin@SpecialObj}
\def\beginplot@xvalues{\Pst@valuesStarfalse\begin@SpecialObj}
\expandafter\def\csname beginplot@xvalues*\endcsname{\Pst@valuesStartrue\begin@SpecialObj}
\def\endplot@values{%
  \Pst@xvaluesfalse%
  \psvalues@ii%
  \pst@stroke
  \end@SpecialObj}
\@namedef{endplot@values*}{\endplot@values}
\def\endplot@xvalues{%
  \Pst@xvaluestrue%
  \psvalues@ii%
  \pst@stroke
  \end@SpecialObj}
\@namedef{endplot@xvalues*}{\endplot@xvalues}
\def\psdataplot{\def\pst@par{}\pst@object{dataplot}}
\def\dataplot{\def\pst@par{}\pst@object{dataplot}}
\def\dataplot@i#1{%
  \pst@killglue
  \begingroup
    \use@par
    \@pstfalse
    \@nameuse{testqp@\psplotstyle}%
    \if@pst
      \dataplot@ii{\addto@pscode{#1}}%
    \else
      \listplot@ii{\addto@pscode{#1}}%
    \fi
  \endgroup
  \ignorespaces}
\def\dataplot@ii#1{%
  \@nameuse{beginplot@\psplotstyle}%
    \addto@pscode{%
      /Dx { \pst@number\psxunit mul /D { Dy } def } def
      /Dy { \pst@number\psyunit mul Do /D { Dx } def } def
      /D { /D { Dx } def } def
      /Do {
        \@nameuse{beginqp@\psplotstyle}%
        /Do { \@nameuse{doqp@\psplotstyle}} def
      } def}%
    #1
    \addto@pscode{ D }%
  \@nameuse{endqp@\psplotstyle}}
\def\psfileplot{\def\pst@par{}\pst@object{fileplot}}
\def\fileplot{\def\pst@par{}\pst@object{fileplot}}
\def\fileplot@i#1{%
  \pst@killglue%
  \begingroup%
    \use@par%
    \@pstfalse%
    \@nameuse{testqp@\psplotstyle}%
    \if@pst\dataplot@ii{\pst@readfile{#1}}\else\listplot@ii{\pst@altreadfile{#1}}\fi%
  \endgroup%
  \ignorespaces}
\def\pslistplot{\pst@object{listplot}}
\def\listplot{\pst@object{listplot}}
\def\listplot@i#1{\listplot@ii{\addto@pscode{#1}}}
\def\listplot@ii#1{%
  \@nameuse{beginplot@\psplotstyle}%
  \addto@pscode{/D {} def mark}%
  #1%
  \addto@pscode{
    \tx@PreparePoints
    \pst@number\psxunit
    \pst@number\psyunit
    \tx@ScalePoints
  }%
  \@nameuse{endplot@\psplotstyle}%
}
\define@boolkey[psset]{pst-plot}[Psk@]{xyValues}[true]{}
\define@boolkey[psset]{pst-plot}[Pst@]{ChangeOrder}[true]{}
\psset[pst-plot]{xyValues,ChangeOrder=false}
\pst@def{PreparePoints}<{%
  counttomark /m exch def
  /maxYValues \psk@plotNoMax\space def
  /YValuePos \psk@plotNo\space def
  /XValuePos \psk@plotNoX\space def
  /maxYvalue \ifx\empty\psk@plotYMax false \else true \fi def
  maxYvalue { /YMaxValue \psk@plotYMax\space def } if
  \ifPsk@xyValues\else 
    /mm m def
    /M m 1 add def
    m { mm exch M 2 roll /M M 1 add def /mm mm 1 sub def } repeat
    /m m dup add def
  \fi
  \ifPst@ChangeOrder
    /m0 m def
    m maxYValues 1 add div 1 sub cvi {
      m0 maxYValues 1 add roll /m0 m0 maxYValues 1 add sub def
    } repeat
  \fi
  /n m maxYValues 1 add div cvi def
  XValuePos 1 gt {
    n {
      maxYValues 1 add XValuePos neg roll    
      dup /XValue ED
      maxYValues 1 add XValuePos 1 sub  roll 
      pop XValue                             
      maxYValues 1 add 1 roll                
      m maxYValues 1 add  roll               
    } repeat
  } if 
  maxYValues 1 gt {
    n {
      maxYValues YValuePos 1 sub neg roll 
      maxYValues 1 sub { pop } repeat     
      /m m maxYValues 1 sub sub def
      m 2 roll
    } repeat
  } if 
  /xMax -99999 def /yMax -99999 def
  /xP 0 def /yP 0 def
  m copy
  n {
    /y exch def /x exch def
    xMax x lt { /xMax x def } if
    yMax y lt {/yMax y def } if
    xP x gt { /xP x def } if
    yP y gt { /yP y def } if
  } repeat
  \psk@xStep\space 0 gt \psk@yStep\space 0 gt or (\psk@xStart) length 0 gt or
  (\psk@yStart) length 0 gt or (\psk@xEnd) length 0 gt or (\psk@yEnd) length 0 gt or {
    (\psk@xStart) length 0 gt {\psk@xStart\space }{ xP } ifelse /xStart exch def
    (\psk@yStart) length 0 gt {\psk@yStart\space }{ yP } ifelse /yStart exch def
    (\psk@xEnd) length 0 gt { \psk@xEnd\space }{ xMax } ifelse /xEnd exch def
    (\psk@yEnd) length 0 gt { \psk@yEnd\space }{ yMax } ifelse /yEnd exch def
    n {
      m -2 roll
      2 copy /yVal exch def /xVal exch def
      xVal xP ge
      yVal yP ge and
      xVal xEnd le and
      yVal yEnd le and
      xVal xStart ge and
      yVal yStart ge and {
        /xP xP \psk@xStep\space add def
        /yP yP \psk@yStep\space add def
      }{%
        pop pop
        /m m 2 sub def
      } ifelse
    } repeat
  }{%
    /ncount 1 def
    (\psk@nEnd) length 0 gt { \psk@nEnd\space }{ m } ifelse 
    /nEnd exch def
    n {
      m -2 roll
      \psk@nStep\space 1 gt { ncount \psk@nStart\space sub \psk@nStep\space mod 0 eq }{ true } ifelse
        ncount nEnd le and
        ncount \psk@nStart\space ge and not {
          pop pop
          /m m 2 sub def
        } if
        /ncount ncount 1 add def
      } repeat
  } ifelse
  maxYvalue { 
    /m0 m def
    n {
      dup YMaxValue gt { pop pop /m m 2 sub def } if
      m -2 roll 
    } repeat
  } if
}>
\define@boolkey[psset]{pst-plot}[Pst@]{polarplot}[true]{}
\define@boolkey[psset]{pst-plot}[Pst@]{VarStep}[true]{}
\define@key[psset]{pst-plot}{PlotDerivative}{\def\psk@PlotDerivative{#1}}%
\define@key[psset]{pst-plot}{VarStepEpsilon}{\def\psk@VarStepEpsilon{#1}}%
\define@key[psset]{pst-plot}{method}{\def\psk@method{#1}}
\def\@rkiv{rk4}
\def\@varrkiv{varrkiv}
\def\@adams{adams}
\def\@default{default}
\psset[pst-plot]{VarStep=false,PlotDerivative=none,VarStepEpsilon=default,polarplot=false,method={}}
\def\psplotinit#1{\xdef\psplot@init{#1 }}
\def\psplot@init{}
\def\psplot{\def\pst@par{}\pst@object{psplot}}
\def\psplot@i#1#2{\@ifnextchar[{\psplot@x{#1}{#2}}{\psplot@x{#1}{#2}[]}}
\def\psplot@x#1#2[#3]#4{%
  \pst@killglue
  \begingroup
    \use@par
    \@nameuse{beginplot@\psplotstyle}%
    \ifPst@polarplot
      \addto@pscode{
        \psplot@init
        #3 
        /x #1 def
        /x1 #2 def
        /dx x1 x sub \psk@plotpoints div def
        /F@pstplot \ifPst@algebraic (#4)
                    \ifx\psk@PlotDerivative\@none\else
                      \psk@PlotDerivative\space { (x) tx@Derive begin Derive end } repeat
                    \fi\space
                    tx@AlgToPs begin AlgToPs end cvx
                 \else { #4 } \fi  def
        \ifPst@VarStep
          /StillZero 0 def /LastNonZeroStep dx def
          /F2@pstplot tx@Derive begin (#4) (x) Derive (x) Derive end
                     \ifx\psk@PlotDerivative\@none\else
                       \psk@PlotDerivative\space { (x) tx@Derive begin Derive end } repeat
                     \fi\space
                    tx@AlgToPs begin AlgToPs end cvx def
          /epsilon12 \ifx\psk@VarStepEpsilon\@default tx@Derive begin F2@pstplot end dx 3 exp abs mul abs
                    \else\psk@VarStepEpsilon\space 12 mul \fi def
          /ComputeStep {
            dup 1e-4 lt
            { pop StillZero 2 ge { LastNonZeroStep 2 mul } { LastNonZeroStep } ifelse /StillZero StillZero 1 add def }
            { epsilon12 exch div 1 3 div exp /StillZero 0 def }
            ifelse } bind def
        \fi
        /xy {
          F@pstplot x \ifPst@algebraic RadtoDeg \fi PtoC
          \pst@number\psyunit mul exch
          \pst@number\psxunit mul exch
        } def}%
    \else
    \addto@pscode{
      \psplot@init
      #3 
      /x #1 def
      /x1 #2 def
      /dx x1 x sub \psk@plotpoints div def
      /F@pstplot \ifPst@algebraic (#4)
                    \ifx\psk@PlotDerivative\@none\else
                      \psk@PlotDerivative\space { (x) tx@Derive begin Derive end } repeat
                    \fi\space
                    tx@AlgToPs begin AlgToPs end cvx
                 \else { #4 } \fi  def
      \ifPst@VarStep
         /StillZero 0 def /LastNonZeroStep dx def
         /F2@pstplot tx@Derive begin (#4) (x) Derive (x) Derive end
                     \ifx\psk@PlotDerivative\@none\else
                       \psk@PlotDerivative\space { (x) tx@Derive begin Derive end } repeat
                     \fi\space
                    tx@AlgToPs begin AlgToPs end cvx def
         /epsilon12 \ifx\psk@VarStepEpsilon\@default tx@Derive begin F2@pstplot end dx 3 exp abs mul abs
                    \else\psk@VarStepEpsilon\space 12 mul \fi def
         /ComputeStep {
           dup 1e-4 lt
           { pop StillZero 2 ge { LastNonZeroStep 2 mul } { LastNonZeroStep } ifelse /StillZero StillZero 1 add def }
           { epsilon12 exch div 1 3 div exp /StillZero 0 def }
           ifelse } bind def
      \fi
      /xy { x \pst@number\psxunit mul F@pstplot \pst@number\psyunit mul
      } def}%
    \fi
    \gdef\psplot@init{}%
    \ifx\pslinestyle\psls@@symbol
      \psplot@iii
    \else
      \@pstfalse
      \@nameuse{testqp@\psplotstyle}%
      \if@pst\psplot@ii\else\psplot@iii\fi
    \fi
  \endgroup
  \ignorespaces}
\def\psplot@ii{%
  \ifPst@VarStep%
    \addto@pscode{%
      mark xy \@nameuse{beginqp@\psplotstyle}
      { F2@pstplot abs ComputeStep
        x 2 copy add dup x1 gt {pop x1} if /x exch def F2@pstplot abs ComputeStep
        /x 3 -1 roll def 2 copy gt { exch } if pop
        /x x 3 -1 roll add dup x1 gt {pop x1} if def
        xy \@nameuse{doqp@\psplotstyle}
        x x1 eq { exit } if} loop}%
  \else
    \pst@killglue%
    \addto@pscode{
      /ps@Exit false def
      xy \@nameuse{beginqp@\psplotstyle}
      \ifx\psk@method\@varrkiv\else\psk@plotpoints 1 sub \fi {
        /x x dx add \ifx\psk@method\@varrkiv  dup x1 gt { pop x1 } if \fi def
        xy \@nameuse{doqp@\psplotstyle}
        \ifx\psk@method\@varrkiv  x x1 eq { exit } if \fi
      } 
      ps@Exit { exit } if
      \ifx\psk@method\@varrkiv loop \else repeat \fi
      ps@Exit not {
        /x x1 def
        xy \@nameuse{doqp@\psplotstyle}
      } if }%
  \fi%
  \@nameuse{endqp@\psplotstyle}}
\def\psplot@iii{%
  \ifPst@VarStep%
    \addto@pscode{
      /n 2 def
      mark
      { xy n 2 roll F2@pstplot abs
        ComputeStep x 2 copy add dup x1 gt {pop x1} if
        /x exch def F2@pstplot abs ComputeStep
        /x 3 -1 roll def 2 copy gt { exch } if pop
        /x x 3 -1 roll dup /LastNonZeroStep exch def add dup x1 gt {pop x1} if def /n n 2 add def
        x x1 eq { exit } if } loop
      xy 
      n 2 roll}%
  \else\pst@killglue%
    \addto@pscode{
      mark
      /n 2 def
      \ifx\psk@method\@varrkiv\else\psk@plotpoints\fi {
        xy
        n 2 roll
        /n n 2 add def
        /x x dx add \ifx\psk@method\@varrkiv  dup x1 gt { pop x1 } if \fi def
        \ifx\psk@method\@varrkiv  x x1 eq { exit } if \fi
      } \ifx\psk@method\@varrkiv loop\else repeat \fi \space
      /x x1 def
      xy 
      2 copy \tx@UserCoor 2 array astore /FinalState ED
      n 2 roll}%
  \fi%
  \@nameuse{endplot@\psplotstyle}}
\def\psparametricplot{\pst@object{parametricplot}}
\def\parametricplot{\pst@object{parametricplot}}
\def\parametricplot@i#1#2{\@ifnextchar[{\parametricplot@x{#1}{#2}}{\parametricplot@x{#1}{#2}[]}}
\def\parametricplot@x#1#2[#3]{\@ifnextchar[{\parametricplot@xi{#1}{#2}[#3]}{\parametricplot@xi{#1}{#2}[#3][]}}
\def\parametricplot@xi#1#2[#3][#4]#5{%
  \pst@killglue%
  \begingroup%
    \use@par%
    \@nameuse{beginplot@\psplotstyle}%
    \addto@pscode{%
      #3 
      \psplot@init
      /t #1 def
      /t1 #2 def
      /dt t1 t sub \psk@plotpoints div def
      /F@pstplot \ifPst@algebraic (#5)
                    \ifx\psk@PlotDerivative\@none\else
                      \psk@PlotDerivative\space { (t) tx@Derive begin Derive end } repeat
                    \fi\space
                    tx@AlgToPs begin AlgToPs end cvx
                 \else { #5 } \fi  def
      \ifPst@VarStep
         /StillZero 0 def /LastNonZeroStep dt def
         /F2@pstplot tx@Derive begin (#5) (t) Derive (t) Derive end
                     \ifx\psk@PlotDerivative\@none\else
                       \psk@PlotDerivative\space { (t) tx@Derive begin Derive end } repeat
                     \fi\space
                    tx@AlgToPs begin AlgToPs end cvx def
         /epsilon12 \ifx\psk@VarStepEpsilon\@default
                       tx@Derive begin F2@pstplot end Pyth
                       dt 3 exp abs mul
                    \else\psk@VarStepEpsilon\space 12 mul \fi def
         /ComputeStep {
           dup 1e-4 lt
           { pop StillZero 2 ge { LastNonZeroStep 2 mul } { LastNonZeroStep } ifelse /StillZero StillZero 1 add def }
           { epsilon12 exch div 1 3 div exp /StillZero 0 def }
           ifelse } bind def
      \fi
      /xy {
        \ifPst@algebraic F@pstplot \else #5 \fi
        \pst@number\psyunit mul exch
        \pst@number\psxunit mul exch
      } def
      }%
    \gdef\psplot@init{}%
    \@pstfalse
    \@nameuse{testqp@\psplotstyle}%
    \if@pst\parametricplot@ii{#4}\else\parametricplot@iii{#4}\fi
  \endgroup%
  \ignorespaces}
\def\parametricplot@ii#1{
  \ifPst@VarStep%
    \addto@pscode{%
      mark xy \@nameuse{beginqp@\psplotstyle}
      { F2@pstplot Pyth ComputeStep
        t 2 copy add dup t1 gt {pop t1} if /t exch def F2@pstplot Pyth ComputeStep
        /t 3 -1 roll def 2 copy gt { exch } if pop
        /t t 3 -1 roll add dup t1 gt {pop t1} if def
        xy \@nameuse{doqp@\psplotstyle}
        t t1 eq { exit } if } loop}%
  \else\pst@killglue%
    \addto@pscode{%
      /ps@Exit false def
      xy \@nameuse{beginqp@\psplotstyle}
      \psk@plotpoints 1 sub {
        /t t dt add def
        xy \@nameuse{doqp@\psplotstyle}
        ps@Exit { exit } if 
      } repeat
      ps@Exit not {
        /t t1 def
        xy \@nameuse{doqp@\psplotstyle}
      } if 
    }%
  \fi%
  \addto@pscode{ #1 }%
  \@nameuse{endqp@\psplotstyle}}
\def\parametricplot@iii#1{%
  \ifPst@VarStep%
    \addto@pscode{%
      /n 2 def
      mark
      { xy n 2 roll F2@pstplot Pyth
        ComputeStep t 2 copy add dup t1 gt {pop t1} if
        /t exch def F2@pstplot Pyth ComputeStep
        /t 3 -1 roll def 2 copy gt { exch } if pop
        /t t 3 -1 roll dup /LastNonZeroStep exch def add dup t1 gt {pop t1} if def /n n 2 add def
        t t1 eq { exit } if } loop
      xy 
      2 copy \tx@UserCoor 2 array astore /FinalState ED
      n 2 roll}%
  \else\pst@killglue%
    \addto@pscode{
      mark
      /n 2 def
      \psk@plotpoints {
        xy
        n 2 roll
        /n n 2 add def
        /t t dt add def
      } repeat
      /t t1 def
      xy
      n 2 roll}%
  \fi%
  \addto@pscode{ #1 }%
  \@nameuse{endplot@\psplotstyle}}
\newdimen\psk@subticksize\psk@subticksize=\z@
\newdimen\pst@xticksizeA
\newdimen\pst@xticksizeB
\newdimen\pst@xticksizeC
\newdimen\pst@yticksizeA
\newdimen\pst@yticksizeB
\newdimen\pst@yticksizeC
\define@key[psset]{pst-plot}{ticks}[all]{\pst@expandafter\psset@@ticks{#1}\@nil\psk@ticks}
\def\psset@@ticks#1#2\@nil#3{%
  \ifx#1a\let#3\z@\else
    \ifx#1x\let#3\@ne\else
      \ifx#1y\let#3\tw@\else
        \ifx#1n\let#3\thr@@\else
          \@pstrickserr{Bad argument: `#1#2'}\@ehpa
  \fi\fi\fi\fi}
\psset[pst-plot]{ticks=all}
\define@key[psset]{pst-plot}{labels}[all]{\pst@expandafter\psset@@ticks{#1}\@nil\psk@labels}
\psset[pst-plot]{labels=all}
\define@key[psset]{pst-plot}{Ox}[0]{\def\psk@Ox{#1}}
\psset[pst-plot]{Ox=0}
\define@key[psset]{pst-plot}{Dx}[1]{\def\psk@Dx{#1}}
\psset[pst-plot]{Dx=1}
\define@key[psset]{pst-plot}{dx}[0]{%
  \pssetxlength\pst@dimg{#1}%
  \edef\psk@dx{\number\pst@dimg}}
\psset[pst-plot]{dx=0}
\define@key[psset]{pst-plot}{Oy}[0]{\def\psk@Oy{#1}}
\psset[pst-plot]{Oy=0}
\define@key[psset]{pst-plot}{Dy}[1]{\def\psk@Dy{#1}}
\psset[pst-plot]{Dy=1}
\define@key[psset]{pst-plot}{dy}[0]{%
  \pssetylength\pst@dimg{#1}%
  \edef\psk@dy{\number\pst@dimg}}
\psset[pst-plot]{dy=0}
\define@boolkey[psset]{pst-plot}[]{showXorigin}[true]{}
\define@boolkey[psset]{pst-plot}[]{showYorigin}[true]{}
\define@boolkey[psset]{pst-plot}[]{showorigin}[true]{%
  \ifshoworigin
    \showXorigintrue\showYorigintrue
  \else
    \showXoriginfalse\showYoriginfalse
  \fi
}
\psset[pst-plot]{showorigin=true}
\long\def\psrotatebox#1#2{%
  \leavevmode
  \Grot@setangle{#1}%
  \setbox\z@\hbox{{#2}}%
  \Grot@x\z@
  \Grot@y\z@
  \Grot@box}
\def\Grot@setangle#1{\edef\Grot@angle{#1}}
\def\Grot@Px#1#2#3{%
        #1\Grot@cos#2%
        \advance#1-\Grot@sin#3}
\def\Grot@Py#1#2#3{%
        #1\Grot@sin#2%
        \advance#1\Grot@cos#3}
\def\Grot@box{%
  \begingroup
  \CalculateSin\Grot@angle
  \CalculateCos\Grot@angle
  \edef\Grot@sin{\UseSin\Grot@angle}%
  \edef\Grot@cos{\UseCos\Grot@angle}%
  \Grot@r\wd\z@  \advance\Grot@r-\Grot@x
  \Grot@l\z@     \advance\Grot@l-\Grot@x
  \Grot@h\ht\z@  \advance\Grot@h-\Grot@y
  \Grot@d-\dp\z@ \advance\Grot@d-\Grot@y
  \ifdim\Grot@sin\p@>\z@
    \ifdim\Grot@cos\p@>\z@
      \Grot@Py\Grot@height \Grot@r\Grot@h
      \Grot@Px\Grot@right  \Grot@r\Grot@d
      \Grot@Px\Grot@left   \Grot@l\Grot@h
      \Grot@Py\Grot@depth  \Grot@l\Grot@d
    \else
      \Grot@Py\Grot@height \Grot@r\Grot@d
      \Grot@Px\Grot@right  \Grot@l\Grot@d
      \Grot@Px\Grot@left   \Grot@r\Grot@h
      \Grot@Py\Grot@depth  \Grot@l\Grot@h
    \fi
  \else
    \ifdim\Grot@cos\p@<\z@
      \Grot@Py\Grot@height \Grot@l\Grot@d
      \Grot@Px\Grot@right  \Grot@l\Grot@h
      \Grot@Px\Grot@left   \Grot@r\Grot@d
      \Grot@Py\Grot@depth  \Grot@r\Grot@h
    \else
      \Grot@Py\Grot@height \Grot@l\Grot@h
      \Grot@Px\Grot@right  \Grot@r\Grot@h
      \Grot@Px\Grot@left   \Grot@l\Grot@d
      \Grot@Py\Grot@depth  \Grot@r\Grot@d
    \fi
  \fi
  \advance\Grot@height\Grot@y
  \advance\Grot@depth\Grot@y
  \Grot@Px\dimen@  \Grot@x\Grot@y
  \Grot@Py\dimen@ii \Grot@x\Grot@y
  \dimen@-\dimen@     \advance\dimen@-\Grot@left
  \dimen@ii-\dimen@ii \advance\dimen@ii\Grot@y
  \setbox\z@\hbox{%
    \kern\dimen@
    \raise\dimen@ii\hbox{\Grot@start\box\z@\Grot@end}}%
  \ht\z@\Grot@height
  \dp\z@-\Grot@depth
  \advance\Grot@right-\Grot@left\wd\z@\Grot@right
  \leavevmode\box\z@
  \endgroup}
\define@key[psset]{pst-plot}{labelFontSize}[{}]{\def\psk@xlabelFontSize{#1}\def\psk@ylabelFontSize{#1}}%
\define@key[psset]{pst-plot}{xlabelFontSize}[{}]{\def\psk@xlabelFontSize{#1}}%
\define@key[psset]{pst-plot}{ylabelFontSize}[{}]{\def\psk@ylabelFontSize{#1}}%
\define@boolkey[psset]{pst-plot}[Pst@]{mathLabel}[true]{%
  \ifPst@mathLabel%
    \Pst@xmathLabeltrue \Pst@ymathLabeltrue
    \def\pshlabel##1{$\psk@xlabelFontSize##1$}%
    \def\psvlabel##1{$\psk@ylabelFontSize##1$}%
  \else%
    \Pst@xmathLabelfalse \Pst@ymathLabelfalse
    \def\pshlabel##1{\psk@xlabelFontSize##1}%
    \def\psvlabel##1{\psk@ylabelFontSize##1}%
  \fi}
\define@boolkey[psset]{pst-plot}[Pst@]{xmathLabel}[true]{%
  \ifPst@xmathLabel%
    \def\pshlabel##1{$\psk@xlabelFontSize##1$}\else\def\pshlabel##1{\psk@xlabelFontSize##1}\fi}
\define@boolkey[psset]{pst-plot}[Pst@]{ymathLabel}[true]{%
  \ifPst@ymathLabel%
    \def\psvlabel##1{$\psk@ylabelFontSize##1$}\else\def\psvlabel##1{\psk@ylabelFontSize##1}\fi}
\psset[pst-plot]{labelFontSize={},mathLabel}
\define@boolkey[psset]{pst-plot}[Pst@]{xAxis}[true]{}
\define@boolkey[psset]{pst-plot}[Pst@]{yAxis}[true]{}
\define@boolkey[psset]{pst-plot}[Pst@]{xyAxes}[true]{%
    \@nameuse{Pst@xAxis#1}\@nameuse{Pst@yAxis#1}}%
\psset[pst-plot]{xAxis,yAxis}%
\define@key[psset]{pst-plot}{xlabelPos}[b]{\pst@expandafter\psset@@xlabelPos#1\@nil}
\define@key[psset]{pst-plot}{ylabelPos}[l]{\pst@expandafter\psset@@ylabelPos#1\@nil}
\def\psset@@xlabelPos#1#2\@nil{%
  \ifx#1t\relax
    \let\psk@xlabelPos\tw@
    \pst@xticksizeC=\pst@xticksizeB
  \else
    \ifx#1a\relax
      \let\psk@xlabelPos\@ne 
      \pst@xticksizeC=\z@
    \else
      \def\psk@xlabelPos{\z@}
      \pst@xticksizeC=\pst@xticksizeA
  \fi\fi
}
\def\psset@@ylabelPos#1#2\@nil{%
  \ifx#1r\relax
    \def\psk@ylabelPos{\tw@}
    \pst@yticksizeC=\pst@yticksizeB
  \else
    \ifx#1a\relax
      \def\psk@ylabelPos{\@ne}
      \pst@yticksizeC=\z@
    \else 
      \def\psk@ylabelPos{\z@}
      \pst@yticksizeC=\pst@yticksizeA
  \fi\fi
}
\psset[pst-plot]{xlabelPos=b, ylabelPos=l}%
\define@key[psset]{pst-plot}{xyDecimals}[{}]{\def\psk@xDecimals{#1}\def\psk@yDecimals{#1}}
\define@key[psset]{pst-plot}{xDecimals}[{}]{\def\psk@xDecimals{#1}}
\define@key[psset]{pst-plot}{yDecimals}[{}]{\def\psk@yDecimals{#1}}
\psset[pst-plot]{xyDecimals={}}%
\define@key[psset]{pst-plot}{xlogBase}[{}]{\def\psk@xlogBase{#1}}
\define@key[psset]{pst-plot}{ylogBase}[{}]{\def\psk@ylogBase{#1}}
\define@key[psset]{pst-plot}{xylogBase}[{}]{\def\psk@xlogBase{#1}\def\psk@ylogBase{#1}}%
\psset[pst-plot]{xylogBase={}}%
\define@key[psset]{pst-plot}{trigLabelBase}[0]{\pst@getint{#1}{\psk@xtrigLabelBase}\let\psk@ytrigLabelBase\psk@xtrigLabelBase}
\define@key[psset]{pst-plot}{xtrigLabelBase}[0]{%
  \pst@getint{#1}{\psk@xtrigLabelBase}\psset{xtrigLabels}}
\define@key[psset]{pst-plot}{ytrigLabelBase}[0]{%
  \pst@getint{#1}{\psk@ytrigLabelBase}\psset{ytrigLabels}}
\psset[pst-plot]{trigLabelBase=0}
\define@key[psset]{pst-plot}{fractionLabelBase}[0]{\pst@getint{#1}{\psk@xfractionLabelBase}\let\psk@yfractionLabelBase\psk@xfractionLabelBase}
\define@key[psset]{pst-plot}{xfractionLabelBase}[0]{%
  \pst@getint{#1}{\psk@xfractionLabelBase}\psset{xfractionLabels}}
\define@key[psset]{pst-plot}{yfractionLabelBase}[0]{%
  \pst@getint{#1}{\psk@yfractionLabelBase}\psset{yfractionLabels}}
\psset[pst-plot]{fractionLabelBase=0}
\def\setDefaulthLabels{%
  \ifPst@xmathLabel\def\pshlabel##1{$\psk@xlabelFontSize##1$}\else\def\pshlabel##1{\psk@xlabelFontSize##1}\fi
  \def\pst@@@hlabel##1{%
      \edef\@xyDecimals{\psk@xDecimals}%
      \ifnum\psk@labels<\tw@\relax
        \ifx\psk@xlogBase\@empty
          \pshlabel{\psk@xlabelFontSize\expandafter\@LabelComma##1..\@nil\psk@xlabelFactor}%
        \else
          \ifPst@xmathLabel
            \pshlabel{\psk@xlabelFontSize\psk@xlogBase^{\expandafter\@stripDecimals##1..\@nil}}%
          \else
            \pshlabel{\psk@xlabelFontSize\psk@xlogBase\textsuperscript{\expandafter\@stripDecimals##1..\@nil}}%
          \fi
        \fi
      \fi
    }%
    \ifPst@xmathLabel\def\pshlabel##1{$\psk@xlabelFontSize##1$}\else\def\pshlabel##1{\psk@xlabelFontSize##1}\fi
}
\def\setTrighLabels{%
    \def\pst@@@hlabel##1{\pshlabel{##1}}%
    \def\pshlabel##1{%
      \ifnum\psk@xtrigLabelBase<2
        \def\de@nominator{\@ne}\else\def\de@nominator{\psk@xtrigLabelBase}\fi
      \def\pst@tempA{##1}%
      \pst@abs{\pst@tempA}\pst@cntm 
      \pst@mod{\pst@cntm}{\de@nominator}\pst@cntp 
      \ifnum\@ne>\pst@cntp                  
        \pst@cnto=\pst@cntm \divide\pst@cnto by \de@nominator  
	\ifPst@xmathLabel
          $\psk@xlabelFontSize
  	  \ifnum\pst@tempA<0 -\fi
          \ifnum\pst@cnto=\@ne                
            \pi                 	      
          \else
            \ifnum\pst@cnto=\z@ 0\else
            \the\pst@cnto\pi 	              
          \fi\fi$%
	\else%
          \psk@xlabelFontSize
  	  \ifnum\pst@tempA<0 -\fi
          \ifnum\pst@cnto=\@ne
            $\pi$
          \else%
            \the\pst@cnto$\pi$
          \fi%
	\fi%
      \else%
	\ifPst@xmathLabel%
          $\psk@xlabelFontSize%
          \ifnum\pst@cntp=\@ne
            \if\pst@cntm=\@ne%
              \frac{\pi}{\de@nominator}
            \else\ifnum\pst@tempA=-1 \frac{-\pi}{\de@nominator}%
              \else \ifnum\pst@tempA=1 \frac{\pi}{\de@nominator}%
                \else\frac{\pst@tempA\pi}{\de@nominator}
            \fi\fi\fi%
          \else%
            \ifnum\pst@tempA=1 \frac{\pi}{\de@nominator}%
            \else\ifnum\pst@tempA=\de@nominator \pi%
              \else\frac{\pst@tempA\pi}{\de@nominator}%
          \fi\fi\fi$%
	\else%
          \psk@xlabelFontSize%
          \ifnum\pst@cntp=\@ne
            \if\pst@cntm=\@ne%
              $\frac{\pi}{\de@nominator}$
            \else\ifnum\pst@tempA=-1 $\frac{-\pi}{\de@nominator}$%
              \else \ifnum\pst@tempA=1 $\frac{\pi}{\de@nominator}$%
                \else$\frac{\pst@tempA\pi}{\de@nominator}$
            \fi\fi\fi
          \else
            \ifnum\pst@tempA=1 $\frac{\pi}{\de@nominator}$%
            \else\ifnum\pst@tempA=\de@nominator $\pi$%
              \else$\frac{\pst@tempA\pi}{\de@nominator}$%
          \fi\fi\fi
	\fi
      \fi
    }%
}
\def\setDefaultvLabels{%
  \ifPst@ymathLabel\def\psvlabel##1{$\psk@ylabelFontSize##1$}\else\def\psvlabel##1{\psk@ylabelFontSize##1}\fi
    \def\pst@@@vlabel##1{%
      \edef\@xyDecimals{\psk@yDecimals}%
      \ifodd\psk@labels 
      \else%
        \ifx\psk@ylogBase\@empty
          \psvlabel{\expandafter\@LabelComma##1..\@nil\psk@ylabelFactor}%
        \else%
          \ifPst@ymathLabel%
            \psvlabel{\psk@ylogBase^{\expandafter\@stripDecimals##1..\@nil }}%
	  \else
            \psvlabel{\psk@ylogBase\textsuperscript{\expandafter\@stripDecimals##1..\@nil }}%
          \fi%
        \fi%
      \fi%
    }%
}%
\def\setTrigvLabels{%
  \def\pst@@@vlabel##1{\psvlabel{##1}}%
    \def\psvlabel##1{%
      \ifnum\psk@ytrigLabelBase<2 \def\de@nominator{\@ne}\else\def\de@nominator{\psk@ytrigLabelBase}\fi
      \def\pst@tempA{##1} 
      \pst@abs{\pst@tempA}\pst@cntm 
      \pst@mod{\pst@cntm}{\de@nominator}\pst@cntp 
      \ifnum\@ne>\pst@cntp                  
        \pst@cnto=\pst@cntm \divide\pst@cnto by \de@nominator  
	\ifPst@ymathLabel%
          $\psk@ylabelFontSize
  	  \ifnum\pst@tempA<0 -\fi
          \ifnum\pst@cnto=\@ne                
            \pi                 	      
          \else
            \the\pst@cnto\pi 	              
          \fi$%
	\else%
          \psk@ylabelFontSize%
  	  \ifnum\pst@tempA<0 -\fi
          \ifnum\pst@cnto=\@ne
            $\pi$
          \else
            \the\pst@cnto$\pi$
          \fi
	\fi
      \else
	\ifPst@ymathLabel%
          $\psk@ylabelFontSize
          \ifnum\pst@cntp=\@ne
            \if\pst@cntm=\@ne%
              \frac{\pi}{\de@nominator}
            \else\ifnum\pst@tempA=-1 \frac{-\pi}{\de@nominator}%
              \else \ifnum\pst@tempA=1 \frac{\pi}{\de@nominator}%
                \else\frac{\pst@tempA\pi}{\de@nominator}
            \fi\fi\fi%
          \else%
            \ifnum\pst@tempA=1 \frac{\pi}{\de@nominator}%
            \else\ifnum\pst@tempA=\de@nominator \pi%
              \else\frac{\pst@tempA\pi}{\de@nominator}%
          \fi\fi\fi$%
	\else
          \psk@ylabelFontSize
          \ifnum\pst@cntp=\@ne
            \if\pst@cntm=\@ne
              $\frac{\pi}{\de@nominator}$
            \else\ifnum\pst@tempA=-1 $\frac{-\pi}{\de@nominator}$%
              \else \ifnum\pst@tempA=1 $\frac{\pi}{\de@nominator}$%
                \else$\frac{\pst@tempA\pi}{\de@nominator}$
            \fi\fi\fi
          \else
            \ifnum\pst@tempA=1 $\frac{\pi}{\de@nominator}$%
            \else\ifnum\pst@tempA=\de@nominator $\pi$%
              \else$\frac{\pst@tempA\pi}{\de@nominator}$%
          \fi\fi\fi
	\fi
      \fi
    }%
}
\def\setFractionvLabels{%
  \def\pst@@@vlabel##1{\psvlabel{##1}}%
  \def\psvlabel##1{%
      \ifnum\psk@yfractionLabelBase<2 \def\de@nominator{\@ne}\else\def\de@nominator{\psk@yfractionLabelBase}\fi
      \def\pst@tempA{##1}%
      \pst@abs{\pst@tempA}\pst@cntm 
      \pst@mod{\pst@cntm}{\de@nominator}\pst@cntp 
      \ifnum\@ne>\pst@cntp                  
        \pst@cnto=\pst@cntm \divide\pst@cnto by \de@nominator  
	\ifPst@ymathLabel$\psk@ylabelFontSize\ifnum\pst@tempA<0 -\fi\the\pst@cnto\psk@ylabelFactor$%
	\else             \psk@ylabelFontSize\ifnum\pst@tempA<0 -\fi\the\pst@cnto\psk@ylabelFactor
	\fi
      \else
	\ifPst@ymathLabel
          $\psk@ylabelFontSize
          \ifnum\pst@cntp=\@ne                
            \if\pst@cntm=\@ne
              \frac{1}{\de@nominator}\psk@ylabelFactor
            \else\ifnum\pst@tempA=-1 \frac{-1}{\de@nominator}\psk@ylabelFactor%
              \else \ifnum\pst@tempA=1 \frac{1}{\de@nominator}\psk@ylabelFactor%
                \else\frac{\pst@tempA}{\de@nominator}\psk@ylabelFactor
            \fi\fi\fi
          \else
            \ifnum\pst@tempA=1 \frac{1}{\de@nominator}\psk@ylabelFactor%
            \else\ifnum\pst@tempA=\de@nominator 1\psk@xlabelFactor \else\frac{\pst@tempA}{\de@nominator}\psk@ylabelFactor%
          \fi\fi\fi$
	\else
          \psk@ylabelFontSize
          \ifnum\pst@cntp=\@ne
            \if\pst@cntm=\@ne
              $\frac{1}{\de@nominator}\psk@ylabelFactor$
            \else\ifnum\pst@tempA=-1 $\frac{-1}{\de@nominator}\psk@ylabelFactor$%
              \else \ifnum\pst@tempA=1 $\frac{1}{\de@nominator}\psk@ylabelFactor$%
                \else$\frac{\pst@tempA}{\de@nominator}\psk@ylabelFactor$
            \fi\fi\fi%
          \else%
            \ifnum\pst@tempA=1 $\frac{1}{\de@nominator}\psk@ylabelFactor$%
            \else\ifnum\pst@tempA=\de@nominator 1\psk@ylabelFactor
              \else$\frac{\pst@tempA}{\de@nominator}\psk@ylabelFactor$
          \fi\fi\fi
	\fi
      \fi
    }%
}
\def\setFractionhLabels{%
  \def\pst@@@hlabel##1{\pshlabel{##1}}%
  \def\pshlabel##1{%
      \ifnum\psk@xfractionLabelBase<2 \def\de@nominator{\@ne}\else\def\de@nominator{\psk@xfractionLabelBase}\fi
      \def\pst@tempA{##1}%
      \pst@abs{\pst@tempA}\pst@cntm 
      \pst@mod{\pst@cntm}{\de@nominator}\pst@cntp
      \ifnum\@ne>\pst@cntp                  
        \pst@cnto=\pst@cntm \divide\pst@cnto by \de@nominator  
	\ifPst@xmathLabel$\psk@xlabelFontSize\ifnum\pst@tempA<0 -\fi\the\pst@cnto\psk@xlabelFactor$%
	\else             \psk@xlabelFontSize\ifnum\pst@tempA<0 -\fi\the\pst@cnto\psk@xlabelFactor
	\fi
      \else
	\ifPst@xmathLabel
          $\psk@xlabelFontSize
          \ifnum\pst@cntp=\@ne
            \if\pst@cntm=\@ne \frac{1}{\de@nominator}\psk@xlabelFactor
            \else\ifnum\pst@tempA=-1 \frac{-1}{\de@nominator}\psk@xlabelFactor%
              \else\ifnum\pst@tempA=1 \frac{1}{\de@nominator}\psk@xlabelFactor%
                \else\frac{\pst@tempA}{\de@nominator}\psk@xlabelFactor
            \fi\fi\fi%
          \else%
            \ifnum\pst@tempA=1 \frac{1}{\de@nominator}\psk@xlabelFactor%
            \else\ifnum\pst@tempA=\de@nominator 1\psk@xlabelFactor\else\frac{\pst@tempA}{\de@nominator}\psk@xlabelFactor%
          \fi\fi\fi$
	\else
          \psk@xlabelFontSize
          \ifnum\pst@cntp=\@ne
            \if\pst@cntm=\@ne $\frac{1}{\de@nominator}\psk@xlabelFactor$
            \else\ifnum\pst@tempA=-1 $\frac{-1}{\de@nominator}\psk@xlabelFactor$%
              \else \ifnum\pst@tempA=1 $\frac{1}{\de@nominator}\psk@xlabelFactor$%
                \else$\frac{\pst@tempA}{\de@nominator}\psk@xlabelFactor$
            \fi\fi\fi
          \else
            \ifnum\pst@tempA=1 $\frac{1}{\de@nominator}\psk@xlabelFactor$%
            \else\ifnum\pst@tempA=\de@nominator 1\psk@xlabelFactor%
              \else$\frac{\pst@tempA}{\de@nominator}\psk@xlabelFactor$
          \fi\fi\fi
	\fi
      \fi
    }%
}
\define@boolkey[psset]{pst-plot}[Pst@]{xtrigLabels}[true]{%
  \ifPst@xtrigLabels \setTrighLabels \else \setDefaulthLabels\fi}
\define@boolkey[psset]{pst-plot}[Pst@]{ytrigLabels}[true]{%
  \ifPst@ytrigLabels \setTrigvLabels \else \setDefaultvLabels \fi}
\define@boolkey[psset]{pst-plot}[Pst@]{trigLabels}[true]{%
  \ifPst@trigLabels\psset[pst-plot]{xtrigLabels,ytrigLabels=false}
  \else            \psset[pst-plot]{xtrigLabels=false,ytrigLabels=false}%
  \fi}
\psset[pst-plot]{trigLabels=false}
\define@boolkey[psset]{pst-plot}[Pst@]{xfractionLabels}[true]{%
  \ifPst@xfractionLabels \setFractionhLabels \else \setDefaulthLabels \fi}
\define@boolkey[psset]{pst-plot}[Pst@]{yfractionLabels}[true]{%
  \ifPst@yfractionLabels \setFractionvLabels \else \setDefaultvLabels \fi}
\define@boolkey[psset]{pst-plot}[Pst@]{fractionLabels}[true]{%
  \ifPst@fractionLabels \setFractionhLabels\setFractionvLabels\Pst@xfractionLabelstrue\Pst@yfractionLabelstrue\fi}
\psset[pst-plot]{fractionLabels=false}
\def\psk@logLines{3}
\define@key[psset]{pst-plot}{logLines}[none]{\pst@expandafter\psset@@logLines#1\@nil\psk@logLines}
\def\psset@@logLines#1#2\@nil#3{%
  \ifx#1a\relax
    \let#3\z@
    \Pst@maxxTickstrue\Pst@maxyTickstrue
    \set@xticksize{0 4pt}\set@yticksize{0 4pt}%
    \def\psk@xsubticksize{1}\def\psk@ysubticksize{1}%
  \else
    \ifx#1x\relax
      \let#3\@ne
      \Pst@maxxTickstrue\Pst@maxyTicksfalse
      \set@xticksize{0 4pt}\def\psk@xsubticksize{1}%
    \else
      \ifx#1y\relax
        \let#3\tw@
	\Pst@maxyTickstrue\Pst@maxxTicksfalse
	\set@yticksize{0 4pt}\def\psk@ysubticksize{1}%
      \else
        \ifx#1n\let#3\thr@@\else
          \@pstrickserr{Bad argument: `#1#2'}\@ehpa
  \fi\fi\fi\fi}
\psset[pst-plot]{logLines=none}%
\define@key[psset]{pst-plot}{ylabelFactor}[\relax]{\def\psk@ylabelFactor{#1}}
\define@key[psset]{pst-plot}{xlabelFactor}[\relax]{\def\psk@xlabelFactor{#1}}
\define@boolkey[psset]{pst-plot}[Pst@]{showOriginTick}[true]{}%
\psset[pst-plot]{xlabelFactor=\relax,ylabelFactor=\relax,showOriginTick}%

\def\psxTick{\pst@object{psxTick}}
\def\psxTick@i{\@ifnextchar({\psxTick@ii{0}}\psxTick@ii}
\def\psxTick@ii#1(#2)#3{{%
  \pst@killglue
  \addbefore@par{arrows=-,linewidth=\psk@xtickwidth\pslinewidth}
  \ifPst@xtrigLabels\addto@par{xtrigLabels=false}\fi 
  \use@par
  \edef\temp@coor{(!#2 \pst@number\pst@xticksizeB \pst@number\psyunit div)(!#2 \pst@number\pst@xticksizeA \pst@number\psyunit div)}%
  \expandafter\psline\temp@coor
  \rput[t]{#1}(! \psk@origin 
                 #2 \pst@number\psxlabelsep \pst@number\pst@xticksizeB add
                 \pst@number\psyunit div neg ){\pshlabel{#3\vphantom{1}}}%
  }\ignorespaces}
\def\psyTick{\pst@object{psyTick}}
\def\psyTick@i{\@ifnextchar({\psyTick@ii{0}}\psyTick@ii}
\def\psyTick@ii#1(#2)#3{{%
  \pst@killglue
  \addbefore@par{arrows=-,linewidth=\psk@ytickwidth\pslinewidth}
  \ifPst@ytrigLabels \setDefaultvLabels \fi
  \use@par
  \edef\temp@coor{(!\pst@number\pst@yticksizeB \pst@number\psxunit div #2)(!\pst@number\pst@yticksizeA \pst@number\psxunit div #2)}%
  \expandafter\psline\temp@coor
    \rput[r]{#1}(!\psk@origin
                  \pst@number\pst@yticksizeB \pst@number\psylabelsep add
                  \pst@number\psxunit div neg #2){\psvlabel{#3}}}\ignorespaces}
\define@boolkey[psset]{pst-plot}[Pst@]{markPoint}[true]{}%
\psset[pst-plot]{markPoint}
\def\psCoordinates{\pst@object{psCoordinates}}
\def\psCoordinates@i(#1){%
  \pst@killglue%
  \begingroup
  \addbefore@par{showpoints=false,markPoint}
  \use@par
  \psline(#1|0,0)(#1)
  \psline(#1)(0,0|#1)%
  \ifPst@markPoint\psdot(#1)\fi%
  \endgroup
  \ignorespaces
}
\def\stripDecimals#1{\expandafter\@stripDecimals#1..\@nil}
\def\@stripDecimals#1.#2.#3\@nil{%
  \def\pst@dummy{#1}%
  \ifx\pst@dummy\@empty\the\@zero\else#1\fi
}
\newcount\@digitcounter\@digitcounter=0\relax
\def\@inc@digitcounter{\global\advance\@digitcounter by 1\relax}
\def\@get@digitcounter{\the\@digitcounter\relax}
\def\@Reset@digitcounter{\global\@digitcounter=0\relax}
\def\@zeroFill{%
  \ifnum \@xyDecimals>\@get@digitcounter
    \bgroup
      0\@inc@digitcounter\@zeroFill
    \egroup
  \fi
}
\def\@process@digits#1#2;{%
  \ifx *#1\@zeroFill\else#1\@inc@digitcounter 
  \ifnum\@xyDecimals>\@get@digitcounter\expandafter\@process@digits#2;\fi\fi%
}
\def\@writeDecimals#1{%
  \ifx\@xyDecimals\@empty
    \def\@tempa{#1}
    \ifx\@tempa\@empty
    \else\ifmmode\expandafter\mathord\expandafter{\psk@decimalSeparator}\else\psk@decimalSeparator\fi#1\fi%
  \else
    \ifnum\@xyDecimals>\@zero
      \ifmmode\expandafter\mathord\expandafter{\psk@decimalSeparator}\else\psk@decimalSeparator\fi%
        \@Reset@digitcounter
        \expandafter\@process@digits#1*;%
      \fi%
  \fi%
}
\def\@LabelComma#1.#2.#3\@nil{%
  \def\pst@tempA{#1}%
  \ifx\pst@tempA\@empty\the\@zero\else#1\fi
  \def\pst@tempA{#2}%
  \ifx\pst@tempA\@empty\@writeDecimals{}\else\@writeDecimals{#2}\fi
}
\def\set@xticksize#1{%
  \pst@expandafter\pst@getydimdim{#1} {} {}\@nil
  \ifdim\pst@dimm>\pst@dimn
    \pst@xticksizeA=\the\pst@dimn%
    \pst@xticksizeB=\the\pst@dimm%
  \else%
    \pst@xticksizeA=\the\pst@dimm%
    \pst@xticksizeB=\the\pst@dimn
  \fi%
  \edef\psk@xticksize{\pst@number\pst@xticksizeA \pst@number\pst@xticksizeB}%
  \ifnum\psk@xlabelPos<\z@\relax
    \pst@xticksizeC=\pst@dimn
  \else
    \pst@xticksizeC=\pst@dimm
  \fi
}
\def\set@yticksize#1{%
  \pst@expandafter\pst@getxdimdim{#1} {} {}\@nil
  \ifdim\pst@dimm>\pst@dimn\relax
    \pst@yticksizeA=\the\pst@dimn%
    \pst@yticksizeB=\the\pst@dimm%
  \else%
    \pst@yticksizeA=\the\pst@dimm%
    \pst@yticksizeB=\the\pst@dimn
  \fi%
  \edef\psk@yticksize{\pst@number\pst@yticksizeA \pst@number\pst@yticksizeB}%
  \ifnum\psk@ylabelPos<\z@	
    \pst@yticksizeC=\pst@dimn%
  \else%
      \pst@yticksizeC=\pst@dimo
  \fi%
}
\newif\ifPst@maxxTicks
\newif\ifPst@maxyTicks
\define@key[psset]{pst-plot}{ticksize}[-4pt 4pt]{%
  \def\pst@tempA{max}%
  \def\pst@tempB{#1}%
  \ifx\pst@tempA\pst@tempB%
    \Pst@maxxTickstrue\Pst@maxyTickstrue%
    \set@xticksize{0 4pt}\set@yticksize{0 4pt}%
  \else%
    \Pst@maxxTicksfalse\Pst@maxyTicksfalse%
    \set@xticksize{#1}\set@yticksize{#1}%
  \fi}
\define@key[psset]{pst-plot}{xticksize}{%
  \def\pst@tempA{max}%
  \def\pst@tempB{#1}%
  \ifx\pst@tempA\pst@tempB
    \Pst@maxxTickstrue\set@xticksize{0 4pt}%
  \else\set@xticksize{#1}\Pst@maxxTicksfalse\fi}
\define@key[psset]{pst-plot}{yticksize}{%
  \def\pst@tempA{max}%
  \def\pst@tempB{#1}%
  \ifx\pst@tempA\pst@tempB%
    \Pst@maxyTickstrue\set@yticksize{0 4pt}%
  \else\set@yticksize{#1}\Pst@maxyTicksfalse\fi}%
\psset[pst-plot]{ticksize=-4pt 4pt}
\define@key[psset]{pst-plot}{tickstyle}[full]{\pst@expandafter\psset@@tickstyle{#1}\@nil}
\def\psset@@tickstyle#1#2\@nil{%
  \ifx#1f\let\psk@tickstyle\z@\else			
    \ifx#1t\let\psk@tickstyle\@ne			
      \edef\psk@xticksize{0 \pst@number\pst@xticksizeB}%
      \edef\psk@yticksize{0 \pst@number\pst@yticksizeB}%
    \else\ifx#1b\let\psk@tickstyle\m@ne			
      \edef\psk@xticksize{\pst@number\pst@xticksizeA 0}%
      \edef\psk@yticksize{\pst@number\pst@yticksizeA 0}%
      \else\ifx#1i\let\psk@tickstyle\tw@
        \else\@pstrickserr{Bad tick style: `#1#2'}\@ehpa
  \fi\fi\fi\fi}
\psset[pst-plot]{tickstyle=full}
\define@key[psset]{pst-plot}{subticks}[1]{\def\psk@xsubticks{#1}\def\psk@ysubticks{#1}}
\define@key[psset]{pst-plot}{xsubticks}[1]{\def\psk@xsubticks{#1}}
\define@key[psset]{pst-plot}{ysubticks}[1]{\def\psk@ysubticks{#1}}
\define@key[psset]{pst-plot}{subticksize}[0.75]{\def\psk@xsubticksize{#1}\def\psk@ysubticksize{#1}}
\define@key[psset]{pst-plot}{xsubticksize}[0.75]{\def\psk@xsubticksize{#1}}
\define@key[psset]{pst-plot}{ysubticksize}[0.75]{\def\psk@ysubticksize{#1}}
\define@key[psset]{pst-plot}{tickwidth}[0.5\pslinewidth]{%
  \pst@getlength{#1}\psk@xtickwidth%
  \pst@getlength{#1}\psk@ytickwidth}
\define@key[psset]{pst-plot}{xtickwidth}[0.5\pslinewidth]{\pst@getlength{#1}\psk@xtickwidth}
\define@key[psset]{pst-plot}{ytickwidth}[0.5\pslinewidth]{\pst@getlength{#1}\psk@ytickwidth}
\define@key[psset]{pst-plot}{subtickwidth}[0.25\pslinewidth]{%
  \pst@getlength{#1}\psk@xsubtickwidth%
  \pst@getlength{#1}\psk@ysubtickwidth}
\define@key[psset]{pst-plot}{xsubtickwidth}[0.25\pslinewidth]{\pst@getlength{#1}\psk@xsubtickwidth}
\define@key[psset]{pst-plot}{ysubtickwidth}[0.25\pslinewidth]{\pst@getlength{#1}\psk@ysubtickwidth}
\define@key[psset]{pst-plot}{labelOffset}[0pt]{%
  \pst@getlength{#1}\psk@xlabelOffset%
  \pst@getlength{#1}\psk@ylabelOffset}
\define@key[psset]{pst-plot}{xlabelOffset}[0pt]{\pst@getlength{#1}\psk@xlabelOffset}
\define@key[psset]{pst-plot}{ylabelOffset}[0pt]{\pst@getlength{#1}\psk@ylabelOffset}
\define@key[psset]{pst-plot}{frameOffset}[0pt]{\pst@getlength{#1}\psk@frameOffset}
\define@key[psset]{pst-plot}{tickcolor}[black]{%
    \pst@getcolor{#1}\psk@xtickcolor%
    \pst@getcolor{#1}\psk@ytickcolor}
\define@key[psset]{pst-plot}{xtickcolor}[black]{\pst@getcolor{#1}\psk@xtickcolor}
\define@key[psset]{pst-plot}{ytickcolor}[black]{\pst@getcolor{#1}\psk@ytickcolor}
\define@key[psset]{pst-plot}{subtickcolor}[gray]{%
  \pst@getcolor{#1}\psk@xsubtickcolor%
  \pst@getcolor{#1}\psk@ysubtickcolor}
\define@key[psset]{pst-plot}{xsubtickcolor}[gray]{\pst@getcolor{#1}\psk@xsubtickcolor}
\define@key[psset]{pst-plot}{ysubtickcolor}[gray]{\pst@getcolor{#1}\psk@ysubtickcolor}
\define@key[psset]{pst-plot}{xticklinestyle}[solid]{%
  \@ifundefined{psls@#1}%
    {\@pstrickserr{Line style `#1' not defined}\@eha}%
    {\def\psxticklinestyle{#1}}}
\define@key[psset]{pst-plot}{xsubticklinestyle}[solid]{%
  \@ifundefined{psls@#1}%
    {\@pstrickserr{Line style `#1' not defined}\@eha}%
    {\def\psxsubticklinestyle{#1}}}
\define@key[psset]{pst-plot}{yticklinestyle}[solid]{%
  \@ifundefined{psls@#1}%
    {\@pstrickserr{Line style `#1' not defined}\@eha}%
    {\def\psyticklinestyle{#1}}}
\define@key[psset]{pst-plot}{ysubticklinestyle}[solid]{%
  \@ifundefined{psls@#1}%
    {\@pstrickserr{Line style `#1' not defined}\@eha}%
    {\def\psysubticklinestyle{#1}}}
\define@key[psset]{pst-plot}{ticklinestyle}[solid]{%
  \@ifundefined{psls@#1}%
    {\@pstrickserr{Line style `#1' not defined}\@eha}%
    {\def\psxticklinestyle{#1}\def\psyticklinestyle{#1}}}
\define@key[psset]{pst-plot}{subticklinestyle}[solid]{%
  \@ifundefined{psls@#1}%
    {\@pstrickserr{Line style `#1' not defined}\@eha}%
    {\def\psxsubticklinestyle{#1}\def\psysubticklinestyle{#1}}}
\psset[pst-plot]{subticksize=0.75,subticks=1,tickcolor=black,ticklinestyle=solid,
  subticklinestyle=solid,subtickcolor=gray,tickwidth=0.5\pslinewidth,
  subtickwidth=0.25\pslinewidth,labelOffset=0pt,frameOffset=0pt}
\define@key[psset]{pst-plot}{nStep}[1]{\def\psk@nStep{#1}}
\define@key[psset]{pst-plot}{nStart}[0]{\def\psk@nStart{#1}}
\define@key[psset]{pst-plot}{nEnd}[{}]{\def\psk@nEnd{#1}}
\define@key[psset]{pst-plot}{xStep}[0]{\def\psk@xStep{#1}}
\define@key[psset]{pst-plot}{yStep}[0]{\def\psk@yStep{#1}}
\define@key[psset]{pst-plot}{xStart}[{}]{\def\psk@xStart{#1}}
\define@key[psset]{pst-plot}{xEnd}[{}]{\def\psk@xEnd{#1}}
\define@key[psset]{pst-plot}{yStart}[{}]{\def\psk@yStart{#1}}
\define@key[psset]{pst-plot}{yEnd}[{}]{\def\psk@yEnd{#1}}
\define@key[psset]{pst-plot}{plotNoX}[1]{\def\psk@plotNoX{#1}}
\define@key[psset]{pst-plot}{plotNo}[1]{\def\psk@plotNo{#1}}
\define@key[psset]{pst-plot}{plotNoMax}[1]{\def\psk@plotNoMax{#1}}
\define@key[psset]{pst-plot}{plotYMax}[1]{\def\psk@plotYMax{#1}}
\psset[pst-plot]{nStep=1, nStart=0, nEnd={},%
  xStep=0, yStep=0, xStart={}, xEnd={},  yStart={}, yEnd={}, 
  plotNo=1,plotNoMax=1,plotNoX=1,plotYMax={}}%
\def\pstScalePoints(#1,#2)#3#4{%
  \def\pstXScale{#1 }%
  \def\pstYScale{#2 }%
  \def\pstXPSScale{#3 }%
  \def\pstYPSScale{#4 }%
  \pst@def{ScalePoints}<%
    /yVal ED /xVal ED
    /yPSOp { #4 yVal mul #2 mul } def
    /xPSOp { #3 xVal mul #1 mul } def
    counttomark dup dup cvi eq not { exch pop } if
    /m exch def /n m 2 div cvi def
    n {
      \ifPst@polarplot exch cvi 360 mod PtoC \fi  
      yPSOp m 1 roll xPSOp m 1 roll 
      /m m 2 sub
      def } repeat>%
}
\pstScalePoints(1,1){}{}
\def\psxs@none{\let\psk@arrowA\@empty\let\psk@arrowB\@empty\psxs@axes}
\def\psxs@axes{{%
  \ifPst@xAxis\psxs@@axes\pst@dima\pst@dimb\pst@dimc\pst@dimd{}{x}\fi%
  \ifPst@yAxis\psxs@@axes\pst@dima\pst@dimb\pst@dimc\pst@dimd{exch}{y}\fi%
}}
\newif\ifSpecialLabelsDone
\def\psaxes{\pst@object{psaxes}}
\def\psaxes@i{%
  \let\pst@par@save\pst@par
  \pst@getarrows\psaxes@ii}
\def\psaxes@ii(#1){\@ifnextchar({\psaxes@iii(#1)}{\psaxes@iv(0,0)(0,0)(#1)}}
\def\psaxes@iii(#1)(#2){\@ifnextchar({\psaxes@iv(#1)(#2)}{\psaxes@iv(#1)(#1)(#2)}}
\def\psaxes@iv(#1)(#2)(#3){\@ifnextchar[{\psaxes@v(#1)(#2)(#3)}{\psaxes@vii(#1)(#2)(#3)}}%
\def\psaxes@v(#1)(#2)(#3)[#4]{\@ifnextchar[{\psaxes@vi(#1)(#2)(#3)[#4]}{\psaxes@vi(#1)(#2)(#3)[#4][]}}%
\def\psaxes@vi(#1)(#2)(#3)[#4,#5][#6,#7]{%
  \psaxes@vii(#1)(#2)(#3)%
  \let\pst@par\pst@par@save
  \begingroup
  \SpecialCoor
  \use@par
  \ifshowgrid\psgrid[style=gridstyleA]\fi
  \uput{\psxlabelsep}[#5](#3|#1){#4}\uput{\psylabelsep}[#7](#1|#3){#6}%
  \endgroup
  \ignorespaces
}
\def\psaxes@vii(#1,#2)(#3,#4)(#5,#6){%
  \pst@killglue
  \begingroup
  \ifdim\pst@dimc<\z@\relax 
    \ifdim\pst@dimd<\z@\relax 
      \addbefore@par{xlabelPos=t,ylabelPos=r}%
  \fi\fi
  \use@par
  \pssetxlength\pst@dimc{#5}
  \pssetylength\pst@dimd{#6}
    \pssetxlength\pst@dimg{#1}
    \pssetylength\pst@dimh{#2}
    \pssetxlength\pst@dima{#3}
    \pssetylength\pst@dimb{#4}
    \pst@dima=\dimexpr\pst@dima-\pst@dimg\relax
    \pst@dimb=\dimexpr\pst@dimb-\pst@dimh\relax
    \pst@dimc=\dimexpr\pst@dimc-\pst@dimg\relax
    \pst@dimd=\dimexpr\pst@dimd-\pst@dimh\relax
   \setbox\pst@hbox=\hbox\bgroup
    \ifshowgrid\psgrid[style=gridstyleA]\fi
    \@nameuse{psxs@\psk@axesstyle}
    \ifPst@xAxis
      \SpecialLabelsDonefalse
      \begingroup
      \ifnum\psk@dx=\z@
        \pst@dimg=\psk@Dx\psxunit
        \ifdim\pst@dimg<\p@ 
          \pst@cnta=\psk@Dx
          \edef\psk@Dx{\the\numexpr-1*\pst@cnta}%
        \fi
        \edef\psk@dx{\number\pst@dimg}%
      \fi
      \pst@hlabels{\pst@dimc}{\psk@arrowB}{#3}{#5}
      \ifPst@yAxis\showoriginfalse\fi
      \pst@hlabels{\pst@dima}{\psk@arrowA}{#3}{#5}
      \endgroup
    \fi
    \ifPst@yAxis
      \SpecialLabelsDonefalse
      \begingroup
      \ifdim\pst@dima=\z@ \else\ifPst@xtrigLabels\showoriginfalse\fi\fi
      \ifnum\psk@dy=\z@
        \pst@dimg=\psk@Dy\psyunit
        \ifdim\pst@dimg<\p@ 
          \pst@cnta=\psk@Dy
          \edef\psk@Dy{\the\numexpr-1*\pst@cnta}%
        \fi
        \edef\psk@dy{\number\pst@dimg}%
      \fi
      \pst@vlabels{\pst@dimb}{\psk@arrowA}{#4}{#6}%
      \ifPst@xAxis\ifdim\pst@dima<\z@ \showoriginfalse\fi\fi 
      \pst@vlabels{\pst@dimd}{\psk@arrowB}{#4}{#6}%
      \endgroup
    \fi
  \egroup%
  \pssetxlength\pst@dimg{#1}%
  \pssetylength\pst@dimh{#2}%
  \leavevmode
  \psput@cartesian\pst@hbox
  \endgroup
  \ignorespaces
}
\newif\ifis@yAxis%
\def\psxs@@axes#1#2#3#4#5#6{
  \pst@killglue
  \begin@SpecialObj
    \ifx#6x\relax
      \is@yAxisfalse
      \ifnum\psk@dx=\z@
        \pst@dimg=\psk@Dx\psxunit
        \def\psk@dx{\number\pst@dimg}%
      \fi
    \else
      \is@yAxistrue
      \ifnum\psk@dy=\z@
        \pst@dimg=\psk@Dy\psyunit
        \def\psk@dy{\number\pst@dimg}%
      \fi
    \fi
    \let\pst@linetype\pst@arrowtype
    \def\pst@axes{axes}%
    \pst@addarrowdef
    \addto@pscode{
      /showOrigin \ifPst@showOriginTick true \else false \fi def 	
      \ifis@yAxis 0 \pst@number#4 \else \pst@number#3 0 \fi
      \ifis@yAxis 0 \pst@number#2 \else \pst@number#1 0 \fi
      ArrowA
      CP 4 2 roll
      ArrowB 
      2 copy
      /yEnd exch def /xEnd exch def
      \ifx\psk@axesstyle\@none   
        pop pop 
      \else
        gsave                              		
        L                                  		
        \@nameuse{psls@\pslinestyle}                 	
        stroke                                       	
        grestore
      \fi
      /yStart exch def
      /xStart exch def
      \number\psk@ticks\space dup 2 mod 0 eq \ifis@yAxis true \else false \fi and 
      exch 2 lt \ifis@yAxis false \else true \fi and or {
      /viceversa 
        \ifis@yAxis\pst@number#2 \pst@number#4 \else\pst@number#1 \pst@number#3 \fi
         gt { true }{ false } ifelse def           
      /epsilon 0.01 def                            
      /minTickline \ifis@yAxis \pst@number#1 \else \pst@number#2 \fi def
      /maxTickline \ifis@yAxis \pst@number#3 \else \pst@number#4 \fi def
      /dT \ifis@yAxis \psk@dy \else \psk@dx \fi\space abs  
        65536 div viceversa { neg } if def                 
      /DT \ifis@yAxis \psk@Dy \else \psk@Dx \fi\space abs viceversa { neg } if def  
      /subTNo \ifis@yAxis\psk@ysubticks\else\psk@xsubticks\fi \space def
      subTNo 0 gt { /dsubT dT subTNo div def}{ /dsubT 0 def } ifelse  
      \ifis@yAxis \psk@yticksize \else \psk@xticksize \fi
      /tickend exch def /tickstart exch def
      /Twidth \ifis@yAxis \psk@ytickwidth \else \psk@xtickwidth \fi\space def
      /subTwidth \ifis@yAxis \psk@ysubtickwidth \else \psk@xsubtickwidth \fi\space def
      /STsize \ifis@yAxis \psk@ysubticksize \else \psk@xsubticksize \fi\space def
      /TColor {
        \ifis@yAxis\pst@usecolor\psk@ytickcolor
        \else\pst@usecolor\psk@xtickcolor\fi\space } def
      /subTColor {
        \ifis@yAxis\pst@usecolor\psk@ysubtickcolor
        \else\pst@usecolor\psk@xsubtickcolor\fi\space } def
      /MinValue { \ifis@yAxis yStart \else xStart \fi
        \ifx\psk@arrowA\@empty\else 
          \psk@arrowsize\space CLW mul add \psk@arrowlength\space mul 
           viceversa { sub epsilon add }{ add epsilon sub } ifelse \fi } def
      /MaxValue { \ifis@yAxis yEnd \else xEnd \fi 
        \ifx\psk@arrowB\@empty\else
          \psk@arrowsize\space CLW mul add \psk@arrowlength\space mul 
           viceversa { add epsilon sub }{ sub epsilon add } ifelse \fi } def
      /logLines {
        \ifnum\psk@logLines=\z@ true \else         
          \ifnum\psk@logLines<\tw@                 
            \ifis@yAxis false \else true \fi       
          \else
            \ifnum\psk@logLines<\thr@@             
              \ifis@yAxis true \else false \fi     
            \else 
              false                                
            \fi
          \fi
        \fi
      } def
      /LSstroke {                                  
        \ifis@yAxis \@nameuse{psls@\psyticklinestyle}
        \else       \@nameuse{psls@\psxticklinestyle}\fi 
        stroke} def
      /subLSstroke {                               
        \ifis@yAxis \@nameuse{psls@\psysubticklinestyle}
        \else       \@nameuse{psls@\psxsubticklinestyle}\fi 
        stroke} def
      0 dT MaxValue 1 add {                        
        /cntTick exch def                          
        logLines {                                 
          gsave
          1 1 DT {
           1 sub /OffSet exch def
          -10 subTNo 1 add div dup 10 add exch dup -0.1 mul 1 add {                   
            /dx exch def                           
            /x dx log OffSet add \ifis@yAxis\pst@number\psyunit\else\pst@number\psxunit\fi\space mul cntTick add def       %
            x abs MaxValue abs le {                
	      \ifis@yAxis
	        \ifPst@maxyTicks true \else false \fi
	      \else
	        \ifPst@maxxTicks true \else false \fi
	      \fi
                { x minTickline #5 moveto
                  x maxTickline #5 lineto }
                { x tickstart STsize mul #5 moveto
                  x tickend STsize mul #5 lineto } ifelse
            } if
          } for } for
          subTwidth SLW subTColor                  
          subLSstroke
          grestore                                 
          stroke
          /dsubT 0 def                             
        } if 					   
        dsubT abs 0 gt {                           
          gsave                                    
          /cntsubTick cntTick dsubT add def
          subTNo 1 sub {
            cntsubTick abs MaxValue abs le {       
    	    \ifis@yAxis
              \ifPst@maxyTicks true \else false \fi
    	    \else
              \ifPst@maxxTicks true \else false \fi
    	    \fi
              { cntsubTick minTickline STsize mul #5 moveto
                cntsubTick maxTickline STsize mul #5 lineto }
              { cntsubTick tickstart STsize mul #5 moveto
                cntsubTick tickend STsize mul #5 lineto } ifelse
            }{ exit }  ifelse
            /cntsubTick cntsubTick dsubT add def
          } repeat 
          subTwidth SLW subTColor               
          subLSstroke
          grestore                              
        } if
        showOrigin {
          gsave
          \ifis@yAxis
            \ifPst@maxyTicks true \else false \fi
          \else
            \ifPst@maxxTicks true \else false \fi
          \fi
            { cntTick minTickline #5 moveto
              cntTick maxTickline #5 lineto }
            { cntTick tickstart #5 moveto        
              cntTick tickend #5 lineto } ifelse 
          Twidth SLW TColor                      
          LSstroke
          grestore
        }{ /showOrigin true def } ifelse         
      } for
      /showOrigin \ifPst@showOriginTick true \else false \fi def 
      /dT dT neg def                               
      /dsubT dsubT neg def
      0 dT MinValue epsilon viceversa { add }{ sub } ifelse {
        /cntTick exch def
        logLines {                                 
          gsave
          1 1 DT cvi {
            1 sub /OffSet exch def
          -10 subTNo 1 add div dup 10 add exch dup -0.1 mul 1 add {                   
            /dx exch def                           
            /x dx log OffSet add \ifis@yAxis\pst@number\psyunit\else\pst@number\psxunit\fi\space mul cntTick add def
            x abs MinValue abs le {                
	      \ifis@yAxis
	        \ifPst@maxyTicks true \else false \fi
	      \else
	        \ifPst@maxxTicks true \else false \fi
	      \fi
                { x minTickline #5 moveto
                  x maxTickline #5 lineto }
                { x tickstart STsize mul #5 moveto
                  x tickend STsize mul #5 lineto } ifelse
            } if
          } for } for
          /dsubT 0 def 
          subTwidth SLW subTColor                  
          subLSstroke
          grestore
        }                                          
        dsubT abs 0 gt {                           
          gsave                                    
          /cntsubTick cntTick dsubT add def
          subTNo 1 sub {
            cntsubTick abs MinValue abs le {       
              cntsubTick tickstart STsize mul #5 moveto
              cntsubTick tickend STsize mul #5 lineto
            }{ exit } ifelse
            /cntsubTick cntsubTick dsubT add def
          } repeat 
          subTwidth SLW subTColor                  
          subLSstroke
          grestore                                 
        } if
        showOrigin {
          gsave
          cntTick tickstart #5 moveto         	
          cntTick tickend #5 lineto    	       	
          Twidth SLW TColor                         
          LSstroke
          grestore
        }{ /showOrigin true def } ifelse         
      } for
    } if
   }
  \end@SpecialObj%
  \ifx\psk@axesstyle\@none\else
    \ifPst@yAxis\psline[linecolor=\pslinecolor](0,#2)(0,#4)\fi
    \ifPst@xAxis\psline[linecolor=\pslinecolor](#1,0)(#3,0)\fi
  \fi
  \ignorespaces
}%
\def\psxs@frame{%
  \psset{axesstyle=none}%
  \begin@SpecialObj%
    \addto@pscode{					
      \pst@number\pst@dima \psk@frameOffset sub \pst@number\pst@dimb \psk@frameOffset sub moveto 	
      \pst@number\pst@dimc \psk@frameOffset add \pst@number\pst@dimb \psk@frameOffset sub L	
      \pst@number\pst@dimc \psk@frameOffset add \pst@number\pst@dimd \psk@frameOffset add L 	
      \pst@number\pst@dima \psk@frameOffset sub \pst@number\pst@dimd \psk@frameOffset add L 	
      closepath 
      }%
    \pst@stroke%
    \psk@fillstyle%
  \end@SpecialObj%
  \let\psk@arrowA\@empty%
  \let\psk@arrowB\@empty%
  \pst@xticksizeC=\z@\pst@yticksizeC=\z@  
  \ifPst@xAxis\psxs@@axes\pst@dima\pst@dimb\pst@dimc\pst@dimd{}{x}\fi
  \ifPst@yAxis\psxs@@axes\pst@dima\pst@dimb\pst@dimc\pst@dimd{ exch }{y}\fi
  \ifnum\psk@tickstyle=\tw@	
    \psDEBUG[psxs@frame]{psk@tickstyle=2 (inner)}%
    \psDEBUG[psxs@frame]{pst@dima=\pst@number\pst@dima}%
    \psDEBUG[psxs@frame]{pst@dimb=\pst@number\pst@dimb}%
    \psDEBUG[psxs@frame]{pst@dimc=\pst@number\pst@dimc}%
    \psDEBUG[psxs@frame]{pst@dimd=\pst@number\pst@dimd}%
    \ifPst@xAxis\psxs@@axes\pst@dima\pst@dimb\pst@dimc\pst@dimd{ neg \pst@number\pst@dimd add }{x}\fi
    \ifPst@yAxis\psxs@@axes\pst@dima\pst@dimb\pst@dimc\pst@dimd{ neg \pst@number\pst@dimc add exch }{y}\fi
  \fi%
}
\def\psxs@polar{
  \pst@killglue
  \begingroup
  \edef\pst@dimC{\strip@pt\pst@dimc}
  \pstFPDiv\pstR@dius{\pst@dimC}{\strip@pt\psxunit}
  \edef\pst@dimD{\strip@pt\pst@dimd}
  \pstFPDiv\psk@EndAngle{\pst@dimD}{\strip@pt\psyunit}
  \ifnum\psk@EndAngle=0 \def\psk@EndAngle{360}\fi
  \use@keep@par
  \pstFPDiv\pstN@lpha{\psk@EndAngle}{\psk@Dy}
  \pstFPdiv\pstd@lpha{\psk@Dy}{\psk@ysubticks}
  \pstFPdiv\pstdR@dius{1}{\psk@xsubticks}
  \pst@cntm=\psk@xsubticks\advance\pst@cntm by \m@ne
  \multido{\iA=\psk@Dx+\psk@Dx,\rB=\pstdR@dius+\psk@Dx,\iB=0+1}{\pstR@dius}{%
    \multido{\rA=\rB+\pstdR@dius}{\the\pst@cntm}{%
      \psarc[linestyle=\psxsubticklinestyle,
         linecolor=\psk@xsubtickcolor,linewidth=\psk@xsubtickwidth pt](0,0){\rA}{0}{\psk@EndAngle}}    
    \psarc[linestyle=\psxticklinestyle,linecolor=\psk@xtickcolor,
		linewidth=\psk@xtickwidth pt](0,0){\iA}{0}{\psk@EndAngle}%
    \ifnum\psk@labels<2\relax
      \uput[-45](\iB,0){\pshlabel{\iB}}\uput[45](0,\iB){\pshlabel{\iB}}%
    \fi%
  }%
  \pst@cntm=\psk@ysubticks\advance\pst@cntm by \m@ne
  \multido{\iA=\psk@Oy+\psk@Dy,\rB=\pstd@lpha+\psk@Dy}{\pstN@lpha}{%
    \multido{\rA=\rB+\pstd@lpha}{\the\pst@cntm}{\psline[linestyle=\psysubticklinestyle,
      linecolor=\psk@ysubtickcolor,linewidth=\psk@ysubtickwidth pt](\pstR@dius;\rA)} 
    \psline[linestyle=\psyticklinestyle,
      linecolor=\psk@ytickcolor,linewidth=\psk@ytickwidth pt](\pstR@dius;\iA)%
    \ifodd\psk@labels\else
      \uput[\iA](\pstR@dius;\iA){\psvlabel{\iA\psk@ylabelFactor}}%
    \fi%
  }%
  \ifnum\psk@EndAngle<360 \psline[linestyle=\psyticklinestyle,
      linecolor=\psk@ytickcolor,linewidth=\psk@ytickwidth pt](\pstR@dius;0)\fi
  \endgroup\ignorespaces%
  \Pst@xAxisfalse\Pst@yAxisfalse%
}
\def\@polar{polar}
\define@key[psset]{pst-plot}{axesstyle}[axes]{%
  \@ifundefined{psxs@#1}%
    {\@pstrickserr{Axes style `#1' not defined}\@eha}%
    {\def\psk@axesstyle{#1}%
     \ifx\psk@axesstyle\@polar\psset{Dy=30}\fi}}
\psset[pst-plot]{axesstyle=axes}
\define@key[psset]{pst-plot}{xLabels}[]{\def\psk@xLabels{#1}}
\define@key[psset]{pst-plot}{xLabelsRot}[0]{\pst@getangle{#1}\pst@xLabelsRot}
\psset[pst-plot]{xLabels=,xLabelsRot=0}
\define@key[psset]{pst-plot}{yLabels}[]{\def\psk@yLabels{#1}}
\define@key[psset]{pst-plot}{yLabelsRot}[0]{\pst@getangle{#1}\pst@yLabelsRot}
\psset[pst-plot]{yLabels=,yLabelsRot=0}
\def\pst@hlabels#1#2#3#4{%
  \ifSpecialLabelsDone
  \else
    \kern\psk@xlabelOffset pt            
    \ifx\empty\psk@xLabels
      \ifdim#1=\z@
      \else                   
        \ifx#2\empty
        \else
          \advance#1\ifdim#1>\z@-\fi7\pslinewidth
        \fi
        \pst@cnta=#1\relax                
        \divide\pst@cnta\psk@dx\relax     
        \ifnum\pst@cnta=\z@
        \else
          \pst@dimb=\psk@dx sp            
          \ifnum\psk@labels<\tw@ \ifPst@xAxis\pst@@hlabels\fi\fi
          \showoriginfalse
        \fi
      \fi
   \else
     \ifnum\psk@xlabelPos=\tw@ \def\pst@tempC{90}\else\def\pst@tempC{-90}\fi
       \pstFPsub\pst@pmtempa{#4}{#3}%
       \pstFPDiv\pst@pmtempb{\pst@pmtempa}{\psk@Dx}%
       \pstFPadd\pst@pmtempc{\pst@pmtempb}{-1}%
       \pstFPadd\pst@pmtempd{\pst@pmtempb}{1}%
       \ifdim\pst@pmtempb pt < \z@ 
         \def\pst@pmtempe{\pst@int{\pst@pmtempc}}%
       \else
         \def\pst@pmtempe{\pst@int{\pst@pmtempd}}%
       \fi
       \multido{\nA=0+1,\rA=#3+\psk@Dx}{\pst@pmtempe}{%
         \ifdim \nA pt < \z@ \def\nB{-\nA} \else \def\nB{\nA} \fi
         \uput{\psxlabelsep}[\pst@tempC]{\pst@xLabelsRot}(\rA,0){%
              \strut\expandafter\pshlabel\expandafter{\psPutXLabel{\nB}}}}%
       \SpecialLabelsDonetrue
    \fi
  \fi
}
\def\pst@@hlabels{%
  \setbox\z@=\vbox{
    \ifcase\psk@xlabelPos
      \vskip-\pst@xticksizeA\vskip\psxlabelsep\or
      \vskip-1ex\vskip-\pslabelsep\or
      \vskip-\pst@xticksizeB\vskip-\psxlabelsep\vskip-1ex
    \fi
    \ifnum\pst@cnta<\z@ \pst@dimb=-\pst@dimb\fi
    \hbox to \z@{%
      \ifshoworigin\hbox to \z@{\hss\pst@@@hlabel{\psk@Ox}\hss}\fi
      \mmultido{\nA=\psk@Ox+\psk@Dx}{\pst@cnta}{%
        \hskip\pst@dimb \hbox to \z@{\hss
          \ifdim\nA pt=\z@\relax\ifshoworigin\pst@@@hlabel{0}\fi
          \else\expandafter\pst@@@hlabel{\nA}%
          \fi
        \hss}%
      }\hss
    }%
  }\ht\z@\z@ \dp\z@\z@ \box\z@}
\def\pst@vlabels#1#2#3#4{%
  \ifSpecialLabelsDone\else
      \ifx\empty\psk@yLabels
        \ifdim#1=\z@\else
          \ifx#2\empty\else\ifdim#1>\z@ \advance#1 by -7\pslinewidth\else\advance#1 by 7\pslinewidth\fi\fi
          \pst@cnta=#1\relax           
          \divide\pst@cnta\psk@dy\relax
          \ifnum\pst@cnta=\z@\else
            \pst@dima=\psk@dy sp
            \ifodd\number\psk@labels\else\ifPst@yAxis\pst@@vlabels\fi
          \fi
          \showoriginfalse
        \fi
      \fi
    \else
	\pstFPsub\pst@pmtempa{#4}{#3}%
	\pstFPDiv\pst@pmtempb{\pst@pmtempa}{\psk@Dy}%
	\pstFPadd\pst@pmtempc{\pst@pmtempb}{-1}%
	\pstFPadd\pst@pmtempd{\pst@pmtempb}{1}%
	\ifdim\pst@pmtempb pt < \z@ \def\pst@pmtempe{\pst@int{\pst@pmtempc}}\else\def\pst@pmtempe{\pst@int{\pst@pmtempd}}\fi
	\multido{\nA=0+1,\rA=#3+\psk@Dy}{\pst@pmtempe}{%
	  \ifdim \nA pt < \z@ \def\nB{-\nA}\else \def\nB{\nA}\fi
	  \ifnum\psk@ylabelPos=0
            \uput{\psylabelsep}[180]{\pst@yLabelsRot}(0,\rA){%
              \strut\expandafter\psvlabel\expandafter{\psPutYLabel{\nB}}}%
          \else
            \uput{\psylabelsep}[0]{\pst@yLabelsRot}(0,\rA){%
              \strut\expandafter\psvlabel\expandafter{\psPutYLabel{\nB}}}%
          \fi
        }%
      \SpecialLabelsDonetrue
    \fi
  \fi
}
\def\pst@@vlabels{%
  \vbox to\z@{%
   \vbox to -\psk@ylabelOffset pt{}
    \ifnum\pst@cnta>\z@ \pst@dima=-\pst@dima\fi
    \offinterlineskip
    \ifshoworigin
      \vbox to \z@{\vss\hbox to\z@{%
        \ifcase\psk@ylabelPos
	  \hss\pst@@@vlabel{\psk@Oy}\hskip\psylabelsep\hskip-\pst@yticksizeA\or%
	  \hskip\pslabelsep\hss\pst@@@vlabel{\psk@Oy}\hss\or
	  \hskip\pst@yticksizeB\hskip\psylabelsep\pst@@@vlabel{\psk@Oy}%
	\fi}\vss}%
    \fi
    \mmultido{\nA=\psk@Oy+\psk@Dy}{\pst@cnta}{%
      \vbox to\pst@dima{\vss}%
      \vbox to \z@{%
        \vss\hbox to\z@{%
        \ifcase\psk@ylabelPos 
	  \hss\ifdim\nA pt=\z@ \ifshoworigin\pst@@@vlabel{0}\fi\else\pst@@@vlabel{\nA}\fi
	    \hskip\psylabelsep\hskip-\pst@yticksizeA\or
	  \hss\ifdim\nA pt=\z@ \ifshoworigin\pst@@@vlabel{0}\fi\else\pst@@@vlabel{\nA}\fi
	  \ifdim\psylabelsep=\z@\hss\else\kern-\psylabelsep\fi\or
	  \hskip\pst@yticksizeB\hskip\psylabelsep
	  \ifdim\nA pt=\z@ \ifshoworigin\pst@@@vlabel{0}\fi\else\pst@@@vlabel{\nA}\fi
	\fi}\vss}%
    }\vss}%
}
\define@key[psset]{pst-plot}{xAxisLabel}[x]{\def\psk@xAxisLabel{#1}}
\define@key[psset]{pst-plot}{yAxisLabel}[y]{\def\psk@yAxisLabel{#1}}
\psset[pst-plot]{xAxisLabel=x,yAxisLabel=y}
\define@key[psset]{pst-plot}{xAxisLabelPos}[{}]{\def\psk@xAxisLabelPos{#1}}
\define@key[psset]{pst-plot}{yAxisLabelPos}[{}]{\def\psk@yAxisLabelPos{#1}}
\psset[pst-plot]{yAxisLabelPos={},xAxisLabelPos={}}
\newdimen\psk@llx
\newdimen\psk@lly
\newdimen\psk@urx
\newdimen\psk@ury
\define@key[psset]{pst-plot}{llx}[\z@]{\pssetxlength\psk@llx{#1}}
\define@key[psset]{pst-plot}{lly}[\z@]{\pssetylength\psk@lly{#1}}
\define@key[psset]{pst-plot}{urx}[\z@]{\pssetxlength\psk@urx{#1}}
\define@key[psset]{pst-plot}{ury}[\z@]{\pssetylength\psk@ury{#1}}
\psset[pst-plot]{llx=\z@, lly=\z@, urx=\z@, ury=\z@}
\define@boolkey[psset]{pst-plot}[Pst@]{psgrid}[true]{}
\define@key[psset]{pst-plot}{gridpara}[{}]{\def\psk@gridpara{#1}}
\define@key[psset]{pst-plot}{gridcoor}[\relax]{\def\psk@gridcoor{#1}}
\psset[pst-plot]{psgrid=false,gridpara={gridlabels=0pt,gridcolor=red!30,subgridcolor=green!30,subgridwidth=0.5\pslinewidth,
  subgriddiv=5},gridcoor=\relax}
\define@key[psset]{pst-plot}{axespos}[bottom]{\pst@expandafter\psset@@axespos{#1}\@nil}
\def\psset@@axespos#1#2\@nil{%
  \ifx#1b\let\psk@axespos\z@\else		
    \ifx#1t\let\psk@axespos\@ne			
      \else\@pstrickserr{Bad axes position: `#1#2'}\@ehpa
  \fi\fi}
\psset[pst-plot]{axespos=b}
\newdimen\pst@xunit
\newdimen\pst@yunit
\def\pslegend{\@ifnextchar[\pslegend@i{\pslegend@i[rt]}}
\def\pslegend@i[#1]{\@ifnextchar({\pslegend@ii[#1]}{\pslegend@ii[#1](\pst@number\pslabelsep,\pst@number\pslabelsep)}}
\def\pslegend@ii[#1](#2,#3)#4{%
  \gdef\pslegend@ref{#1}%
  \xdef\pslegend@sepx{#2 }%
  \xdef\pslegend@sepy{#3 }%
  \gdef\pslegend@text{#4}}
\newpsstyle{legendstyle}{fillstyle=solid,fillcolor=white,linewidth=0.5pt}
\def\pslegend@iii[#1](#2){\rput[#1](#2){\psframebox[style=legendstyle]{%
  \footnotesize\tabcolsep=2pt%
  \tabular[t]{@{}ll@{}}\pslegend@text\endtabular}}\global\let\pslegend@text\relax}
\let\pslegend@text\relax
\def\psgraph{\pst@object{psgraph}}
\def\psgraph@i{%
  \let\psgraph@para\pst@par
  \let\psk@save@arrowA\psk@arrowA
  \let\psk@save@arrowB\psk@arrowB
  \pst@getarrows\psgraph@ii}
\def\psgraph@ii(#1,#2){\catcode`\!=12\relax
  \@ifnextchar({\psgraph@iii(#1,#2)}{\psgraph@iv(0,0)(#1,#2)}}
\def\psgraph@iii(#1,#2)(#3,#4){\@ifnextchar({\psgraph@v(#1,#2)(#3,#4)}{\psgraph@iv(#1,#2)(#3,#4)}}
\def\psgraph@iv(#1,#2)(#3,#4)#5#6{
  \pst@killglue%
  \begingroup
  \use@keep@par
  \pstFPsub\pst@tempA{#3}{#1}%
  \pst@dimm=#5
  \pst@dimo=\pst@tempA pt
  \pstFPdiv\pst@@dx{\strip@pt\pst@dimm}{\pst@tempA}%
  \pst@xunit=\pst@@dx\p@
  \ifx!#6\let\pst@yunit=\pst@xunit\else
    \pst@dimm=#6
    \pstFPsub\pst@tempA{#4}{#2}%
    \pstFPdiv\pst@@dy{\strip@pt\pst@dimm}{\pst@tempA}%
    \pst@yunit=\pst@@dy\p@
  \fi
  \pst@dimm=#1\pst@xunit\advance\pst@dimm by \psk@llx
  \pst@dimn=#2\pst@yunit\advance\pst@dimn by \psk@lly
  \pst@dimo=#3\pst@xunit\advance\pst@dimo by \psk@urx
  \pst@dimp=#4\pst@yunit\advance\pst@dimp by \psk@ury
  \if@star\pspicture*(\pst@dimm,\pst@dimn)(\pst@dimo,\pst@dimp)\else
  \pspicture(\pst@dimm,\pst@dimn)(\pst@dimo,\pst@dimp)\fi
  \let\psxunit\pst@xunit \let\psyunit\pst@yunit
  \ifdim\pst@xunit=\pst@yunit\relax\psset{runit=\pst@xunit}\fi%
  \bgroup
    \use@par
  \ifPst@psgrid
     \expandafter\psset\expandafter{\psk@gridpara}%
      \rput[lb](0,0){\expandafter\psgrid\psk@gridcoor}  
  \fi
    \ifnum\psk@axespos=0
      \expandafter\psaxes\expandafter[\psgraph@para](#1,#2)(#3,#4)%
    \else
      \xdef\psgraph@coor{(#1,#2)(#3,#4)(#5,#6)}%
    \fi
  \egroup
  \psgraph@vi(#1,#2)(#1,#2)(#3,#4)%
}
\def\psgraph@v(#1,#2)(#3,#4)(#5,#6)#7#8{
  \pst@killglue%
  \let\psgraph@para\pst@par
  \begingroup%
  \use@keep@par
  \pstFPsub\pst@tempA{#5}{#3}%
  \pst@dimm=#7%
  \pst@dimo=\pst@tempA pt%
  \pstFPdiv\pst@@dx{\strip@pt\pst@dimm}\pst@tempA%
  \pst@xunit=\pst@@dx\p@%
  \ifx!#8\let\pst@yunit=\pst@xunit\else
    \pst@dimm=#8%
    \pstFPsub\pst@tempA{#6}{#4}%
    \pstFPdiv\pst@@dy{\strip@pt\pst@dimm}\pst@tempA%
    \pst@yunit=\pst@@dy\p@%
  \fi%
  \pst@dima=#3\pst@xunit \advance\pst@dima by \psk@llx%
  \pst@dimb=#4\pst@yunit \advance\pst@dimb by \psk@lly%
  \pst@dimc=#5\pst@xunit \advance\pst@dimc by \psk@urx%
  \pst@dimd=#6\pst@yunit \advance\pst@dimd by \psk@ury%
  \if@star\pspicture*(\pst@dima,\pst@dimb)(\pst@dimc,\pst@dimd)\else%
          \pspicture(\pst@dima,\pst@dimb)(\pst@dimc,\pst@dimd)\fi%
  \psset{xunit=\pst@xunit,yunit=\pst@yunit}
  \ifdim\pst@xunit=\pst@yunit \psset{runit=\pst@xunit}\fi%
  \bgroup%
    \use@par%
  \ifPst@psgrid
     \expandafter\psset\expandafter{\psk@gridpara}%
      \rput[lb](0,0){\expandafter\psgrid\psk@gridcoor}
  \fi%
    \ifnum\psk@axespos=0
      \psaxes(#1,#2)(#3,#4)(#5,#6)%
    \else
      \xdef\psgraph@coor{(#1,#2)(#3,#4)(#5,#6)}%
    \fi
  \egroup
  \psgraph@vi(#1,#2)(#3,#4)(#5,#6)%
}
\def\setxLabelC@@r#1,#2(#3,#4)(#5){%
  \pst@getcoor{#5}\pst@tempB%
  \ifx c#1 
    \pssetylength\pst@dimm{#2}%
    \rput(! #4 #3 add 2 div \pst@number\pst@dimm \pst@tempB\space exch pop add 
      \pst@number\psyunit div ){\psk@xAxisLabel}%
  \else%
    \pst@getcoor{\psk@xAxisLabelPos}\pst@tempA%
    \rput(! \pst@tempA\space \pst@tempB\space exch pop add \tx@UserCoor ){\psk@xAxisLabel}%
  \fi}
\def\setyLabelC@@r#1,#2(#3,#4)(#5){%
  \pst@getcoor{#5}\pst@tempB%
  \ifx c#2
    \pssetxlength\pst@dimm{#1}%
    \rput{90}(! \pst@number\pst@dimm \pst@tempB\space pop add \pst@number\psxunit div #4 #3 add 2 div ){\psk@yAxisLabel}%
  \else%
    \pst@getcoor{\psk@yAxisLabelPos}\pst@tempA%
    \rput{90}(! \pst@tempB\space pop \pst@tempA\space 3 1 roll add exch \tx@UserCoor ){\psk@yAxisLabel}%
  \fi}
\def\psgraph@vi(#1,#2)(#3,#4)(#5,#6){%
  \ifx\psk@xAxisLabel\@empty\else%
    \ifx\psk@xAxisLabelPos\@empty\uput[0](#5,#2){\psk@xAxisLabel}%
    \else\expandafter\setxLabelC@@r\psk@xAxisLabelPos(#3,#5)(#1,#2)\fi%
  \fi%
  \ifx\psk@yAxisLabel\@empty\else%
    \ifx\psk@yAxisLabelPos\@empty\uput[90](#1,#6){\psk@yAxisLabel}%
    \else\expandafter\setyLabelC@@r\psk@yAxisLabelPos(#4,#6)(#1,#2)\fi%
  \fi%
  \def\lt@@{lt}\def\lb@@{lb}\def\rb@@{rb}%
  \ifx\pslegend@ref\lb@@    \gdef\pslegend@coor{#3 \pslegend@sepx \pst@number\psxunit div add 
                                                   \pslegend@sepy \pst@number\psyunit div}%
  \else%
    \ifx\pslegend@ref\lt@@  \gdef\pslegend@coor{#3 \pslegend@sepx \pst@number\psxunit div add 
                                                #6 \pslegend@sepy \pst@number\psyunit div sub}%
    \else%
      \ifx\pslegend@ref\rb@@\gdef\pslegend@coor{#5 \pslegend@sepx \pst@number\psxunit div sub 
                                                   \pslegend@sepy \pst@number\psyunit div}%
      \else                 \gdef\pslegend@coor{#5 \pslegend@sepx \pst@number\psxunit div sub 
                                                #6 \pslegend@sepy \pst@number\psyunit div sub}%
      \fi%
    \fi%
  \fi%
  \xdef\psgraphLLx{#3}\xdef\psgraphLLy{#4}\xdef\psgraphURx{#5}\xdef\psgraphURy{#6}%
  \global\let\psk@arrowA\psk@save@arrowA
  \global\let\psk@arrowB\psk@save@arrowB
  \ignorespaces
}
\def\endpsgraph{%
  \ifx\relax\pslegend@text\relax \else\pslegend@iii[\pslegend@ref](!\pslegend@coor)\fi
  \expandafter\psset\expandafter{\psgraph@para}%
  \ifnum\psk@axespos>0
    \expandafter\psaxes\psgraph@coor
  \fi
  \endpspicture
  \endgroup\ignorespaces}
\@namedef{psgraph*}{\psgraph*}
\@namedef{endpsgraph*}{\endpsgraph}
\def\psPutXLabel#1{%
  \global\pst@cnto=0\relax
  \global\pst@cntp=#1\relax
  \expandafter\get@Label\psk@xLabels,,\@nil%
}
\def\psPutYLabel#1{%
  \global\pst@cnto=0\relax
  \global\pst@cntp=#1\relax
  \expandafter\get@Label\psk@yLabels,,\@nil%
}
\def\get@Label#1,#2,#3\@nil{%
    \ifnum\the\pst@cnto<\the\pst@cntp
      \global\advance\pst@cnto by \@ne 
      \ifx\relax#3\relax\else\expandafter\get@Label#2,#3\@nil\fi%
    \else #1\fi%
}
\def\psVectorfield{\pst@object{psVectorfield}}
\def\psVectorfield@i(#1,#2)(#3,#4)#5{{%
  \addbefore@par{Dx=0.1,Dy=0.1,Ox=3,arrows=->,linewidth=0.2pt}%
  \begin@SpecialObj
  \SpecialCoor
  \pstFPsub\pst@tempA{#3}{#1}%
  \pstFPsub\pst@tempB{#4}{#2}%
  \pstFPDiv{\pst@tempC}{\pst@tempA}{\psk@Dx}%
  \pstFPDiv{\pst@tempD}{\pst@tempB}{\psk@Dy}%
  \pstVerb{ /yStrich \ifPst@algebraic (#5) tx@AlgToPs begin AlgToPs end cvx
                \else { #5 } \fi def }%
  \multido{\rX=#1+\psk@Dx}{\numexpr\pst@tempC+1}{%
    \multido{\rY=#2+\psk@Dy}{\numexpr\pst@tempD+1}{%
       \psline%
         (! /x \rX\space def 
            /y \rY\space def 
            /yTemp yStrich \psk@Dx\space \psk@Ox\space div mul def 
            \rX\space \psk@Dx\space \psk@Ox\space div sub \rY\space yTemp sub)%
         (! /x \rX\space def 
            /y \rY\space def 
            /yTemp yStrich \psk@Dx\space \psk@Ox\space div mul def 
            \rX\space \psk@Dx\space \psk@Ox\space div add \rY\space yTemp add)%
   }}%
  \end@SpecialObj
}\ignorespaces}  
\def\psFixpoint{\pst@object{psFixpoint}}
\def\psFixpoint@i#1#2#3{
  \pst@killglue%
  \begingroup%
  \use@par%
  \@nameuse{beginplot@\psplotstyle}%
  \addto@pscode{
    \psplot@init
      /x #1 def
      /F@pstplot \ifPst@algebraic (#2) tx@AlgToPs begin AlgToPs end cvx
                 \else { #2 } \fi  def
      /xy { x \pst@number\psxunit mul F@pstplot dup /x ED \pst@number\psyunit mul } def 
  }%
  \gdef\psplot@init{}%
  \@pstfalse%
  \@nameuse{testqp@\psplotstyle}%
  \addto@pscode{
      mark
      x \pst@number\psxunit mul 0
      /n 2 def
      #3 {
        xy 
        dup dup 
        /n n 4 add def
      } repeat 
  }%
  \@nameuse{endplot@\psplotstyle}%
  \endgroup%
  \ignorespaces}
\define@boolkey[psset]{pst-plot}[Pst@]{showDerivation}[true]{}
\psset{showDerivation}
\def\psNewton{\pst@object{psNewton}}
\def\psNewton@i#1#2{\@ifnextchar[{\psNewton@ii{#1}{#2}}{\psNewton@iii{#1}{#2}}}
\def\psNewton@ii#1#2[#3]#4{
  \pst@killglue%
  \begingroup%
  \addbefore@par{showDerivation}%
  \use@par%
  \@nameuse{beginplot@\psplotstyle}%
  \addto@pscode{
    \psplot@init
      /x #1 def
      /F@pstplot \ifPst@algebraic (#2) tx@AlgToPs begin AlgToPs end cvx \else { #2 } \fi  def
      /F@pstplotDerive \ifPst@algebraic (#3) tx@AlgToPs begin AlgToPs end cvx \else { #3 } \fi  def
      /newxVal { 
        F@pstplotDerive 
        div neg 
      } def
  }%
  \gdef\psplot@init{}%
  \@pstfalse%
  \@nameuse{testqp@\psplotstyle}%
  \addto@pscode{
      mark
      x 0 \tx@ScreenCoor 
      /n 2 def
      #4 {
        F@pstplot /yVal ED
        x yVal \tx@ScreenCoor
        /n n 2 add def
        yVal newxVal x add /x ED
        x 0 \tx@ScreenCoor 
        \ifPst@showDerivation /n n 4 add def \else moveto /n n 2 add def\fi
      } repeat 
      pstack
  }%
  \@nameuse{endplot@\psplotstyle}%
  \endgroup%
  \ignorespaces}
\def\psNewton@iii#1#2#3{
  \pst@killglue%
  \begingroup%
  \addbefore@par{VarStepEpsilon=0.01,showDerivation}%
  \use@par%
  \@nameuse{beginplot@\psplotstyle}%
  \addto@pscode{
    \psplot@init
      /epsilon \psk@VarStepEpsilon\space def
      /x #1 def
      /F@pstplot \ifPst@algebraic (#2) tx@AlgToPs begin AlgToPs end cvx \else { #2 } \fi  def
      /newxVal { 
        /saveX x def
        saveX epsilon add /x ED F@pstplot saveX epsilon sub /x ED F@pstplot sub epsilon dup add div 
        div neg 
        /x saveX def
      } def
  }%
  \gdef\psplot@init{}%
  \@pstfalse%
  \@nameuse{testqp@\psplotstyle}%
  \addto@pscode{
      mark
      x 0 \tx@ScreenCoor 
      /n 2 def
      #3 {
        F@pstplot /yVal ED
        x yVal \tx@ScreenCoor
        yVal newxVal x add /x ED
        x 0 \tx@ScreenCoor 
        \ifPst@showDerivation /n n 4 add def \else moveto /n n 2 add def\fi
      } repeat 
  }%
  \@nameuse{endplot@\psplotstyle}%
  \endgroup%
  \ignorespaces}
\def\psResetPlotValues{%
  \psset{method={}}%
}%
\catcode`\@=\TheAtCode\relax
 
\csname PSTnodesLoaded\endcsname
\let\PSTnodesLoaded 
\ifx\PSTricksLoaded \else\input pstricks.tex \fi\relax
\ifx\PSTXKeyLoaded \else \input pst-xkey \fi
\def\fileversion{1.42}
\def\filedate{2019/03/03}
\edef\TheAtCode{\the\catcode`\@}
\catcode`\@=11
\pstheader{pst-node.pro}
\pst@addfams{pst-node}
\SpecialCoor
\define@boolkey[psset]{pst-node}[Pst@]{trueAngle}[true]{}
\psset[pst-node]{trueAngle=false}
\define@boolkey[psset]{pst-node}[Pst@]{storeNodeInfo}[true]{}
\psset[pst-node]{storeNodeInfo=false}
\def\pst@nodedict{tx@NodeDict begin }
\def\pst@zapspace#1 #2{%
#1%
\ifx#2\@empty\else\expandafter\pst@zapspace\fi
#2}
\def\pst@getnode#1#2{\pst@expandafter\pst@@getnode{#1},,\@nil#2}
\def\pst@@getnode#1,#2,#3\@nil#4{%
  \ifx\@empty#3\@empty
    \edef#4{/N@\pst@zapspace#1 \@empty\space}%
  \else
    \pst@cntg=#1\relax
    \pst@cnth=#2\relax
    \edef#4{/N@M-\ifnum\psmatrixcnt=\z@ 1\else\the\psmatrixcnt\fi
    -\the\pst@cntg-\the\pst@cnth\space}%
  \fi}
\def\tx@NewNode{/NodeScale {\ifx\pstnodescale\@undefined  \else\pstnodescale \fi} def NewNode }
\def\psopenNodeFile{%
  \pst@Verb{ 
    (\jobname.nodes) (w) file /NodeFile exch def 
  }}
\def\pscloseNodeFile{\pstVerb{ tx@NodeDict begin NodeFile closefile end }}
\define@boolkey[psset]{pst-node}[Pst@]{showNode}[true]{\ifPst@showNode\psopenNodeFile\fi}
\define@boolkey[psset]{pst-node}[Pst@]{markNode}[true]{}
\define@boolkey[psset]{pst-node}[Pst@]{saveNodeCoors}[true]{}
\define@key[psset]{pst-node}{NodeCoorPrefix}[]{\def\psk@NodeCoorPrefix{#1}}
\psset[pst-node]{saveNodeCoors=false,showNode=false,markNode=false,NodeCoorPrefix=}
\def\pst@newnode#1#2#3#4{%
\pst@killglue
\leavevmode
\pst@getnode{#1}\pst@thenode
\pst@Verb{
  \ifPst@saveNodeCoors
    \ifx\relax#3\relax 0 0 \else gsave \pst@dict STV CP T end #3 \tx@UserCoor grestore \fi 
    \if$\psk@NodeCoorPrefix$
      /N-#1.y exch def
      /N-#1.x exch def
    \else
      /\psk@NodeCoorPrefix#1y exch def
      /\psk@NodeCoorPrefix#1x exch def
    \fi
  \fi
  \pst@nodedict
  {#3}
  \ifx\psk@name\relax false \else \psk@name true \fi
  \pst@thenode
  #2
  {#4}
  \ifPst@showNode 
  exch dup /NodeType ED 
  exch
   NodeType 10 eq {  
    5 copy 
    cvlit aload pop
    20 string cvs (; )   6 2 roll 
    20 string cvs (; )   7 2 roll 
    20 string cvs (; )   8 2 roll 
    20 string cvs (; )   9 2 roll 
    cvlit dup length 2 eq 
      { aload pop exch 
        20 string cvs (; ) 11 2 roll 
        20 string cvs (, ) 12 2 roll  
        (\string\n)                   
        13 array astore concatstringarray 
      }
      { 255 string cvs (; ) 10 2 roll 
        (\string\n)                   
        11 array astore concatstringarray 
      } ifelse 
    NodeFile exch writestring 
  } if
  NodeType 14 eq {  
    5 copy 
    /@@temp ED 
    @@temp  
    4 -1 roll cvlit pop
    ( OvalNodePos ) (; )  5 2 roll
    20 string cvs (; )   6 2 roll 
    20 string cvs (; )   7 2 roll 
    20 string cvs (; )   8 2 roll 
    Y 20 string cvs (; ) 10 2 roll
    X 20 string cvs (, ) 12 2 roll
    (\string\n)                   
    13 array astore concatstringarray 
    tx@NodeDict begin NodeFile exch writestring end
  } if
  \fi
  \tx@NewNode
  end 
}%
\global\let\psk@name\relax%
\pstree@nodehook%
\global\let\pstree@nodehook\relax}
\let\pstree@nodehook\relax
\define@boolkey[psset]{pst-node}[Pst@]{nodealign}[true]{}
\psset[pst-node]{nodealign=false}

\def\pst@nodealign{%
\pst@dimg=\ht\pst@hbox
\advance\pst@dimg by -\dp\pst@hbox
\divide\pst@dimg by \tw@
\lower\pst@dimg}
\def\tx@InitPnode{InitPnode }
\def\pnode{\@ifnextchar[{\pnode@i}{\pnode@iii}}
\def\pnode@i[#1]{\@ifnextchar({\pnode@ii[#1]}{\pnode@ii[#1](0,0)}}
\def\pnode@ii[#1](#2)#3{%
  \pst@getcoor{#1}\pst@tempA%
  \pst@getcoor{#2}\pst@tempB%
  \pst@newnode{#3}{10}{\pst@tempA \pst@tempB 3 -1 roll add 3 1 roll add exch }{\tx@InitPnode}%
  \ifPst@showNode\psdot(#3)\uput[\ifx\psk@rot\@empty0\else\psk@rot\fi]{0}(#3){#3}\fi
  \ignorespaces}
\def\pnode@iii{\@ifnextchar({\pnode@}{\pnode@(0,0)}}
\def\pnode@(#1)#2{%
  \pst@@getcoor{#1}%
  \pst@newnode{#2}{10}{\pst@coor}{\tx@InitPnode}%
  \ifPst@showNode\psdot(#2)\uput[\ifx\psk@rot\@empty0\else\psk@rot\fi]{0}(#2){#2}\fi
  \ignorespaces}
\def\pnodes{\@ifnextchar[{\pnodes@i}{\pnodes@i[0,0]}}
\def\pnodes@i[#1]{\@ifnextchar({\psnodes@ii[#1]}{\pnodes@ii}}
\def\psnodes@ii[#1](#2)#3{%
  \pnode[#1](#2){#3}%
  \@ifnextchar({\psnodes@ii[#1]}{}%
}
\def\tx@InitCnode{InitCnode }
\def\cnode{\pst@object{cnode}}
\def\cnode@i{\@ifnextchar({\cnode@ii}{\cnode@ii(0,0)}}
\def\cnode@ii(#1)#2#3{%
  \leavevmode
  \hbox{%
    \use@par
    \pst@@getcoor{#1}%
    \pssetlength\pst@dimc{#2}%
    \pst@dimg=\psk@dimen\pslinewidth
    \advance\pst@dimc-\pst@dimg
    \advance\pst@dimc.5\pslinewidth
    \ifPst@nodealign
      \kern\pst@dimc
      \vrule width\z@ height \pst@dimc depth \pst@dimc
    \fi
    \pscircle@do(#1){#2}%
    \pst@newnode{#3}{11}{\pst@coor \pst@number\pst@dimc}{\tx@InitCnode}%
    \ifPst@nodealign\kern\pst@dimc\fi%
  }%
  \ignorespaces}
\def\Cnode{\pst@object{Cnode}}
\def\Cnode@i{\@ifnextchar({\Cnode@ii}{\Cnode@ii(0,0)}}
\def\Cnode@ii(#1)#2{\cnode@ii(#1){\psk@radius}{#2}}%
\def\cnodeput{\pst@object{cnodeput}}
\def\cnodeput@i{\@ifnextchar({\cnodeput@iii}{\cnodeput@ii}}
\def\cnodeput@ii#1{%
  \addto@par{rot={#1}}%
  \@ifnextchar({\cnodeput@iii}{\cnodeput@iii(\z@,\z@)}%
}
\def\cnodeput@iii(#1)#2{%
  \pst@killglue
  \@fixedradiusfalse
  \def\pst@nodehook{\cnodeput@iv{#2}}%
  \pst@makebox{\cput@v{#1}}%
}
\def\cnodeput@iv#1{%
  \pst@newnode{#1}{11}{\pscirclebox@iv \pst@number\pslinewidth add}{\tx@InitCnode}%
  \global\let\pst@nodehook\relax
  \ignorespaces
}
\def\Cnodeput{\pst@object{Cnodeput}}
\def\Cnodeput@i{\@ifnextchar({\Cnodeput@iii}{\Cnodeput@ii}}
\def\Cnodeput@ii#1{%
  \addto@par{rot={#1}}%
  \@ifnextchar({\Cnodeput@iii}{\Cnodeput@iii(\z@,\z@)}}
\def\Cnodeput@iii(#1)#2{%
  \pst@killglue
  \@fixedradiustrue
  \def\pst@nodehook{\Cnodeput@iv{#2}}%
  \pst@makebox{\cput@v{#1}}%
}
\def\Cnodeput@iv#1{%
  \pst@newnode{#1}{11}{%
    \pst@number{\wd\pst@hbox} 2 div \pst@number\pst@dima 
    \pst@number\pst@dimb \pst@number\pslinewidth \psk@dimen .5 sub mul sub }
       {\tx@InitCnode}%
  \global\let\pst@nodehook\relax}
\def\circlenode{\pst@object{circlenode}}
\def\circlenode@i#1{\pst@makebox{\circlenode@ii{#1}}}
\def\circlenode@ii#1{%
  \begingroup
  \pst@useboxpar
  \setbox\pst@hbox=\hbox{%
    \cnodeput@iv{#1}%
    \pscirclebox@iii
    \box\pst@hbox}%
  \ifPst@nodealign \psboxseptrue \fi
  \ifpsboxsep \pscirclebox@sep \fi
  \leavevmode
  \ifPst@nodealign\pst@nodealign\fi
  \box\pst@hbox
  \endgroup}
\def\Circlenode{\pst@object{Circlenode}}
\def\Circlenode@i#1{\pst@makebox{\Circlenode@ii{#1}}}
\def\Circlenode@ii#1{%
\begingroup
  \pst@useboxpar
  \pst@dima=\ht\pst@hbox
  \advance\pst@dima by -\dp\pst@hbox
  \divide\pst@dima by \tw@
  \pssetlength\pst@dimb\psk@radius
  \setbox\pst@hbox=\hbox{%
  \Cnodeput@iv{#1}%
  \pscircle(.5\wd\pst@hbox,\pst@dima){\pst@dimb}%
  \box\pst@hbox}%
  \ifPst@nodealign \psboxseptrue \fi
  \ifpsboxsep \psCirclebox@sep \fi
  \leavevmode
  \ifPst@nodealign\pst@nodealign\fi
  \box\pst@hbox
  \endgroup}
\def\tx@GetRnodePos{GetRnodePos }
\def\tx@InitRnode{InitRnode }
\def\psnode{\pst@object{psnode}}
\def\psnode@i{\@ifnextchar(\psnode@ii{\psnode@ii(0,0)}}
\def\psnode@ii(#1)#2#3{
  \rput(#1){\rnode{#2}{#3}}}
\def\rnode{\@ifnextchar[{\rnode@i}{\def\pst@par{}\rnode@ii}}
\def\rnode@i[#1]{\def\pst@par{ref=#1}\rnode@ii}
\def\rnode@ii#1{\pst@makebox{\rnode@iii\rnode@iv{#1}}}
\def\rnode@iii#1#2{%
\leavevmode
\begingroup
\pst@useboxpar
#1%
\ifPst@nodealign\lower\pst@dimb\fi
\hbox{%
\pst@newnode{#2}{16}{%
\pst@number{\ht\pst@hbox}%
\pst@number{\dp\pst@hbox}%
\pst@number{\wd\pst@hbox}%
\pst@number\pst@dima%
\pst@number\pst@dimb}%
{\tx@InitRnode}%
\box\pst@hbox}%
\endgroup}
\def\rnode@iv{%
\pst@dima=\psk@xref\wd\pst@hbox
\ifx\psk@yref\relax
\pst@dimb=\z@
\else
\pst@dimb=\ht\pst@hbox
\advance\pst@dimb\dp\pst@hbox
\pst@dimb=\psk@yref\pst@dimb
\advance\pst@dimb-\dp\pst@hbox
\fi}
\define@key[psset]{pst-node}{href}[0]{\pst@checknum{#1}\psk@href}
\psset[pst-node]{href=0}
\define@key[psset]{pst-node}{vref}[0.7ex]{\def\psk@vref{#1}}
\psset[pst-node]{vref=0.7ex}
\def\Rnode{\pst@object{Rnode}}
\def\Rnode@i#1{\pst@makebox{\rnode@iii\Rnode@ii{#1}}}
\def\Rnode@ii{%
\use@par
\pst@dima=\psk@href\wd\pst@hbox
\advance\pst@dima\wd\pst@hbox
\divide\pst@dima 2
\pssetlength\pst@dimb{\psk@vref}}
\def\tx@DiaNodePos{DiaNodePos }
\def\dianode{\pst@object{dianode}}
\def\dianode@i#1{\pst@makebox{\dianode@ii{#1}}}
\def\dianode@ii#1{%
\begingroup
\pst@useboxpar
\psdiabox@iii
\setbox\pst@hbox=\hbox{%
\pst@newnode{#1}{14}{}{%
/X \pst@number\pst@dima def
/Y \pst@number\pst@dimb def
/w \pst@number\pst@dimc 2 mul def
/h \pst@number\pst@dimd 2 mul def
/NodePos { \tx@DiaNodePos } def}%
\box\pst@hbox}%
\ifPst@nodealign\psboxseptrue\fi
\ifpsboxsep\psdiabox@sep\fi
\leavevmode
\ifPst@nodealign\lower\pst@dimb\fi
\box\pst@hbox
\endgroup}
\def\tx@TriNodePos{TriNodePos }
\def\tx@InitTriNode{InitTriNode }
\def\trinode{\pst@object{trinode}}
\def\trinode@i#1{\pst@makebox{\trinode@ii{#1}}}
\def\trinode@ii#1{%
  \begingroup%
  \pst@useboxpar%
  \pstribox@iii
  \setbox\pst@hbox=\hbox{%
    \pst@newnode{#1}{14}{}{
      \pst@number\pst@dimc
      \pst@number\pst@dimd
      \ifodd\psk@trimode
        exch
        \pst@number\pst@dima
      \else
        \pst@number\pst@dimb
      \fi
      \psk@trimode
      \pst@number{\wd\pst@hbox}
      \pst@number{\ht\pst@hbox}
      \pst@number{\dp\pst@hbox}
      \tx@InitTriNode
    }%
    \box\pst@hbox%
  }%
  \ifPst@nodealign\psboxseptrue\fi
  \ifpsboxsep\pstribox@sep\fi
  \leavevmode
  \ifPst@nodealign\lower\pst@tempa\fi
  \box\pst@hbox%
  \endgroup}
\def\tx@OvalNodePos{OvalNodePos }
\def\ovalnode{\pst@object{ovalnode}}
\def\ovalnode@i#1{\pst@makebox{\ovalnode@ii{#1}}}
\def\ovalnode@ii#1{%
\begingroup
\pst@useboxpar
\psovalbox@iii
\setbox\pst@hbox=\hbox{%
\pst@newnode{#1}{14}{}{%
/X \pst@number\pst@dima def
/Y \pst@number\pst@dimb def
/w \pst@number\pst@dimc def
/h \pst@number\pst@dimd def
/NodePos { \tx@OvalNodePos } def}%
\unhbox\pst@hbox}%
\ifPst@nodealign\psboxseptrue\fi
\ifpsboxsep\psovalbox@sep\fi
\leavevmode
\ifPst@nodealign\lower\pst@dimb\fi
\box\pst@hbox
\endgroup}
\def\dotnode{\pst@object{dotnode}}
\def\dotnode@i{\@ifnextchar({\dotnode@ii}{\dotnode@ii(\z@,\z@)}}
\def\dotnode@ii(#1)#2{%
  \leavevmode
  \hbox{%
    \use@par
    \pst@@getcoor{#1}%
    \pst@getdotsize
    \pstree@nodehook
    \ifPst@nodealign
      \pst@dima=\pst@dimg
      \kern\pst@dima
      \vrule width\z@ height \pst@dimh depth \pst@dimh
    \fi
    \pst@newnode{#2}{14}{}{
      \pst@coor
      /Y exch def /X exch def
      /w \pst@number\pst@dimg def
      /h \pst@number\pst@dimh def
      /NodePos { \tx@OvalNodePos } def}%
    \psdot@ii(#1)%
    \ifPst@nodealign\kern\pst@dima\fi}%
  \ifPst@markNode\uput[\ifx\psk@rot\@empty0\else\psk@rot\fi]{0}(#2){#2}\fi
  \ignorespaces}
\def\dotnodes{\pst@object{dotnodes}}
\def\dotnodes@i{\use@par\dotnodes@ii}
\def\dotnodes@ii(#1)#2{%
  \dotnode(#1){#2}%
  \@ifnextchar(\dotnodes@ii{\def\pst@par{}}}
\define@key[psset]{pst-node}{framesize}{\pst@expandafter\psset@@framesize{#1} \@nil}
\def\psset@@framesize#1 #2\@nil{%
  \pssetlength\pst@dimg{#1}%
  \divide\pst@dimg2
  \edef\psk@framewidth{\pst@number\pst@dimg}%
  \ifx\@empty#2\@empty
    \let\psk@frameheight\psk@framewidth
  \else
    \pssetlength\pst@dimg{#2}%
    \divide\pst@dimg2
    \edef\psk@frameheight{\pst@number\pst@dimg}%
  \fi}
\psset[pst-node]{framesize=10pt}
\def\fnode{\pst@object{fnode}}
\def\fnode@i{\@ifnextchar({\fnode@ii}{\fnode@ii(\z@,\z@)}}
\def\fnode@ii(#1)#2{%
  \leavevmode
  \pst@killglue
  \hbox{%
    \use@par%
    \begin@ClosedObj%
    \ifPst@nodealign
      \kern\psk@framewidth\p@
      \vrule width\z@ height \psk@frameheight\p@ depth \psk@frameheight\p@
      \edef\pst@coor{0 0 }%
    \else\pst@@getcoor{#1}\fi
    \pst@newnode{#2}{14}{}{
      \pst@coor
      /Y exch def /X exch def
      /d \psk@dimen .5 sub CLW mul neg def
      /r \psk@framewidth d add def
      /l r neg def
      /u \psk@frameheight d add def
      /d u neg def
      /NodePos { \tx@GetRnodePos } def}%
    \addto@pscode{
      /x2 \psk@framewidth CLW \psk@dimen mul sub def
      /y2 \psk@frameheight CLW \psk@dimen mul sub def
      \pst@coor 2 copy
      y2 sub /y1 ED
      x2 sub /x1 exch def
      y2 add /y2 exch def
      x2 add /x2 exch def
      \psk@cornersize
      1 index 0 eq { pop pop \tx@Rect } { \tx@OvalFrame } ifelse}%
    \def\pst@linetype{2}%
    \showpointsfalse%
    \end@ClosedObj%
    \ifPst@nodealign\kern\psk@framewidth\p@\fi}
  \ignorespaces}
\define@key[psset]{}{XnodesepA}{\pst@getlength{#1}\psk@nodesepA\def\psk@nodeseptypeA{2 }}
\define@key[psset]{}{XnodesepB}{\pst@getlength{#1}\psk@nodesepB\def\psk@nodeseptypeB{2 }}
\define@key[psset]{}{Xnodesep}{%
    \pst@getlength{#1}\psk@nodesepA
    \let\psk@nodesepB\psk@nodesepA
    \def\psk@nodeseptypeA{2 }%
    \def\psk@nodeseptypeB{2 }}
\define@key[psset]{}{YnodesepA}{\pst@getlength{#1}\psk@nodesepA\def\psk@nodeseptypeA{1 }}
\define@key[psset]{}{YnodesepB}{\pst@getlength{#1}\psk@nodesepB\def\psk@nodeseptypeB{1 }}
\define@key[psset]{}{Ynodesep}{%
    \pst@getlength{#1}\psk@nodesepA
    \let\psk@nodesepB\psk@nodesepA
    \def\psk@nodeseptypeA{1 }%
    \def\psk@nodeseptypeB{1 }}
\define@key[psset]{pst-node}{nodesepA}[0pt]{\pst@getlength{#1}\psk@nodesepA \def\psk@nodeseptypeA{0 }}
\define@key[psset]{pst-node}{nodesepB}[0pt]{\pst@getlength{#1}\psk@nodesepB \def\psk@nodeseptypeB{0 }}
\define@key[psset]{pst-node}{nodesep}[0pt]{%
  \pst@getlength{#1}\psk@nodesepA
  \let\psk@nodesepB\psk@nodesepA
  \def\psk@nodeseptypeA{0 }%
  \def\psk@nodeseptypeB{0 }}
\psset[pst-node]{nodesep=0pt}
\define@key[psset]{pst-node}{armA}[10pt]{\pst@getlength{#1}\psk@armA \def\psk@armtypeA{0 }}
\define@key[psset]{pst-node}{armB}[10pt]{\pst@getlength{#1}\psk@armB \def\psk@armtypeB{0 }}
\define@key[psset]{pst-node}{arm}[10pt]{%
  \pst@getlength{#1}\psk@armA
  \let\psk@armB\psk@armA
  \def\psk@armtypeA{0 }%
  \def\psk@armtypeB{0 }}
\psset[pst-node]{arm=10pt}
\define@key[psset]{pst-node}{XarmA}[]{\pst@getlength{#1}\psk@armA \def\psk@armtypeA{1 }}
\define@key[psset]{pst-node}{XarmB}[]{\pst@getlength{#1}\psk@armB \def\psk@armtypeB{1 }}
\define@key[psset]{pst-node}{Xarm}{%
  \pst@getlength{#1}\psk@armA
  \let\psk@armB\psk@armA
  \def\psk@armtypeA{1 }%
  \def\psk@armtypeB{1 }}
\define@key[psset]{pst-node}{YarmA}[]{\pst@getlength{#1}\psk@armA \def\psk@armtypeA{2 }}
\define@key[psset]{pst-node}{YarmB}[]{\pst@getlength{#1}\psk@armB \def\psk@armtypeB{2 }}
\define@key[psset]{pst-node}{Yarm}[]{%
  \pst@getlength{#1}\psk@armA
  \let\psk@armB\psk@armA
  \def\psk@armtypeA{2 }%
  \def\psk@armtypeB{2 }}
\define@key[psset]{pst-node}{offsetA}[0pt]{\pst@getlength{#1}\psk@offsetA}
\define@key[psset]{pst-node}{offsetB}[0pt]{\pst@getlength{#1}\psk@offsetB}
\define@key[psset]{pst-node}{offset}[0pt]{\pst@getlength{#1}\psk@offsetA\let\psk@offsetB\psk@offsetA}
\psset[pst-node]{offset=0pt}
\define@key[psset]{pst-node}{angleA}[0]{\pst@getangle{#1}\psk@angleA}
\define@key[psset]{pst-node}{angleB}[0]{\pst@getangle{#1}\psk@angleB}%
\define@key[psset]{pst-node}{angle}[0]{%
  \pst@getangle{#1}\psk@angleA
  \let\psk@angleB\psk@angleA}
\psset[pst-node]{angle=0}
\define@key[psset]{pst-node}{arcangleA}[8]{\pst@getangle{#1}\psk@arcangleA}
\define@key[psset]{pst-node}{arcangleB}[8]{\pst@getangle{#1}\psk@arcangleB}%
\define@key[psset]{pst-node}{arcangle}[8]{%
  \pst@getangle{#1}\psk@arcangleA
  \let\psk@arcangleB\psk@arcangleA}
\psset[pst-node]{arcangle=8}
\define@key[psset]{pst-node}{ncurvA}[0.67]{\pst@checknum{#1}\psk@ncurvA}
\define@key[psset]{pst-node}{ncurvB}[0.67]{\pst@checknum{#1}\psk@ncurvB}%
\define@key[psset]{pst-node}{ncurv}[0.67]{\pst@checknum{#1}\psk@ncurvA\let\psk@ncurvB\psk@ncurvA}
\psset[pst-node]{ncurv=0.67}
\def\tx@GetCenter{GetCenter }
\def\tx@XYPos{XYPos }
\def\tx@GetEdge{GetEdge }
\def\tx@AddOffset{AddOffset }
\def\tx@GetEdgeA{GetEdgeA }
\def\tx@GetEdgeB{GetEdgeB }
\def\tx@GetArmA{GetArmA }
\def\tx@GetArmB{GetArmB }
\def\check@arrow#1#2{%
  \check@@arrow#2-\@nil
  \if@pst\addto@par{arrows=#2}\def\next{#1}%
  \else\def\next{#1{#2}}\fi
  \next}
\def\check@@arrow#1-#2\@nil{%
\ifx\@nil#2\@nil\@pstfalse\else\@psttrue\fi}
\def\tx@InitNC{InitNC }
\def\nc@object#1#2#3#4#5{%
  \csname begin@#1Obj\endcsname
  \showpointsfalse
  \pst@getnode{#2}\pst@tempa
  \pst@getnode{#3}\pst@tempb
  \gdef\npos@default{#4 }%
  \addto@pscode{%
    /NCLW CLW def
    \pst@nodedict
    \psk@offsetA
    \psk@offsetB neg
    \psk@nodesepA
    \psk@nodesepB
    \psk@nodeseptypeA
    \psk@nodeseptypeB
    \pst@tempa
    \pst@tempb
    \tx@InitNC { #5 } if
    end }%
  \def\use@pscode{%
    \pst@Verb{gsave \tx@STV newpath \pst@code\space grestore}%
    \gdef\pst@code{}}%
  \csname end@#1Obj\endcsname
  \pst@shortput}
\def\npos@default{.5 }
\def\pc@object#1{%
  \@ifnextchar({\pc@@object#1}{\pst@getarrows{\pc@@object#1}}}
\def\pc@@object#1(#2)(#3){%
  \pnode(#2){@@A}\pnode(#3){@@B}%
  #1{@@A}{@@B}}
\def\tx@LPutLine{LPutLine }
\def\tx@LPutLines{LPutLines }
\def\tx@BezierMidpoint{BezierMidpoint }
\def\tx@HPosBegin{HPosBegin }
\def\tx@HPosEnd{HPosEnd }
\def\tx@HPutLine{HPutLine }
\def\tx@HPutLines{HPutLines }
\def\tx@VPosBegin{VPosBegin }
\def\tx@VPosEnd{VPosEnd }
\def\tx@VPutLine{VPutLine }
\def\tx@VPutLines{VPutLines }
\def\tx@HPutCurve{HPutCurve }
\def\tx@NCCoor{NCCoor }
\def\tx@NCLine{NCLine }
\def\ncline{\pst@object{ncline}}
\def\ncline@i{\check@arrow{\ncline@ii}}
\def\ncline@ii#1#2{\nc@object{Open}{#1}{#2}{.5}{\tx@NCLine}}
\def\pcline{\pst@object{pcline}}
\def\pcline@i{\pc@object\ncline@ii}
\def\ncLine{\pst@object{ncLine}}
\def\ncLine@i{\check@arrow{\ncLine@ii}}
\def\ncLine@ii#1#2{\nc@object{Open}{#1}{#2}{.5}%
{\tx@NCLine /LPutPos { xB yB xA yA \tx@LPutLine } def}}
\def\tx@NCLines{NCLines }
\def\nclines{\pst@object{nclines}}
\def\nclines@i{\check@arrow\nclines@ii}
\def\nclines@ii#1#2{%
\begingroup
\use@par
\def\pst@aftercoors{\nclines@iii{#1}{#2}}%
\def\pst@coors{}%
\pst@@getcoors}
\def\nclines@iii#1#2{%
\nc@object{Open}{#1}{#2}{.5}{%
tx@Dict begin \psline@iii pop end
mark \pst@coors \tx@NCLines}%
\endgroup
\ignorespaces}
\def\tx@NCCurve{NCCurve }
\def\nccurve{\pst@object{nccurve}}
\def\nccurve@i{\check@arrow{\nccurve@ii}}
\def\nccurve@ii#1#2{\nc@object{Open}{#1}{#2}{.5}{%
  /AngleA \psk@angleA\space def /AngleB \psk@angleB\space def
  \psk@ncurvB\space \psk@ncurvA\space
  \tx@NCCurve}}
\def\pccurve{\pst@object{pccurve}}
\def\pccurve@i{\pc@object\nccurve@ii}
\def\ncarc{\pst@object{ncarc}}
\def\ncarc@i{\check@arrow{\ncarc@ii}}
\def\ncarc@ii#1#2{\nc@object{Open}{#1}{#2}{.5}{%
  yB yA sub xB xA sub \tx@Atan dup
  \psk@arcangleA\space add /AngleA exch def
  \psk@arcangleB\space sub 180 add /AngleB exch def
  \psk@ncurvB\space \psk@ncurvA\space
  \tx@NCCurve}}
\def\pcarc{\pst@object{pcarc}}
\def\pcarc@i{\pc@object\ncarc@ii}
\def\tx@NCAngles{NCAngles }
\define@boolkey[psset]{pst-node}[Pst@]{pcRef}[true]{}
\psset[pst-node]{pcRef=false}
\def\ncangles{\pst@object{ncangles}}
\def\ncangles@i{\check@arrow{\ncangles@ii}}
\def\ncangles@ii#1#2{%
  \nc@object{Open}{#1}{#2}{1.5}{\ncangles@iii \tx@NCAngles}}
\def\ncangles@iii{
  tx@Dict begin \psline@iii pop end
  /AngleA \psk@angleA def
  /AngleB \psk@angleB def
  /ArmA \psk@armA \ifPst@pcRef 
    GetEdgeA yA yA1 sub dup mul xA xA1 sub dup mul add sqrt sub \fi def
  /ArmB \psk@armB def
  /ArmTypeA \psk@armtypeA def
  /ArmTypeB \psk@armtypeB def }
\def\pcangles{\pst@object{pcangles}}
\def\pcangles@i{\pc@object\ncangles@ii}
\def\tx@NCAngle{NCAngle }
\def\ncangle{\pst@object{ncangle}}
\def\ncangle@i{\check@arrow{\ncangle@ii}}
\def\ncangle@ii#1#2{%
\nc@object{Open}{#1}{#2}{1.5}{\ncangles@iii \tx@NCAngle}}
\def\pcangle{\pst@object{pcangle}}
\def\pcangle@i{\pc@object\ncangle@ii}
\def\tx@NCBar{NCBar }
\def\ncbar{\pst@object{ncbar}}
\def\ncbar@i{\check@arrow{\ncbar@ii}}
\def\ncbar@ii#1#2{\nc@object{Open}{#1}{#2}{1.5}{%
\ncangles@iii /AngleB \psk@angleA def \tx@NCBar}}
\def\pcbar{\pst@object{pcbar}}
\def\pcbar@i{\pc@object\ncbar@ii}
\define@key[psset]{pst-node}{lineAngle}[0]{%
  \ifdim#1pt=\z@\else\psset{armB=0.5}\fi
  \def\psk@lineAngle{#1}}%
\psset[pst-node]{lineAngle=0}%
\def\tx@NCDiag{NCDiag }
\def\ncdiag{\pst@object{ncdiag}}
\def\ncdiag@i{\check@arrow{\ncdiag@ii}}
\def\ncdiag@ii#1#2{%
  \nc@object{Open}{#1}{#2}{1.5}{\ncangles@iii \psk@lineAngle\space \tx@NCDiag}}
\def\pcdiag{\pst@object{pcdiag}}
\def\pcdiag@i{\pc@object\ncdiag@ii}
\def\tx@NCDiagg{NCDiagg }
\def\ncdiagg{\pst@object{ncdiagg}}
\def\ncdiagg@i{\check@arrow{\ncdiagg@ii}}
\def\ncdiagg@ii#1#2{%
  \nc@object{Open}{#1}{#2}{.5}{\ncangles@iii \psk@lineAngle\space \tx@NCDiagg}}
\def\pcdiagg{\pst@object{pcdiagg}}
\def\pcdiagg@i{\pc@object\ncdiagg@ii}
\def\tx@NCLoop{NCLoop }
\define@key[psset]{pst-node}{loopsize}{\pst@getlength{#1}\psk@loopsize}
\psset[pst-node]{loopsize=1cm}
\def\ncloop{\pst@object{ncloop}}
\def\ncloop@i{\check@arrow{\ncloop@ii}}
\def\ncloop@ii#1#2{%
\nc@object{Open}{#1}{#2}{2.5}%
{\ncangles@iii /loopsize \psk@loopsize def \tx@NCLoop}}
\def\pcloop{\pst@object{pcloop}}
\def\pcloop@i{\pc@object\ncloop@ii}
\def\tx@NCCircle{NCCircle }
\def\nccircle{\pst@object{nccircle}}
\def\nccircle@i{\check@arrow{\nccircle@ii}}
\def\nccircle@ii#1#2{%
\pssetlength\pst@dima{#2}%
\nc@object{Open}{#1}{#1}{.5}{%
/AngleA \psk@angleA def
/r \pst@number\pst@dima def
\tx@NCCircle \psarc@v end}}
\def\tx@NCBox{NCBox }
\def\ncbox{\pst@object{ncbox}}
\def\ncbox@i{\check@arrow{\ncbox@ii}}
\def\ncbox@ii#1#2{%
\def\pst@linetype{2}%
\nc@object{Closed}{#1}{#2}{.5}{%
tx@Dict begin \psline@iii pop end
\psk@boxheight \psk@boxdepth
\tx@NCBox}}
\def\pcbox{\pst@object{pcbox}}
\def\pcbox@i{\pc@object\ncbox@ii}
\def\tx@NCArcBox{NCArcBox }
\define@key[psset]{pst-node}{boxheight}[0.4cm]{\pst@getlength{#1}\psk@boxheight}
\define@key[psset]{pst-node}{boxdepth}[0.4cm]{\pst@getlength{#1}\psk@boxdepth}
\define@key[psset]{pst-node}{boxsize}[0.4cm]{%
  \pst@getlength{#1}\psk@boxheight%
  \let\psk@boxdepth\psk@boxheight}
\psset[pst-node]{boxsize=0.4cm}
\def\ncarcbox{\pst@object{ncarcbox}}
\def\ncarcbox@i{\check@arrow{\ncarcbox@ii}}
\def\ncarcbox@ii#1#2{%
\def\pst@linetype{1}%
\nc@object{Closed}{#1}{#2}{.5}{%
\psk@arcangleA \psk@boxheight \psk@boxdepth \pst@number\pslinearc
\tx@NCArcBox}}
\def\pcarcbox{\pst@object{pcarcbox}}
\def\pcarcbox@i{\pc@object\ncarcbox@ii}
\def\tx@Tfan{Tfan }
\begingroup
\catcode`\:=13
\gdef\pst@activerot{\def:{\string:}}
\endgroup
\define@key[psset]{pst-node}{nrot}[0]{%
  \begingroup
  \pst@activerot
  \pst@expandafter{\@ifnextchar:{\psset@@nrot}{\psset@@rot}}{#1}\@nil
  \global\let\pst@tempg\psk@rot
  \endgroup
  \let\psk@nrot\pst@tempg}
\def\psset@@nrot:#1\@nil{%
  \psset@@rot#1\@nil
  \edef\psk@rot{NAngle \ifx\psk@rot\@empty\else\psk@rot add \fi}}
\psset[pst-node]{nrot=0}
\def\tx@LPutCoor{LPutCoor }
\def\tx@LPut{LPut }
\define@key[psset]{pst-node}{npos}[{}]{%
  \def\pst@tempa{#1}%
  \ifx\pst@tempa\@empty\def\psk@npos{\npos@default}\else\pst@checknum{#1}\psk@npos\fi}
\psset[pst-node]{npos=}
\def\ncput{\pst@object{ncput}}
\def\ncput@i{\pst@killglue\pst@makebox{\ncput@ii}}
\def\ncput@ii{%
  \begingroup%
  \use@par%
  \if@star\pst@starbox\fi%
  \pst@makesmall\pst@hbox%
  \pst@rotate\psk@nrot\pst@hbox%
  \ncput@iii%
  \endgroup%
  \pst@shortput}
\def\ncput@iii{%
  \leavevmode%
  \hbox{%
    \pst@Verb{
      \pst@nodedict
      /t \psk@npos def
      \tx@LPut
      end
      \tx@PutBegin}%
    \box\pst@hbox%
    \pst@Verb{\tx@PutEnd}}}
\def\naput{\pst@object{naput}}
\def\naput@i{\pst@killglue\pst@makebox{\naput@ii{NAngle 90 add}}}
\def\naput@ii#1{%
  \begingroup
  \use@par
  \if@star\pst@starbox\fi
  \def\psk@refangle{#1 }%
  \let\psk@rot\psk@nrot
  \pst@Verb{ 
    gsave  STV CP T /ps@refangle {#1 } def 
    /ps@rot { \psk@rot } def grestore }
  \uput@vii
  {exch pop add a \tx@PtoC h1 add exch w1 add exch }%
  {tx@Dict /NCLW known { NCLW add } if }%
  \ncput@iii
  \endgroup
  \pst@shortput}
\def\nbput{\pst@object{nbput}}
\def\nbput@i{\pst@killglue\pst@makebox{\naput@ii{NAngle 90 sub}}}
\define@key[psset]{pst-node}{tpos}[0.5]{%
  \pst@checknum{#1}\psk@tpos
  \ifdim\psk@tpos \p@<\z@
    \def\psk@tpos{.5}%
    \@pstrickserr{Bad `tpos' value: `#1'. Must be 0<tpos<1}\@ehpa
  \else
    \ifdim\psk@tpos \p@>\p@
      \def\psk@tpos{.5}%
      \@pstrickserr{Bad `tpos' value: `#1'. Must be 0<tpos<1}\@ehpa%
    \fi%
  \fi}
\psset[pst-node]{tpos=0.5}
\def\nlput{\pst@object{nlput}}
\def\nlput@i(#1)(#2)#3#4{%
  \begin@SpecialObj
  \psLDNode(#1)(#2){#3}{temp@lnput}
  \pcline[linestyle=none](#1)(temp@lnput)%
  \ncput[npos=1]{#4}%
  \end@SpecialObj}
\def\tvput{\pst@object{tvput}}
\def\tvput@i{\pst@makebox{\psput@tput{H}{1}}}
\def\tlput{\pst@object{tlput}}
\def\tlput@i{\pst@makebox{\psput@tput{H}{true}}}
\def\trput{\pst@object{trput}}
\def\trput@i{\pst@makebox{\psput@tput{H}{false}}}
\def\thput{\pst@object{thput}}
\def\thput@i{\pst@makebox{\psput@tput{V}{1}}}
\def\taput{\pst@object{taput}}
\def\taput@i{\pst@makebox{\psput@tput{V}{true}}}
\def\tbput{\pst@object{tbput}}
\def\tbput@i{\pst@makebox{\psput@tput{V}{false}}}
\def\tx@HPutAdjust{HPutAdjust }
\def\tx@VPutAdjust{VPutAdjust }
\def\psput@tput#1#2{%
  \begingroup
  \use@par
  \pst@tputmakesmall
  \leavevmode
  \hbox{%
    \pst@Verb{%
      \pst@nodedict
      /t \psk@tpos \pst@tposflip def
      tx@NodeDict /HPutPos known
        { #1PutPos }
        { CP /Y exch def /X exch def /NAngle 0 def /NCLW 0 def }
      ifelse
      /Sin NAngle sin def
      /Cos NAngle cos def
      /s \pst@number\pslabelsep NCLW add def
      /l \pst@number\pst@dima def
      /r \pst@number\pst@dimb def
      /h \pst@number\pst@dimc def
      /d \pst@number\pst@dimd def
      \ifnum1=0#2 \else
        /flag #2 def
        \csname tx@#1PutAdjust\endcsname
      \fi
      \tx@LPutCoor
      end
      \tx@PutBegin}%
    \box\pst@hbox
    \pst@Verb{\tx@PutEnd}}%
  \endgroup
  \pst@shortput}
\def\pst@tposflip{}
\def\pst@tputmakesmall{%
\pst@dima=\wd\pst@hbox
\divide\pst@dima 2
\pst@dimg=\psk@href\pst@dimg
\pst@dimb\pst@dima
\advance\pst@dima\pst@dimg 
\advance\pst@dimb-\pst@dimg 
\pst@dimd=\psk@vref\relax
\pst@dimc=\ht\pst@hbox
\advance\pst@dimc-\pst@dimd 
\advance\pst@dimd\dp\pst@hbox 
\setbox\pst@hbox=\hbox to\z@{%
\kern-\pst@dima\vbox to\z@{\vss\box\pst@hbox\vskip-\pst@dimd}\hss}}
\def\MakeShortNab#1#2{%
  \def\pst@shortput@nab{%
    \def\pst@tempg{\next}%
    \ifx#1\next
      \let\pst@tempg\naput
    \else
      \ifx#2\next
        \let\pst@tempg\nbput
      \else
        \ifx\@sptoken\next
          \let\pst@tempg\pst@shortput
        \fi
      \fi
    \fi
    \pst@tempg}}
\MakeShortNab{^}{_}
\def\MakeShortTablr#1#2#3#4{%
  \def\pst@shortput@tablr{%
    \def\pst@tempg{\next}%
    \ifx#1\next
      \let\pst@tempg\taput
    \else
      \ifx#2\next
        \let\pst@tempg\tbput
      \else
        \ifx#3\next
          \let\pst@tempg\tlput
        \else
          \ifx#4\next
            \let\pst@tempg\trput
          \else
            \ifx\@sptoken\next
              \let\pst@tempg\pst@shortput
            \fi
          \fi
        \fi
      \fi
    \fi
    \pst@tempg}}
\MakeShortTablr{^}{_}{<}{>}
\def\MakeShortTab#1#2{%
  \def\pst@shortput@tab{%
    \def\pst@tempg{\next}%
    \ifx#1\next
      \def\pst@tempg{%
        \@nameuse{%
          t\ifodd\psk@treemode\ifpstreeflip b\else a\fi
          \else\ifpstreeflip r\else l\fi\fi put}}%
    \else
      \ifx#2\next
        \def\pst@tempg{%
          \@nameuse{%
            t\ifodd\psk@treemode\ifpstreeflip a\else b\fi
            \else\ifpstreeflip l\else r\fi\fi put}}%
      \else
        \ifx\@sptoken\next
          \let\pst@tempg\pst@shortput
        \fi
      \fi
    \fi
    \pst@tempg}}
\MakeShortTab{^}{_}
\define@key[psset]{pst-node}{shortput}[none]{%
  \def\pst@tempg{#1}%
  \ifx\pst@tempg\@none
    \let\pst@shortput\ignorespaces
  \else
    \@ifundefined{pst@shortput@#1}%
     {\@pstrickserr{Bad short put: `#1'}\@ehpa}%
     {\edef\pst@shortput{\noexpand\afterassignment\expandafter\noexpand
      \csname pst@shortput@#1\endcsname\noexpand\let\noexpand\next}}%
  \fi}
\psset[pst-node]{shortput=none}
\def\lput{\def\pst@par{}\pst@ifstar{\@ifnextchar[{\lput@i}{\lput@ii}}}
\def\lput@i[#1]{\addto@par{ref=#1}\lput@ii}
\def\lput@ii{\@ifnextchar({\lput@iv}{\lput@iii}}
\def\lput@iii#1{\addto@par{nrot=#1}\@ifnextchar({\lput@iv}{\ncput@i}}
\def\lput@iv(#1){\addto@par{npos=#1}\ncput@i}
\def\mput{\def\pst@par{}\pst@ifstar{\@ifnextchar[{\mput@i}{\ncput@i}}}
\def\mput@i[#1]{\addto@par{ref=#1}\ncput@i}
\def\Lput{\def\pst@par{}\pst@ifstar{\@ifnextchar[{\Lput@ii}{\Lput@i}}}
\def\Lput@i#1{\addto@par{labelsep=#1}\Lput@ii}
\def\Lput@ii[#1]{\addto@par{ref={#1}}\@ifnextchar({\Lput@iv}{\Lput@iii}}
\def\Lput@iii#1{\addto@par{nrot={#1}}\@ifnextchar({\Lput@iv}{\Lput@v}}
\def\Lput@iv(#1){\addto@par{npos=#1}\Lput@v}
\def\Lput@v{\pst@killglue\pst@makebox{\Lput@vi}}
\def\Lput@vi{%
\begingroup
\use@par
\if@star\pst@starbox\fi
\Rput@vi
\pst@makesmall\pst@hbox
\ifx\psk@rot\@empty\else\pst@rotate{ps@rot }\pst@hbox\fi
\ncput@iii
\endgroup
\pst@shortput}
\def\Mput{\def\pst@par{}\pst@ifstar{\@ifnextchar[{\Mput@ii}{\Mput@i}}}
\def\Mput@i#1{\addto@par{labelsep=#1}\Mput@ii}
\def\Mput@ii[#1]{\addto@par{ref={#1}}\Lput@v}
\def\aput@#1{\def\pst@par{}\pst@ifstar{\@ifnextchar[{\aput@i#1}{\aput@ii#1}}}
\def\aput@i#1[#2]{\addto@par{labelsep=#2}\aput@ii#1}
\def\aput@ii#1{\@ifnextchar({\aput@iv#1}{\aput@iii#1}}
\def\aput@iii#1#2{\addto@par{nrot=#2}\@ifnextchar({\aput@iv#1}{#1}}
\def\aput@iv#1(#2){\addto@par{npos=#2}#1}
\def\aput{\aput@\naput@i}
\def\bput{\aput@\nbput@i}
\def\Aput{\def\pst@par{}\pst@ifstar{\@ifnextchar[{\Aput@i}{\naput@i}}}
\def\Aput@i[#1]{\addto@par{labelsep=#1}\naput@i}
\def\Bput{\def\pst@par{}\pst@ifstar{\@ifnextchar[{\Bput@i}{\nbput@i}}}
\def\Bput@i[#1]{\addto@par{labelsep=#1}\nbput@i}
\def\node@coor#1;#2\@nil{
  \pst@getnode{#1}\pst@tempg
  \edef\pst@coor{%
    \pst@nodedict
    tx@NodeDict \pst@tempg known
    \pslbrace \pst@tempg load \tx@GetCenter \psrbrace
    \pslbrace 0 0 \psrbrace ifelse
    end }}
\def\Node@coor[#1]#2;#3\@nil{
\begingroup
\psset{angle=0,#1}
\@ifnextchar\bgroup{\Node@@@coor}
                   {\Node@@coor}#2\@nil
\endgroup
\let\pst@coor\pst@tempg}
\def\Node@@coor#1\@nil{%
\pst@getnode{#1}\pst@tempg
\xdef\pst@tempg{%
\pst@nodedict
tx@NodeDict \pst@tempg known
  { \psk@nodesepA \psk@angleA 
    \pst@tempg load \psk@nodeseptypeA \tx@GetEdge
    \psk@offsetA \psk@angleA \tx@AddOffset
    \pst@tempg load \tx@GetCenter
    3 -1 roll add 3 1 roll add exch }
  { CP } ifelse end }}
\def\Node@@@coor#1{
\pst@@getcoor{#1}%
\def\psk@angleA{%
  \pst@tempg load \tx@GetCenter \pst@coor
  3 -1 roll sub 3 1 roll sub neg \tx@Atan \psk@angleB add
  }%
\Node@@coor}
\def\nput{\pst@object{nput}}
\def\nput@i#1#2{\pst@killglue\pst@makebox{\nput@ii{#1}{#2}}}
\def\nput@ii#1#2{%
  \begingroup
  \use@par
  \if@star\pst@starbox\fi%
  \psset[pstricks]{refangle=#1}%
  \let\psk@angleA\psk@refangle
  \edef\psk@nodesepA{\pst@number\pslabelsep}%
  \def\psk@nodeseptypeA{0 }%
  \pslabelsep\z@
  \uput@vi
  \Node@@coor#2\@nil
  \let\pst@coor\pst@tempg
  \leavevmode
  \psput@special\pst@hbox
  \endgroup
  \ignorespaces}
\newcount\psrow
\newcount\pscol
\newcount\psmatrixcnt
\newskip\psrowsep
\newskip\pscolsep
\define@key[psset]{pst-node}{colsep}[1.5cm]{\pssetlength\pscolsep{#1}}
\define@key[psset]{pst-node}{rowsep}[1.5cm]{\pssetlength\psrowsep{#1}}
\psset[pst-node]{colsep=1.5cm}
\psset[pst-node]{rowsep=1.5cm}
\newif\ifpsmatrix
\ifx\mscount\@undefined\let\mscount\@multicnt\fi
\def\psmatrix{\begingroup{\ifnum0=`}\fi 
  \@ifnextchar[{\psmatrix@i}{\ifnum0=`{\fi}{}\psmatrix@ii}}
\def\psmatrix@i[#1]{%
  \ifnum0=`{\fi}{}%
  \psset{#1}%
  \psmatrix@ii}
\def\psmatrix@ii{%
  \KillGlue
  \edef\psm@beginmath{%
    \ifmmode$\m@th\ifinner\textstyle\else\displaystyle\fi\fi}%
  \edef\psm@endmath{\ifmmode$\fi}%
  \let\\\psm@cr
  \advance\psmatrixcnt by \@ne
  \def\psm@thenode{M-\the\psmatrixcnt-\the\psrow-\the\pscol}%
  \tabskip\z@
  \psrow=\@ne
  \pscol\z@
  \psset{shortput=tablr}%
  \leavevmode
  \vbox\bgroup\halign\bgroup&%
  \begingroup
  \global\advance\pscol by \@ne
  \csname psrowhook\romannumeral\psrow\endcsname
  \csname pscolhook\romannumeral\pscol\endcsname
  \psm@beginnode##\psm@endnode\endgroup
  \cr}
\def\endpsmatrix{%
  \crcr\egroup\unskip\egroup
  \endgroup}
\def\psm@cr{{\ifnum0=`}\fi\ps@ifnextchar[{\psm@@cr}{\psm@@@cr{}}}
\def\psm@@cr[#1]{\psm@@@cr{\vskip#1\relax}}
\def\psm@@@cr#1{%
  \ifnum0=`{\fi}{}\cr
  \noalign{%
  \global\advance\psrow 1
  \global\pscol\z@
  \vskip\psrowsep
  #1}}
\def\psm@beginnode{%
  \@ifnextchar\psm@endnode
    {\let\psm@endnode@i\relax\setbox\pst@hbox=\hbox{}}%
    {\pst@object{psm@beginnode}}}
\def\psm@beginnode@i{%
  \setbox\pst@hbox=\hbox\bgroup
  \psm@beginmath
  \begingroup
  \ignorespaces}
\def\psm@endnode@i{%
  \unskip
  \endgroup
  \psm@endmath
  \egroup
  \use@par
  \@psttrue}
\def\psm@endnode{%
  \@pstfalse
  \psm@endnode@i
  \ifnum\pscol>1\relax \pshskip\pscolsep \fi
  \psk@mnodesize
  \hfil
  \Pst@nodealigntrue
  \if@pst\csname mnode@\psk@mnode\endcsname
  \else\csname mnode@\psk@emnode\endcsname\fi
  \psk@mcol
  \psk@@mnodesize}
\def\psspan#1{\global\mscount#1\relax\pstloop\ifnum\mscount>\@ne\sp@n\repeat}
\def\pstloop#1\repeat{\gdef\pstiterate{#1\relax\expandafter\pstiterate\fi}%
  \pstiterate
  \let\pstiterate\relax}
\define@key[psset]{pst-node}{name}[\relax]{\pst@getnode{#1}\psk@name}
\let\psk@name\relax
\define@key[psset]{pst-node}{mcol}[c]{%
  \ifx r#1\relax\let\psk@mcol\relax\else
    \ifx l#1\relax\let\psk@mcol\hfill\else
    \let\psk@mcol\hfil\fi\fi}
\psset[pst-node]{mcol=c}
\define@key[psset]{pst-node}{mnodesize}[-1pt]{%
  \pssetlength\pst@dimg{#1}%
  \ifdim\pst@dimg<\z@
    \let\psk@mnodesize\relax
    \let\psk@@mnodesize\relax
  \else
    \edef\psk@mnodesize{\noexpand\hbox to\number\pst@dimg sp\noexpand\bgroup}%
    \let\psk@@mnodesize\egroup
  \fi}
\psset[pst-node]{mnodesize=-1pt}
\def\mnode@R{\rnode@iii\Rnode@ii{\psm@thenode}}
\def\mnode@r{\rnode@iii\rnode@iv{\psm@thenode}}
\def\mnode@oval{\ovalnode@ii{\psm@thenode}}
\def\mnode@tri{\trinode@ii{\psm@thenode}}
\def\mnode@dia{\dianode@ii{\psm@thenode}}
\def\mnode@C{{\Pst@nodealigntrue\cnode@ii(\z@,\z@){\psk@radius}{\psm@thenode}}}
\def\mnode@f{{\Pst@nodealigntrue\fnode@ii(\z@,\z@){\psm@thenode}}}
\def\mnode@circle{\circlenode@ii{\psm@thenode}}
\def\mnode@Circle{\Circlenode@ii{\psm@thenode}}
\def\mnode@p{\pnode(\z@,\z@){\psm@thenode}}
\def\mnode@dot{\dotnode@ii(\z@,\z@){\psm@thenode}}
\def\mnode@none{\box\pst@hbox}
\define@key[psset]{pst-node}{mnode}[R]{%
  \@ifundefined{mnode@#1}%
    {\@pstrickserr{\string\psmatrix\space node `#1' not defined.}\@ehpa}%
    {\edef\psk@mnode{#1}}}
\define@key[psset]{pst-node}{emnode}[none]{%
  \@ifundefined{mnode@#1}%
    {\@pstrickserr{\string\psmatrix\space node `#1' not defined.}\@ehpa}%
    {\edef\psk@emnode{#1}}}
\psset[pst-node]{mnode=R,emnode=none}
\def\nccoil{\pst@object{nccoil}}
\def\nccoil@i{\check@arrow{\nccoil@ii}}
\def\nccoil@ii#1#2{\nc@object{Open}{#1}{#2}{.5}{
  \tx@NCCoor
  tx@Dict begin
  4 2 roll
  \psk@coilwidth \pscoilheight
  \psk@coilarmA \psk@coilarmB
  \psk@coilaspect \psk@coilinc
  \pst@coildict \tx@Coil end
  end}%
}
\def\nczigzag{\pst@object{nczigzag}}
\def\nczigzag@i{\check@arrow{\nczigzag@ii}}
\def\nczigzag@ii#1#2{\nc@object{Open}{#1}{#2}{.5}{
  \tx@NCCoor
  tx@Dict begin
  4 2 roll
  \pscoilheight
  \psk@coilwidth
  \psk@coilarmA
  \psk@coilarmB
  \pst@coildict \tx@ZigZag end
  \psline@iii
  \tx@Line
  end}%
}
\def\psGetNodeCenter#1{ tx@NodeDict begin /N@#1 load GetCenter end 
  \pst@number\psyunit div /#1.y exch def 	
  \pst@number\psxunit div /#1.x exch def }	
\def\psGetEdgeA#1#2{
  tx@NodeDict begin \psk@offsetA \psk@offsetB neg 
    \psk@nodesepA \psk@nodesepB 0 0 
    /N@#1 /N@#2 InitNC { NCCoor } if pop pop \tx@UserCoor end}
\def\psGetEdgeB#1#2{
  tx@NodeDict begin \psk@offsetA \psk@offsetB neg 
    \psk@nodesepA \psk@nodesepB 0 0 
    /N@#1 /N@#2 InitNC { NCCoor } if 4 2 roll pop pop \tx@UserCoor end}
\def\ncbarr{\pst@object{ncbarr}}
\def\ncbarr@i#1#2{%
  \begingroup
  \use@par%
  \psLNode(#1)(#2){0.5}{barr@tempNode}%
  \pst@dimc=\psk@angleA pt
  \pst@dimd=180pt
  \ifdim\pst@dimc=\z@\else\ifdim\pst@dimc=\pst@dimd\else\psset{angleA=0}\fi\fi
  \ncbar[arrows=-]{#1}{barr@tempNode}
  \ifdim\psk@angleA pt=\z@\relax
    \ncbar[angleA=180,angleB=180]{barr@tempNode}{#2}
  \else\ncbar[angleA=0,angleB=0]{barr@tempNode}{#2}\fi%
  \endgroup%
}
\def\psLNode(#1)(#2)#3#4{%
  \pst@getcoor{#1}\pst@tempA%
  \pst@getcoor{#2}\pst@tempB%
  \pnode(!
    \pst@tempA /YA exch \pst@number\psyunit div def
    /XA exch \pst@number\psxunit div def
    \pst@tempB /YB exch \pst@number\psyunit div def
    /XB exch \pst@number\psxunit div def
    /dx XB XA sub def
    /dy YB YA sub def
    XA dx #3\space mul add YA dy #3\space mul add){#4}}
\def\psLCNode(#1)#2(#3)#4#5{%
  \pst@getcoor{#1}\pst@tempA%
  \pst@getcoor{#3}\pst@tempB%
  \pnode(!
    \pst@tempA /YA exch \pst@number\psyunit div def
    /XA exch \pst@number\psxunit div def
    \pst@tempB /YB exch \pst@number\psyunit div def
    /XB exch \pst@number\psxunit div def
    XA #2\space mul XB #4\space mul add
    YA #2\space mul YB #4\space mul add){#5}}
\def\psLDNode(#1)(#2)#3#4{%
  \pst@getcoor{#1}\pst@tempA%
  \pst@getcoor{#2}\pst@tempB%
  \pssetlength\pst@dimb{#3}%
  \pnode(!%
    \pst@tempA /YA exch \pst@number\psyunit div def
    /XA exch \pst@number\psxunit div def
    \pst@tempB /YB exch \pst@number\psyunit div def
    /XB exch \pst@number\psxunit div def
    /dx XB XA sub def
    /dy YB YA sub def
    /angle dy dx Atan def
    /linelength \pst@number\pst@dimb \pst@number\psunit div def
    XA linelength angle cos mul add YA linelength angle sin mul add ){#4}%
}
\def\psRelNode{\pst@object{psRelNode}}
\def\psRelNode@i(#1)(#2)#3#4{{
  \use@par
  \pst@getcoor{#1}\pst@tempA%
  \pst@getcoor{#2}\pst@tempB%
  \pnode(!
    \pst@tempA /YA exch \pst@number\psyunit div def
    /XA exch \pst@number\psxunit div def
    \pst@tempB /YB exch \pst@number\psyunit div def
    /XB exch \pst@number\psxunit div def
    /AlphaStrich \psk@angleA\space def
    /unit \pst@number\psyunit \pst@number\psxunit div def 
    /dx XB XA sub  def
    /dy YB YA sub \ifPst@trueAngle\space unit mul \fi\space def
    /laenge dy dup mul dx dup mul add sqrt #3 mul def
    /Alpha dy dx atan def 
    /beta Alpha AlphaStrich add def
    laenge beta cos mul XA add
    laenge beta sin mul \ifPst@trueAngle\space unit div \fi\space YA add ){#4}%
}\ignorespaces}
\define@cmdkeys[psset]{pstricks-add}[PSTPSPNk@]{
  blName,bcName,brName,
  clName,ccName,crName,
  tlName,tcName,trName}[]{}%
\psset[pstricks-add]{%
  blName=PSPbl,bcName=PSPbc,brName=PSPbr,
  clName=PSPcl,ccName=PSPcc,crName=PSPcr,
  tlName=PSPtl,tcName=PSPtc,trName=PSPtr}
\def\psDefPSPNodes{\def\pst@par{}\pst@object{psDefPSPNodes}}
\def\psDefPSPNodes@i{%
  \pst@killglue
  \begingroup
  \use@par
  \expandafter\psDefPSPNodes@ii\pic@coor}
\def\psDefPSPNodes@ii(#1)(#2)(#3){%
    \pnode(#1){PSPN@temp}\pnode([angle=45]PSPN@temp){\PSTPSPNk@blName}
    \pnode(#3){PSPN@temp}\pnode([angle=-135]PSPN@temp){\PSTPSPNk@trName}
    \pnode(\PSTPSPNk@blName|\PSTPSPNk@trName){\PSTPSPNk@tlName}
    \pnode(\PSTPSPNk@trName|\PSTPSPNk@blName){\PSTPSPNk@brName}
    \ncline[linestyle=none]{\PSTPSPNk@blName}{\PSTPSPNk@tlName}
    \ncput[npos=.5]{\pnode{\PSTPSPNk@clName}}
    \ncline[linestyle=none]{\PSTPSPNk@blName}{\PSTPSPNk@brName}
    \ncput[npos=.5]{\pnode{\PSTPSPNk@bcName}}
    \pnode(\PSTPSPNk@brName|\PSTPSPNk@clName){\PSTPSPNk@crName}
    \pnode(\PSTPSPNk@bcName|\PSTPSPNk@trName){\PSTPSPNk@tcName}
    \pnode(\PSTPSPNk@bcName|\PSTPSPNk@clName){\PSTPSPNk@ccName}
  \endgroup
  \ignorespaces}
\def\psDefBoxNodes#1#2{\rnode[tl]{#1:tl}{\rnode[Bl]{#1:Bl}{\rnode[tr]{#1:tr}{%
\rnode[bl]{#1:bl}{\rnode[Br]{#1:Br}{\rnode[br]{#1:br}{#2}}}}}}%
\pnode(!\psGetNodeCenter{#1:bl}
          \psGetNodeCenter{#1:tl} 
          #1:bl.x #1:tl.x add 2 div #1:bl.y #1:tl.y add 2 div ){#1:Cl}%
\pnode(!\psGetNodeCenter{#1:tr}
          \psGetNodeCenter{#1:br} 
          #1:tr.x #1:br.x add 2 div #1:tr.y #1:br.y add 2 div ){#1:Cr}%
\pnode(!\psGetNodeCenter{#1:Cl}
          \psGetNodeCenter{#1:Cr} 
          #1:Cl.x #1:Cr.x add 2 div #1:Cl.y #1:Cr.y add 2 div ){#1:C}%
\pnode(!\psGetNodeCenter{#1:Br}
          \psGetNodeCenter{#1:Bl} 
          #1:Br.x #1:Bl.x add 2 div #1:Br.y #1:Bl.y add 2 div ){#1:BC}%
\pnode(!\psGetNodeCenter{#1:tr}
          \psGetNodeCenter{#1:tl} 
          #1:tr.x #1:tl.x add 2 div #1:tr.y #1:tl.y add 2 div ){#1:tC}%
\pnode(!\psGetNodeCenter{#1:br}
          \psGetNodeCenter{#1:bl} 
          #1:br.x #1:bl.x add 2 div #1:br.y #1:bl.y add 2 div ){#1:bC}}%
\newcount\pst@args
\newcount\num@pts
\newcount\pst@argcnt
\def\PST@root{}
\let\pst@next\relax
\def\my@tempA{}
\def\my@tempB{}
\def\my@tempC{}
\def\my@tempD{}
\def\my@next{}
\newif\if@paren%
\newif\if@equal%
\newif\if@colon%
\newif\ifshow
\def\plussign{+}\def\minussign{-}
\def\defaultvalue#1#2{
  \ifdefined#1\ifx#1\@empty\xdef#1{#2}\fi\else\xdef#1{#2}\fi}%
\def\testAlg#1|#2\@nil{%
\ifx\relax#2\relax%
   \let\my@next\psparnode\xdef\my@tempD{}%
\else%
   \let\my@next\algparnode\xdef\my@tempD{A}
\fi}%
\def\trim #1{\expandafter\trim@\expandafter{#1 }#1}%
\def\trim@ #1{\trim@@ @#1 @ #1 @ @@}%
\def\trim@@ #1@ #2@ #3@@{\trim@@@\empty #2 @}%
\def\unbrace#1{#1}%
\unbrace{\def\trim@@@ #1 } #2@#3{\expandafter\def%
  \expandafter #3\expandafter {#1}}%
\def\hasparen#1(#2\@nil{
  \ifx\relax#2\relax \@parenfalse \else \@parentrue\fi}%
\def\hasequal#1=#2\@nil{
  \ifx\relax#2\relax \@equalfalse \else \@equaltrue\fi
  \hascolon#2:\@nil}%
\def\hascolon#1:#2\@nil{
\ifx\relax#2\relax \@colonfalse \else \@colontrue\fi}%
\def\equalwhat#1=#2:#3\@nil{{#2}{#3}}%
\def\parsenodexn#1(#2)#3\@nil{%
  \def\coeffA{#1}\edef\nodeA{#2}%
  \trim\coeffA%
  \ifx\nodeA\@empty\else%
    \pnode(#2){@@TMP}%
    \ifx\coeffA\@empty\def\coeffA{1}\else%
      \ifx\coeffA\plussign\def\coeffA{1}\else\ifx\coeffA\minussign\def\coeffA{-1}\fi\fi\fi%
  \edef\cmd{\noexpand\psLCNode(@TMP\the\pst@argcnt){1}(@@TMP){\coeffA}{@TMP}}%
  \cmd%
  \advance\pst@argcnt by \@ne%
  \pnode(@TMP){@TMP\the\pst@argcnt}%
  \parsenodexn#3\@nil%
  \fi}%
\def\normalvec(#1)#2{%
  \psRelNodeVar(0,0)(#1)(0,1){#2}}%
\def\curvepnode#1#2#3{%
  \edef\my@tempA{#2}
  \expandafter\testAlg\my@tempA|\@nil\my@next {#1}{#2}{#3}}
\def\psparnode#1#2#3{%
  \pnode(!/t #1 def #2){#3}%
  \pnode(!/t #1 .001 sub def #2 
          /t #1 .001 add def 
           #2 3 -1 roll sub 3 1 roll sub neg 
           2 copy Pyth dup 3 1 roll div 3 1 roll div ){#3tang}}
\def\algparnode#1#2#3{%
  \pstVerb{tx@Dict begin /Func (#2) AlgParser cvx def end }
  \pnode(!/t #1 def Func){#3}
  \pnode(!/t #1 .001 sub def Func 
          /t #1 .001 add def 
          Func 3 -1 roll sub 3 1 roll sub neg 
          2 copy Pyth dup 3 1 roll div 3 1 roll div ){#3tang}
}%
\def\nodex#1{%
\expandafter\hasparen#1(\@nil%
\if@paren
  \pnode(0,0){@TMP0}%
  \pst@argcnt=0%
  \expandafter\parsenodexn#1()\@nil%
\else%
  \def\my@tempC{#1}%
  \ifx\my@tempC\@empty\pnode(0,0){@TMP}\else\pnode(#1){@TMP}\fi%
\fi}
\def\nodexn#1#2{%
\expandafter\hasparen#1(\@nil
\if@paren
  \pnode(0,0){@TMP0}%
  \pst@argcnt=0%
  \parsenodexn#1()\@nil%
  \pnode(@TMP){#2}%
\else%
  \def\my@tempC{#1}%
  \ifx\my@tempC\@empty\pnode(0,0){#2}\else\pnode(#1){#2}\fi%
\fi}
\def\psxline{\pst@object{psxline}}%
\def\psxline@i{\@ifnextchar({\psxline@iii}{\psxline@ii}}%
\def\psxline@ii#1{%
\addto@par{arrows=#1}%
\psxline@iii}%
\def\psxline@iii(#1)#2#3{{
\pst@killglue%
\use@par%
\nodexn{#2}{@TMP@a}%
\AplusB(#1)(@TMP@a){@TMP@A}%
\nodexn{#3}{@TMP@a}%
\AplusB(#1)(@TMP@a){@TMP@B}%
\psline(@TMP@A)(@TMP@B)%
}%
\ignorespaces}%
\def\curvepnodes{\pst@object{curvepnodes}}
\def\curvepnodes@i#1#2#3#4{{
  \pst@killglue
  \use@par
  \edef\my@tempA{#3}
  \expandafter\testAlg\my@tempA|\@nil %
  \pstVerb{%
	tx@Dict begin 
	/t0 #1 def
	/t1 #2 def  
	 t1 t0 sub end \psk@plotpoints div /dt exch def }%
  \pst@cntc=\psk@plotpoints\relax
  \advance\pst@cntc by \@ne\relax 
  \ifx\my@tempD\@empty\pstVerb{tx@Dict begin /Func (#3) cvx def end }
  \else\pstVerb{tx@Dict begin /Func (#3 ) AlgParser cvx def end }%
  \fi%
    \multido{\i=0+1}{\pst@cntc}{%
      \pnode(! /t #1 dt \i\space mul add def Func ){#4\i}}
    \expandafter\xdef \csname #4nodecount\endcsname {\psk@plotpoints}%
    \ifnum\Pst@Debug>0 \typeout{Created nodes #40 .. #4\psk@plotpoints}\fi%
}\ignorespaces}%
\def\fnpnode{\pst@object{fnpnode}}
\def\fnpnode@i#1#2#3{{
  \pst@killglue
  \use@par
  \ifPst@algebraic\pnode(*#1 {#2}){#3}\else\pnode(! /x #1 def x #2){#3}\fi
}\ignorespaces}%
\def\fnpnodes{\pst@object{fnpnodes}}
\def\fnpnodes@i#1#2#3#4{{
\pst@killglue
\use@par
\pst@dima=#1pt \pst@dimb=#2pt \advance\pst@dimb -\pst@dima%
\pst@cnta=\psk@plotpoints \relax 
\def\PST@root{#4}
\divide\pst@dimb by \pst@cnta
\pst@cntc=\pst@cnta %
\advance\pst@cntc by 1 \relax 
\ifPst@algebraic 
  \multido{\i=0+1}{\pst@cntc}{\pnode(*{\pst@number\pst@dima} {#3}){#4\i}
  \advance\pst@dima \pst@dimb}%
\else
    \multido{\i=0+1}{\pst@cntc}{\pnode(!/x \pst@number\pst@dima\space def x #3){#4\i}%
  \advance\pst@dima \pst@dimb}%
\fi%
  \expandafter\xdef \csname \PST@root nodecount\endcsname {\the\pst@cnta}%
  \ifnum\Pst@Debug>0 \typeout{Created nodes #40 .. #4\the\pst@cnta}\fi%
}\ignorespaces}%
\def\AtoB(#1)(#2)#3{\psLCNodeVar(#1)(#2)(-1,1){#3}}
\def\AplusB(#1)(#2)#3{\psLCNodeVar(#1)(#2)(1,1){#3}}
\def\midAB(#1)(#2)#3{\psLCNodeVar(#1)(#2)(.5,.5){#3}}
\def\psnline{\pst@object{psnline}}
\def\psnline@i{\pst@getarrows{\psnline@ii}}
\def\psnline@ii(#1,#2)#3{{%
\pst@killglue%
\use@par%
\pst@cnta=#2 \relax\advance\pst@cnta by 1
\edef\@tmp{}%
\multido{\i=#1+1}{\pst@cnta}{\xdef\@tmp{\@tmp(#3\i)}}%
\expandafter\psline\@tmp}%
\ignorespaces}%
\def\psnpolygon{\pst@object{psnpolygon}}
\def\psnpolygon@i{\pst@getarrows{\psnpolygon@ii}}
\def\psnpolygon@ii(#1,#2)#3{{%
\pst@killglue%
\use@par%
\pst@cnta=#2 \relax\advance\pst@cnta by 1
\edef\@tmp{}%
\multido{\i=#1+1}{\pst@cnta}{\xdef\@tmp{\@tmp(#3\i)}}%
\expandafter\pspolygon\@tmp}%
\ignorespaces}%
\def\psncurve{\pst@object{psncurve}}
\def\psncurve@i{\pst@getarrows{\psncurve@ii}}
\def\psncurve@ii(#1,#2)#3{{%
\pst@killglue%
\use@par%
\pst@cnta=#2 \relax\advance\pst@cnta by 1
\edef\@tmp{}%
\multido{\i=#1+1}{\pst@cnta}{\xdef\@tmp{\@tmp(#3\i)}}%
\expandafter\pscurve\@tmp}%
\ignorespaces}%
\def\psnccurve{\pst@object{psnccurve}}
\def\psnccurve@i{\pst@getarrows{\psnccurve@ii}}
\def\psnccurve@ii(#1,#2)#3{{%
\pst@killglue%
\use@par%
\pst@cnta=#2 \relax\advance\pst@cnta by 1
\xdef\@tmp{}%
\multido{\i=#1+1}{\pst@cnta}{\xdef\@tmp{\@tmp(#3\i)}}%
\expandafter\psccurve\@tmp}%
\ignorespaces}%
\def\shownode(#1){
  \pst@killglue%
  \pstVerb{%
    gsave tx@Dict begin %
    tx@NodeDict /N@#1 known { 
      /tmpar [(Node #1: ) <28> () (, ) () <29>] def %
      /str 12 string def 
      STV CP T \psGetNodeCenter{#1}\space 
      tmpar 2 #1.x str cvs put 
      /str 12 string def 
      tmpar 4 #1.y str cvs put 
      tmpar concatstringarray = }%
    {
      (Node #1: (NOT KNOWN)) = %
    } ifelse %
    end grestore }%
  \ignorespaces}%
\def\pnodes@ii#1{\getnodelist{#1}{}}
\def\getnodelist#1#2{%
\pst@args=0 \relax%
\def\PST@root{#1}%
\def\pst@next{#2}
\getnext@Node}%
\def\getnext@Node{\@ifnextchar({\getnext@Node@i}%
  {\advance\pst@args by \m@ne \expandafter\xdef \csname \PST@root nodecount\endcsname {\the\pst@args}
  \ifnum\Pst@Debug>0 \typeout{Created nodes \PST@root0 .. \PST@root\the\pst@args}\fi%
  \pst@next}%
}%
\def\getnext@Node@i(#1){%
\pnode(#1){\PST@root\the\pst@args}%
\advance\pst@args by \@ne\relax%
\getnext@Node}%
\def\psLCNodeVar(#1)(#2)(#3)#4{%
\pst@getcoor{#1}\my@tempA%
\pst@getcoor{#2}\my@tempB%
\pnode(#3){tmpLCn@de}%
\pnode(!%
  \my@tempA /YA exch \pst@number\psyunit div def
  /XA exch \pst@number\psxunit div def
  \my@tempB /YB exch \pst@number\psyunit div def
  /XB exch \pst@number\psxunit div def 
  \psGetNodeCenter{tmpLCn@de}\space
  XA tmpLCn@de.x mul XB tmpLCn@de.y mul add
  YA tmpLCn@de.x mul YB tmpLCn@de.y mul add){tmpLCn@deA}%
\pnode(tmpLCn@deA){#4}%
}%
\def\psRelNodeVar{\pst@object{psRelNodeVar}}
\def\psRelNodeVar@i(#1)(#2)(#3)#4{{
  \use@par
  \pst@getcoor{#1}\my@tempA%
  \pst@getcoor{#2}\my@tempB%
   \pnode(#3){tmpn@de}%
\pnode(!
  /unit \pst@number\psyunit \pst@number\psxunit div def 
    \my@tempA /YA exch \pst@number\psyunit div def
    /XA exch \pst@number\psxunit div def
    \my@tempB /YB exch \pst@number\psyunit div YA sub 
    \ifPst@trueAngle\space unit mul \fi\space def
    /XB exch \pst@number\psxunit div XA sub def
    \psGetNodeCenter{tmpn@de}
    XB tmpn@de.x mul YB tmpn@de.y mul sub
    YB tmpn@de.x mul XB tmpn@de.y mul add
    \ifPst@trueAngle\space unit div \fi\space 
   YA add exch XA add exch 
    ){#4}%
}}
\def\psRelLineVar{\pst@object{psRelLineVar}}
\def\psRelLineVar@i{\@ifnextchar({\psRelLineVar@iii}{\psRelLineVar@ii}}
\def\psRelLineVar@ii#1{%
  \addto@par{arrows=#1}%
  \psRelLineVar@iii}
\def\psRelLineVar@iii(#1)(#2)(#3)#4{{%
  \pst@killglue
  \use@par
  \psRelNodeVar(#1)(#2)(#3){#4}%
  \psline(#1)(#4)%
}\ignorespaces}
\def\rhombus#1(#2)(#3)#4#5{
\AtoB(#2)(#3){node@P}
\pnode(! 
/tmp \psGetNodeCenter{node@P} node@P.x node@P.y 
Pyth 2 div def 
/ang tmp #1\space div Acos def 
#1\space tmp 2 mul div 
dup ang cos mul exch ang sin mul ){node@A1}
\pnode(! \psGetNodeCenter{node@A1} node@A1.x node@A1.y neg ){node@A2}
\psRelNodeVar(#2)(#3)(node@A1){#4}%
\psRelNodeVar(#2)(#3)(node@A2){#5}%
}%
\def\psrline{\pst@object{psrline}}
\def\psrline@i{\@ifnextchar({\psrline@iii}{\psrline@ii}}%
\def\psrline@ii#1{%
\addto@par{arrows=#1}%
\psrline@iii}%
\def\psrline@iii{%
\getnodelist{@tmpnode}{\psrline@iv}%
}%
\def\psrline@iv{%
   \ifnum\pst@args<0\else
      \pnode(@tmpnode0){@tmpnodeB0}%
      \multido{\iA=1+1,\iB=0+1}{\pst@args}{%
      \AplusB(@tmpnodeB\iB)(@tmpnode\iA){@tmpnodeB\iA}}%
      \psrline@v%
   \fi%
}%
\def\psrline@v{{
  \pst@killglue%
  \use@par%
  \xdef\tmp{(@tmpnodeB0)}%
  \multido{\i=1+1}{\pst@args}%
{\xdef\tmp{\tmp(@tmpnodeB\i)}}%
\expandafter\psline\tmp%
}\ignorespaces}%
\def\polyIntersections#1#2(#3)(#4){%
\def\nodenameA{#1}\def\nodenameB{#2}%
\pnode(#3){P@A}\pnode(#4){P@B}%
\@ifnextchar({\polyIntersections@next}{\polyIntersections@ii}%
}%
\def\polyIntersections@ii#1#2{%
\def\root@node{#1}\num@pts=#2 \relax%
\polyIntersections@iii}%
\def\polyIntersections@next{
\def\root@node{P@}\getnodelist{P@}{\num@pts=\pst@args \relax\polyIntersections@iii}%
}%
\def\polyIntersections@iii{
\pst@cnta=\num@pts \relax\advance\pst@cnta by 1 \relax%
\pstVerb{%
 /xarray \the\pst@cnta\space array def
 /yarray \the\pst@cnta\space array def  tx@Dict begin }%
\multido{\i=0+1}{\the\pst@cnta}{\pstVerb{ \psGetNodeCenter{\root@node\i} xarray \i\space \root@node\i.x put yarray \i\space \root@node\i.y put }}%
\pstVerb{ /tposmin 100 def /tnegmax -100 def 
\psGetNodeCenter{P@B} \psGetNodeCenter{P@A} 
/dx P@B.x P@A.x sub def 
/dy P@B.y P@A.y sub def 
/lenAB dx dy Pyth def
/oldx xarray 0 get def /oldy yarray 0 get def 
1 1 \the\num@pts\space {/k exch def /newx xarray k get def /newy yarray k get def 
/ddx newx oldx sub def /ddy newy oldy sub def 
/det ddy dx mul ddx dy mul sub def
det abs lenAB ddx ddy Pyth mul .001 mul gt 
{/ac oldx P@A.x sub def /bd oldy P@A.y sub def 
 /tt  ac ddy mul bd ddx mul sub det div def 
 /ss ac  dy mul bd dx mul sub det div def 
ss 0 ge 
   {ss 1 le 
        {tt 0 lt {tt tnegmax gt {/tnegmax tt def} if } {tt tposmin lt {/tposmin tt def} if } ifelse }
    if } 
if }
 if 
 /oldx newx def /oldy newy def} for end }%
\pnode(! \psGetNodeCenter{P@A} \psGetNodeCenter{P@B} P@B.x P@A.x sub  tposmin mul P@A.x add  P@B.y P@A.y sub tposmin  mul P@A.y add ){\nodenameA}%
\pnode(! \psGetNodeCenter{P@A} \psGetNodeCenter{P@B} P@B.x P@A.x sub tnegmax mul P@A.x add P@B.y P@A.y sub tnegmax mul P@A.y add){\nodenameB}%
}%
\def\actualscale#1 #2 scale{
#1}
\def\psGetCenter#1{ tx@NodeDict begin /N@#1 load GetCenter end }
\def\ArrowNotch{\pst@object{ArrowNotch}}
\def\ArrowNotch@i#1#2#3#4{{%
\pst@killglue%
\use@par%
\def\inc{-1}%
\ifx#3<\def\inc{1}\fi
\pstVerb{ 
    1 \psk@arrowinset\space sub \psk@arrowlength\space \psk@arrowsize\space  
    \pst@number\pslinewidth \space mul add  mul mul 
    \expandafter\actualscale\psk@arrowscale \space  mul 
    /hh exch def /hh1 hh .05 sub def }
\def\root@node{#1}\num@pts=\csname\root@node nodecount\endcsname %
\pst@cntb=\num@pts \advance\pst@cntb by \@ne
\pst@cnta=\num@pts \advance\pst@cnta by \thr@@
\pst@cntc=#2 \relax
\ifnum\pst@cntc>\num@pts \pnode(0,0){#4}\else
\pstVerb{%
/PythSq { dup mul exch dup mul add } def
/PtSub {					
  3 -1 roll 		
  sub neg		
  3 1 roll 		
  sub			
  exch                     
} def
  /xarray \the\pst@cnta\space array def
  /yarray \the\pst@cnta\space array def  
  tx@Dict begin }
\multido{\i=0+1,\ib=1+1}{\the\pst@cntb}{\pnode(! \psGetCenter{\root@node\i}\space  
yarray \ib\space 3 -1 roll put xarray \ib\space 3 -1 roll put 0 0 ){@tmp}}
\pnode(! xarray 1 get dup yarray 1 get dup 3 1 roll 
xarray 2 get yarray 2 get PtSub  
2 copy Pyth hh div 2 div dup 
3 1 roll 
div 3 1 roll div 
3 1 roll 
add 3 1 roll add 
 xarray 0 3 -1 roll put yarray 0 3 -1 roll put 
 xarray length 2 sub /topnum exch def 
 xarray topnum get dup yarray topnum get dup 3 1 roll 
topnum 1 sub /topnum exch def xarray topnum get yarray topnum get 
3 -1 roll sub  neg 3 1 roll sub exch 
2 copy Pyth hh div 2 div dup 
3 1 roll div 3 1 roll div 
3 -1 roll add 3 1 roll 
topnum 2 add /topnum exch def xarray topnum 3 -1 roll put yarray topnum 3 -1 roll put 
 /oldcindex \the\pst@cntc\space 1 add def 
 xarray oldcindex get /xc exch def yarray oldcindex get /yc exch def
/inc \inc\space def 
/cindex oldcindex def 
{cindex inc add /cindex exch def xarray cindex get xc sub yarray cindex get yc sub Pyth dup hh1 gt 
{ exit } if } loop 
 hh1 .1 add lt { xarray cindex get yarray cindex get } 
{ xarray cindex inc sub get dup yarray cindex inc sub get dup 4 -1 roll exch 
xarray cindex get yarray cindex get PtSub /dy1 exch def /dx1 exch def dx1 dy1 PythSq /Aterm exch def 
 2 copy xc yc PtSub 
 2 copy 2 copy 3 -1 roll mul 3 1 roll mul add hh dup mul sub 
 Aterm div /Cterm exch def  
 dx1 dy1 
 4 1 roll mul 3 1 roll mul add Aterm div /Bterm exch def 
 Bterm abs neg dup dup mul Cterm sub sqrt add dup /tval exch def
 dup dx1 dy1 4 1 roll mul 3 1 roll mul  
 PtSub } ifelse 
 \pst@number\psyunit div exch \pst@number\psxunit div exch  
){#4}\fi%
\pstVerb { end } 
}\ignorespaces}%
\def\saveDataAsNodes#1#2{
  \psLoopIndex=0\relax
  \typeout{Open file #1}%
  \openin7=#1
  \loop
    \read7 to \@Data
    \ifeof7\else
      \ifx\@Data\@empty
      \else
        \pnode(!\@Data){#2\the\psLoopIndex}%
        \typeout{#2\the\psLoopIndex -> \@Data}%
	\advance\psLoopIndex by 1
        \let\@oldData\@Data
      \fi
  \repeat
  \closein7
  \advance\psLoopIndex by -1
  \pnode(!\@oldData){#2Last}%
}
\catcode`\@=\TheAtCode\relax
 
\csname PSTcoilsLoaded\endcsname
\let\PSTcoilsLoaded 
\ifx\PSTricksLoaded   \else\input pstricks.tex\fi
\ifx\PSTnodeLoaded    \else\fi
\ifx\PSTXKeyLoaded    \else\input pst-xkey \fi
\def\fileversion{1.07}
\def\filedate{2015/05/13}
\edef\TheAtCode{\the\catcode`\@}
\catcode`\@=11
\pst@addfams{pst-coil}
\pstheader{pst-coil.pro}
\edef\pst@theheaders{\pst@theheaders,pst-coil.pro}
\def\pst@CoilDict{tx@CoilDict begin }
\def\tx@CoilLoop  {\pst@CoilDict CoilLoop   end }
\def\tx@Coil      {\pst@CoilDict Coil       end }
\def\tx@AltCoil   {\pst@CoilDict AltCoil    end }
\def\tx@ZigZag    {\pst@CoilDict ZigZag     end }
\def\tx@ZigZagCirc{\pst@CoilDict ZigZagCirc end }
\def\tx@Sin       {\pst@CoilDict Sin        end }
\define@key[psset]{pst-coil}{coilwidth}[1cm]{\pst@getlength{#1}\psk@coilwidth}
\define@key[psset]{pst-coil}{coilheight}[1]{\pst@checknum{#1}\pscoilheight}
\define@key[psset]{pst-coil}{coilarmA}[0.5cm]{\pst@getlength{#1}\psk@coilarmA}
\define@key[psset]{pst-coil}{coilarmB}[0.5cm]{\pst@getlength{#1}\psk@coilarmB}
\define@key[psset]{pst-coil}{coilarm}[0.5cm]{%
  \pst@getlength{#1}\psk@coilarmA%
  \let\psk@coilarmB\psk@coilarmA}
\define@key[psset]{pst-coil}{coilaspect}[45]{\pst@getangle{#1}\psk@coilaspect}
\define@key[psset]{pst-coil}{coilinc}[10]{\pst@getangle{#1}\psk@coilinc}
\psset[pst-coil]{coilaspect=45,coilarm=.5cm,coilheight=1,coilwidth=1cm,coilinc=10}
\def\pscoil{\def\pst@par{}\pst@object{pscoil}}
\def\pscoil@i{\pst@getarrows\pscoil@ii}
\def\pscoil@ii(#1){\@ifnextchar({\pscoil@iii{1}(#1)}{\pscoil@iii{\z@}(0,0)(#1)}}
\def\pscoil@iii#1(#2)(#3){%
  \begin@OpenObj
  \pst@getcoor{#2}\pst@tempa
  \pst@getcoor{#3}\pst@tempb
  \pst@optcp{#1}\pst@tempa
  \addto@pscode{%
    \pst@tempa \pst@tempb
    \psk@coilwidth \pscoilheight
    \psk@coilarmA \psk@coilarmB
    \psk@coilaspect \psk@coilinc
    \tx@Coil }%
    \showpointsfalse
  \end@OpenObj}
\def\psCoil{\def\pst@par{}\pst@object{psCoil}}
\def\psCoil@i#1#2{%
  \begin@AltOpenObj
  \showpointsfalse
  \pst@getangle{#1}\pst@tempa
  \pst@getangle{#2}\pst@tempb
  \addto@pscode{%
    \pst@tempa
    \pst@tempb
    \psk@coilwidth
    \pscoilheight
    \psk@coilaspect
    \psk@coilinc
    \tx@AltCoil  
    \@nameuse{psls@\pslinestyle} }%
  \end@OpenObj}
\define@key[psset]{pst-coil}{bow}[0]{%
  \pst@getlength{#1}\psk@bow%
  \pst@dimm=\psk@bow pt%
  \pst@absdim{\pst@dimm}{\pst@dimn}%
  \ifdim\pst@dimn<1pt \def\psk@bow{0}\fi}%
\psset[pst-coil]{bow=0}
\def\pszigzag{\def\pst@par{}\pst@object{pszigzag}}
\def\pszigzag@i{\pst@getarrows\pszigzag@ii}
\def\pszigzag@ii(#1){\@ifnextchar({\pszigzag@iii{1}(#1)}{\pszigzag@iii{\z@}(0,0)(#1)}}
\def\pszigzag@iii#1(#2)(#3){%
  \addbefore@par{bow=0}%
  \begin@OpenObj%
  \pst@getcoor{#2}\pst@tempA%
  \pst@getcoor{#3}\pst@tempB%
  \pst@optcp{#1}\pst@tempA%
  \addto@pscode{%
    \pst@tempA
    \pst@tempB
    \pscoilheight
    \psk@coilwidth
    \psk@coilarmA
    \psk@coilarmB 
    \ifdim\psk@bow pt=\z@ \tx@ZigZag \else \psk@bow\space \tx@ZigZagCirc \fi
    \psline@iii
    \tx@Line }%
  \end@OpenObj}
\def\nccoil{\pst@object{nccoil}}
\def\nccoil@i{\check@arrow{\nccoil@ii}}
\def\nccoil@ii#1#2{\nc@object{Open}{#1}{#2}{.5}{%
  \tx@NCCoor
  tx@Dict begin
  4 2 roll
  \psk@coilwidth \pscoilheight
  \psk@coilarmA \psk@coilarmB
  \psk@coilaspect \psk@coilinc
  \tx@Coil 
  end }}
\def\pccoil{\def\pst@par{}\pst@object{pccoil}}
\def\pccoil@i{\pc@object\nccoil@ii}
\def\nczigzag{\pst@object{nczigzag}}
\def\nczigzag@i{\check@arrow{\nczigzag@ii}}
\def\nczigzag@ii#1#2{\nc@object{Open}{#1}{#2}{.5}{%
  \tx@NCCoor
  tx@Dict begin
  4 2 roll
  \pscoilheight
  \psk@coilwidth
  \psk@coilarmA
  \psk@coilarmB
  \ifdim\psk@bow pt=\z@\tx@ZigZag\else\psk@bow\space\tx@ZigZagCirc\fi 
  \psline@iii
  \tx@Line
  end }}
\def\pczigzag{\def\pst@par{}\pst@object{pczigzag}}
\def\pczigzag@i{\pc@object\nczigzag@ii}
\def\pst@checkUnit#1#2{\expandafter\pst@checkUnit@i#1!!#2}
\def\pst@checkUnit@i{\@ifnextchar*%
  {\def\pst@roundValue{0 }\pst@checkUnit@ii}%
  {\def\pst@roundValue{-1 }\pst@checkUnit@iii**}}
\def\pst@checkUnit@ii*{\@ifnextchar*%
  {\def\pst@roundValue{1 }\pst@checkUnit@iii*}%
  {\pst@checkUnit@iii**}}
\def\pst@checkUnit@iii**#1!!#2{%
  \edef\ps@next{#1}%
  \ifx\ps@next\@empty\let\pst@num\z@%
  \else\expandafter\pst@@checknum\ps@next..\@nil%
  \fi%
  \ifnum\pst@num=\z@\pst@getlength{#1}{#2}\def\pst@relativePeriod{false }%
  \else%
    \def\pst@relativePeriod{true }%
    \edef#2{\ifnum\pst@num=\tw@-\fi\the\pst@cntg.%
    \expandafter\@gobble\the\pst@cnth\space}%
  \fi}
\define@key[psset]{pst-coil}{periods}[1]{\pst@checkUnit{#1}{\psk@periods}}
\define@key[psset]{pst-coil}{amplitude}[1]{\def\psk@amplitude{#1 }}
\define@key[psset]{pst-coil}{ppoints}[360]{\def\psk@ppoints{#1 }}
\define@key[psset]{pst-coil}{function}[sin]{\def\psk@function{#1 }}
\psset[pst-coil]{periods=1,amplitude=1,ppoints=360,function=sin}
\def\pssin{\pst@object{pssin}}
\def\pssin@i{\pst@getarrows\pssin@ii}
\def\pssin@ii(#1){\@ifnextchar({\pssin@iii{1}(#1)}{\pssin@iii{\z@}(0,0)(#1)}}
\def\pssin@iii#1(#2)(#3){%
  \begin@OpenObj
  \pst@getcoor{#2}\pst@tempa
  \pst@getcoor{#3}\pst@tempb
  \pst@optcp{#1}\pst@tempa
  \addto@pscode{%
    \pst@tempa \pst@tempb
    \psk@periods 
    \pst@relativePeriod 
    \pst@roundValue
    \psk@amplitude \pst@number\psyunit mul
    \psk@coilarmA \psk@coilarmB 
    \psk@ppoints
    { \psk@function }
    \tx@Sin
  }%
  \showpointsfalse%
  \end@OpenObj}
\def\ncsin{\pst@object{ncsin}}
\def\ncsin@i{\check@arrow{\ncsin@ii}}
\def\ncsin@ii#1#2{\nc@object{Open}{#1}{#2}{.5}{%
  \tx@NCCoor
  tx@Dict begin
  4 2 roll
  \psk@periods 
  \pst@relativePeriod 
  \pst@roundValue
  \psk@amplitude \pst@number\psyunit mul
  \psk@coilarmA \psk@coilarmB 
  \psk@ppoints
  { \psk@function }
  \tx@Sin 
  end }}
\def\pcsin{\def\pst@par{}\pst@object{pcsin}}
\def\pcsin@i{\pc@object\ncsin@ii}
\catcode`\@=\TheAtCode\relax

\def\UrlFont{\rmfamily}
\newcommand{\boldphi}{\boldsymbol{\varphi}}
\newcommand{\bgamma}{\boldsymbol{\gamma}}
    \usepackage{latexsym}
\usepackage{subfigure}
\usepackage{amssymb}
\usepackage{url}
\usepackage{amsmath}
\usepackage{amssymb}
\usepackage{psfrag}

\usepackage[section]{placeins}
\usepackage{arydshln}

\def\Ct{{c_\theta}}
\def\St{{s_\theta}}
\def\Cf{{c_\phi}}
\def\Sf{{s_\phi}}
\def\Cp{{c_\psi}}
\def\Sp{{s_\psi}}
\def\atan2{{\mbox{atan2}}}

\newcommand{\eq}[1]{eq.~(\ref{eq:#1})}
\newcommand{\Eq}[1]{Eq.~(\ref{eq:#1})}
\newcommand{\eqn}[1]{(\ref{eq:#1})}
\newcommand{\Eqn}[1]{(\ref{eq:#1})}
\newcommand{\fig}[1] {fig.~\ref{fig:#1}}
\newcommand{\Fig}[1]{Fig.~\ref{fig:#1}}
\newcommand{\fign}[1] {\ref{fig:#1}}
\newcommand{\Fign}[1]{(\ref{fig:#1})}
\newcommand{\bib}[1]{~\cite{#1}}
\newcommand{\sez}[1]{Sez.~\ref{sec:#1}}
\newcommand{\sect}[1]{Sec.~\ref{sec:#1}}
\newcommand{\Sec}[1]{Sec.~\ref{sec:#1}}
\newcommand{\Section}[1]{Section~\ref{sec:#1}}
\newcommand{\ch}[1]{Chap.~\ref{ch:#1}}
\newcommand{\Chapter}[1]{Chapter~\ref{ch:#1}}
\newcommand{\tab}[1] {tab.~\ref{tab:#1}}
\newcommand{\Tab}[1] {Tab.~\ref{tab:#1}}

\newcommand{\bdot}[1] {\dot{\bf #1}}

\newcommand{\crx}[1]{{\bf #1}\times}
\newcommand{\grad}[2]{\frac{\partial #1}{\partial {#2}}}
\newcommand{\hess}[2]{\frac{\partial^2 #1}{\partial{\bf #2}^2}}
\newcommand{\bg}[1]{\mbox{\boldmath $#1$}}
\newcommand{\dt}[1]{\frac{d~#1}{dt}}

\def\fitem{\item[$\Rightarrow~$]}

\def\DEIS{Dipartimento di Elettronica Informatica e Sistemistica}
\def\DIEM{Dipartimento di Meccanica}

\def\12{\frac{1}{2}}

\def\a{\mbox{\boldmath $a$}}  \def\b{\mbox{\boldmath $b$}}
\def\c{\mbox{\boldmath $c$}} \def\d{\mbox{\boldmath $d$}}
\def\e{\mbox{\boldmath $e$}} \def\f{\mbox{\boldmath $f$}} \def\g{\mbox{\boldmath $g$}} \def\h{\mbox{\boldmath $h$}}
\def\ib{\mbox{\boldmath $i$}} \def\j{\mbox{\boldmath $j$}} \def\k{\mbox{\boldmath $k$}} \def\l{\mbox{\boldmath $l$}}
\def\m{\mbox{\boldmath $m$}} \def\n{\mbox{\boldmath $n$}} \def\o{\mbox{\boldmath $o$}} \def\p{\mbox{\boldmath $p$}}
\def\q{\mbox{\boldmath $q$}} \def\r{\mbox{\boldmath $r$}} \def\s{\mbox{\boldmath $s$}} \def\t{\mbox{\boldmath $t$}}
\def\u{\mbox{\boldmath $u$}} \def\v{\mbox{\boldmath $v$}} \def\w{\mbox{\boldmath $w$}} \def\x{\mbox{\boldmath $x$}}
\def\y{\mbox{\boldmath $y$}} \def\z{\mbox{\boldmath $z$}}

\def\A{\mbox{\boldmath $A$}} \def\B{\mbox{\boldmath $B$}} \def\C{\mbox{\boldmath $C$}} \def\D{\mbox{\boldmath $D$}}
\def\E{\mbox{\boldmath $E$}} \def\F{\mbox{\boldmath $F$}} \def\G{\mbox{\boldmath $G$}} \def\H{\mbox{\boldmath $H$}}
\def\I{\mbox{\boldmath $I$}} \def\J{\mbox{\boldmath $J$}} \def\K{\mbox{\boldmath $K$}} \def\L{\mbox{\boldmath $L$}}
\def\M{\mbox{\boldmath $M$}} \def\N{\mbox{\boldmath $N$}} \def\O{\mbox{\boldmath $O$}} \def\P{\mbox{\boldmath $P$}}
\def\Q{\mbox{\boldmath $Q$}} \def\R{\mbox{\boldmath $R$}} \def\S{\mbox{\boldmath $S$}} \def\T{\mbox{\boldmath $T$}}
\def\U{\mbox{\boldmath $U$}} \def\V{\mbox{\boldmath $V$}} \def\W{\mbox{\boldmath $W$}} \def\X{\mbox{\boldmath $X$}}
\def\Y{\mbox{\boldmath $Y$}} \def\Z{\mbox{\boldmath $Z$}}

 \def\udot{\dot{\bf u}} \def\xdot{\dot{\bf x}}
\def\ydot{\dot{\bf y}} \def\ccdot{\dot{\bf c}} \def\pdot{\dot{\bf p}}
\def\vdot{\dot{\bf v}} \def\rdot{\dot{\bf r}}  \def\bddot{\dot{\bf d}}

 \def\dotx{\dot{x}}

\def\0{{\bf 0}}

\def\cA{{\cal A}}
\def\cB{{\cal B}}
\def\cC{{\cal C}}
\def\cD{{\cal D}}
\def\cE{{\cal E}}
\def\cM{{\cal M}}
\def\cP{{\cal P}}
\def\cQ{{\cal Q}}
\def\cR{{\cal R}}
\def\cU{{\cal U}}
\def\cT{{\cal T}}
\def\cV{{\cal V}}
\def\cW{{\cal W}}
\def\cX{{\cal X}}
\def\cY{{\cal Y}}

 \def\btau{{\mbox{\boldmath $\tau$}}}
 \def\bom{{\mbox{\boldmath $\omega$}}}
 \def\bomega{{\mbox{\boldmath $\omega$}}}
 \def\bphi{{\mbox{\boldmath $\phi$}}}
 \def\bpsi{{\mbox{\boldmath $\psi$}}}
 \def\balpha{{\mbox{\boldmath $\alpha$}}}
 \def\bbeta{{\mbox{\boldmath $\beta$}}}
 \def\bupsilon{{\mbox{\boldmath $\upsilon$}}}
 \def\bnu{{\mbox{\boldmath $\nu$}}}
 \def\b0{{\mbox{\boldmath $0$}}}

 \def\bA{{\mbox{\boldmath $A$}}}
 \def\bB{{\mbox{\boldmath $B$}}}
 \def\oB{\overline{\mbox{{\boldmath $B$}}}}
 \def\bC{{\mbox{\boldmath $C$}}}
 \def\oC{\overline{\mbox{{\boldmath $C$}}}}
 \def\bD{{\mbox{\boldmath $D$}}}
 \def\bI{{\mbox{\boldmath $I$}}}
 \def\bK{{\mbox{\boldmath $K$}}}
 \def\bL{{\mbox{\boldmath $L$}}}
 \def\bM{{\mbox{\boldmath $M$}}}
 \def\bP{{\mbox{\boldmath $P$}}}
 \def\oP{\ \overline{\mbox{{\boldmath $P$}}}}
 \def\bR{{\mbox{\boldmath $R$}}}
 \def\bO{{\mbox{\boldmath $0$}}}
 \def\oQ{\overline{\mbox{{\boldmath $Q$}}}}
 \def\bW{{\mbox{\boldmath $W$}}}
 \def\oW{\overline{\mbox{{\boldmath $W$}}}}
 \def\bX{{\mbox{\boldmath $X$}}}
 \def\bb{{\mbox{\boldmath $b$}}}
 \def\bet{{\mbox{\scriptsize \boldmath $e$}}}
 \def\bg{{\mbox{\boldmath $g$}}}
 \def\bm{{\mbox{\boldmath $m$}}}
 \def\bp{{\mbox{\boldmath $p$}}}
 \def\bps{{\mbox{\scriptsize \boldmath $p$}}}
 \def\bpt{{\mbox{\tiny \boldmath $p$}}}
 \def\bq{{\mbox{\boldmath $q$}}}
 \def\bqt{{\mbox{\tiny \boldmath $q$}}}
 \def\bs{{\mbox{\boldmath $s$}}}
 \def\bt{{\mbox{\boldmath $t$}}}
 \def\bu{{\mbox{\boldmath $u$}}}
 \def\bx{{\mbox{\boldmath $x$}}}
 \def\bz{{\mbox{\boldmath $z$}}}

 \def\qdot{\dot{\bq}}

\newcommand{\Rm}{{\rm R\hspace*{-0.9ex}\rule{0.15ex}%
{1.5ex}\hspace*{0.9ex}}}
\newcommand{\Nm}{{\rm N\hspace*{-1.0ex}\rule{0.15ex}%
{1.3ex}\hspace*{1.0ex}}}
\def\natural{\mbox{\rm I\kern-0.2em N}}  
\newcommand{\Qm}{{\rm Q\hspace*{-1.1ex}\rule{0.15ex}%
{1.5ex}\hspace*{1.1ex}}}
\newcommand{\Cm}{{\sf C\hspace*{-0.9ex}\rule{0.15ex}%
       {1.3ex}\hspace*{0.9ex}}}

\def\diag{\mbox{\sf diag $\;$}}
\def\real{\mbox{\rm I\kern-0.2em R}}  
\def\=def{\stackrel{def}{=}}
\def\wh{\noindent}
\def\ni{\noindent}
\def\nis{\hspace*{-1.5mm}}
\def\beq{\begin{equation}}
\def\eeq{\end{equation}}
\def\be{\begin{equation}}
\def\ee{\end{equation}}
\def\bea{\begin{eqnarray}}
\def\eea{\end{eqnarray}}
\def\beann{\begin{eqnarray*}}
\def\eeann{\end{eqnarray*}}

\def\ds{\displaystyle}
\def\sbf{\bf \sf}
\def\sit{\it \sf}
\newcommand{\fracd}[2]{\frac{\ds #1}{\ds #2}}

\def\vsp{\vspace*{5mm}}
\def\vspi{\vspace*{5mm}}
\def\vspii{\vspace*{10mm}}
\def\vspiii{\vspace*{15mm}}
\def\vspiiii{\vspace*{20mm}}
\def\hsp{\hspace*{5mm}}
\def\hspi{\hspace*{5mm}}
\def\hspii{\hspace*{10mm}}
\def\hspiii{\hspace*{15mm}}
\def\hspiiii{\hspace*{20mm}}

\def\ra{\rightarrow}
\def\Ra{\Rightarrow}
\def\Lra{\Longrightarrow}

\def\la{\leftarrow}
\def\La{\Leftarrow}
\def\Lla{\Longleftarrow}

\def\lra{\Leftrightarrow}
\def\Llra{\Longleftrightarrow}

  {\setlength{\arraycolsep}{0.14em} 
   \begin{eqnarray}}
  {\end{eqnarray}}           

\def\Beginboxit
   {\par
    \vbox\bgroup
           \hrule height1pt
           \hbox\bgroup
                  \vrule width1pt \kern2pt %
                  \vbox\bgroup\kern2pt
   }
\def\Endboxit{%
                             \kern2pt
                             \vspace*{4pt}
                       \egroup
                  \kern2pt\vrule width1pt
                \egroup
           \hrule height1pt
         \egroup
   }

\newenvironment{boxit}{\Beginboxit}{\Endboxit}
\newenvironment{boxit*}{\Beginboxit\hbox to\hsize{}}{\Endboxit}

\newcommand{\cbox}[1]{
                        \begin{boxit}
                        #1
                        \end{boxit}
                        }

\def\WDT{0.47\textwidth}
\def\HWDT{0.25\textwidth}
\def\BWDT{0.7\textwidth}

\def\ds{\displaystyle}

 \newcommand{\preamble}[1]{
 \begin{center}
 \begin{minipage}{0.8\textwidth}
 {\small \sf
 #1

 \underline{\hspace{\textwidth}}
 \vspace{2mm}
 }
 \end{minipage}
 \end{center}
 }

 \newcommand{\Esempio}[1]{
 \begin{esemp}
 {\rm  #1 }
 \hspace*{\fill} $\Box$ \\
 \end{esemp}
 }

 \newcommand{\putab}{\centerline{
     \parbox[c]{0.5\textwidth}{\hspace{0.25\textwidth}\small (a)}
     \parbox[c]{0.5\textwidth}{\hspace{0.25\textwidth}\small (b)}}}

 \newcommand{\puttimed}{\vspace{-1.5mm}\centerline{
     \parbox[c]{0.5\textwidth}{\hspace{0.25\textwidth}\small t}
     \parbox[c]{0.5\textwidth}{\hspace{0.25\textwidth}\small t}}}

 \newif\ifSmallFigure \SmallFiguretrue
 \ifSmallFigure
    \def\FIGT2{c:/usr/claudio/work/ucima/tlibro/tfigure1} 
 \else
    \def\FIGT2{c:/usr/claudio/work/ucima/tlibro/tfigure} 
 \fi

\def\sign{\mathrm{sign}}

\def\vq{{\bq}}

\def\yq{{y}}
\def\dotq{\dot{q}}
\def\ddotq{\ddot{q}}
\def\dddotq{q^{(3)}}
\def\ddddotq{q^{(4)}}

\def\ac{\ddotq}
\def\j{\dddotq}
\def\s{\ddddotq}

\def\t{t}
\def\tzero{t_0}
\def\tuno{t_1}

\def\qzero{q_0}
\def\quno{q_1}

\def\vel{\texttt{v}}
\def\vzero{\vel_0}
\def\vuno{\vel_1}
\def\vmax{\vel_{max}}
\def\vmin{\vel_{min}}
\def\hvmax{\hat \vel_{max}}
\def\hvmin{\hat \vel_{min}}

\def\deltav{\Delta V}
\def\deltaq{h}

\def\acc{\texttt{a}}
\def\azero{\acc_{0}}
\def\auno{\acc_{1}}
\def\amax{\acc_{max}}
\def\amin{\acc_{min}}
\def\hamax{\hat\acc_{max}}
\def\hamin{\hat\acc_{min}}

\def\jerk{\texttt{j}}
 \def\snap{\texttt{s}}
\def\jzero{\jerk_{0}}
\def\juno{\jerk_{1}}
\def\jmax{\jerk_{max}}
\def\jmin{\jerk_{min}}
\def\hjmax{\hat \jerk_{max}}
\def\hjmin{\hat \jerk_{min}}
\def\szero{\snap_{0}}
\def\suno{\snap_{1}}

\def\smax{\snap_{max}}
\def\smin{\snap_{min}}

\def\alfa{a}

\newcommand{\blambda}{\mbox{\boldmath$\lambda$}}
\newcommand{\bdelta}{\mbox{\boldmath$\delta$}}

\def\qd{\q_k}
\def\vd{\dotq_k}
\def\acd{\ddotq_k}
\def\jd{\j_k}
\def\sd{\s_k}

\def\qdm1{\q_{k-1}}
\def\vdm1{\dotq_{k-1}}
\def\acdm1{\ac_{k-1}}
\def\jdm1{\j_{k-1}}

\def\hq{\hat q}
\def\hv{\hat \vel}
\def\hac{\hat \acc}

\def\hdotq{\dot{\hat{q}}}
\def\hddotq{\ddot{\hat{q}}}
\def\hdddotq{{\hat{q}}^{(3)}}

\def\im{{\rm{j}}}

 \def\qn{{q _{\mbox{\tiny{$N$}}}}}
 \def\tqn{\tilde q _{\mbox{\tiny{$N$}}}}
 \newcommand{\qnpow}[1]{q ^{#1}_{\mbox{\tiny{$N$}}}}
 \newcommand{\tqnpow}[1]{\tilde q ^{#1}_{\mbox{\tiny{$N$}}}}
 \def\dqn{\dot q _{\mbox{\tiny{$N$}}}}
 \def\ddqn{\ddot q _{\mbox{\tiny{$N$}}}}

 \def\Vn{V_{\mbox{\tiny{$N$}}}}

\newcommand{\tras}{^{\mbox{\tiny T}}}
\newcommand{\mtras}{^{\mbox{\tiny -T}}}
\newcommand{\dtras}{^{\mbox{\tiny T}\dag}}
\newcommand{\muno}{^{\mbox{\tiny -1}}}
\newcommand{\tracon}{^{*}}
\newcommand{\mtracon}{^{-*}}
\newcommand{\ts}{\textstyle}
\newcommand{\scr}{\scriptstyle}
\newcommand{\sscr}{\scriptscriptstyle}
\newcommand{\bbar}[1]{\bar{\bar{#1}}}

 \newcommand{\evidenzia}[1]{\psframebox[framesep=5pt]{#1}}
 \newcommand{\BoxedEPSF}{\epsfbox} 
 \newcommand{\bo}{\mbox{\bf o}}
 \newcommand{\und}{\underline}
 \newcommand{\Span}{\mbox{\rm span}}
 \newcommand{\spann}{\mbox{\rm span}}
 \newcommand{\Agg}{\mbox{\rm agg}}
 \newcommand{\adj}{\mbox{\rm adj}}
 \newcommand{\Imm}{\mbox{\rm Im}}
 \newcommand{\Ker}{\mbox{\rm ker}}
 \newcommand{\rango}{\mbox{\rm rango}}
 \newcommand{\mat}[2]{\left[\begin{array}{#1} #2 \end{array}\right]}
 \newcommand{\mdet}[2]{\left|\begin{array}{#1} #2 \end{array}\right|}
 \newcommand{\cofbin}[2]{\left(\!\begin{array}{c}#1\\ #2 \end{array}\!\right)}

\def\a{{\bf a}} \def\b{{\bf b}} \def\c{{\bf c}} \def\d{{\bf d}}
\def\e{{\bf e}} \def\f{{\bf f}} \def\g{{\bf g}} \def\h{{\bf h}}
\def\ib{{\bf i}} \def\j{{\bf j}} \def\k{{\bf k}} \def\l{{\bf l}}
\def\m{{\bf m}} \def\n{{\bf n}} \def\o{{\bf o}} \def\p{{\bf p}}
\def\q{{\bf q}} \def\r{{\bf r}} \def\s{{\bf s}} \def\t{{\bf t}}
\def\u{{\bf u}} \def\v{{\bf v}} \def\w{{\bf w}} \def\x{{\bf x}}
\def\y{{\bf y}} \def\z{{\bf z}}

\def\A{{\bf A}} \def\B{{\bf B}} \def\C{{\bf C}} \def\D{{\bf D}}
\def\E{{\bf E}} \def\F{{\bf F}} \def\G{{\bf G}} \def\H{{\bf H}}
\def\I{{\bf I}} \def\J{{\bf J}} \def\K{{\bf K}} \def\L{{\bf L}}
\def\M{{\bf M}} \def\N{{\bf N}} \def\O{{\bf O}} \def\P{{\bf P}}
\def\Q{{\bf Q}} \def\R{{\bf R}} \def\S{{\bf S}} \def\T{{\bf T}}
\def\U{{\bf U}} \def\V{{\bf V}} \def\W{{\bf W}} \def\X{{\bf X}}
\def\Y{{\bf Y}} \def\Z{{\bf Z}}

\def\cA{{\cal A}} \def\cB{{\cal B}} \def\cC{{\cal C}} \def\cD{{\cal D}}
\def\cE{{\cal E}} \def\cF{{\cal F}} \def\cG{{\cal G}} \def\cH{{\cal H}}
\def\cI{{\cal I}} \def\cJ{{\cal J}} \def\cK{{\cal K}} \def\cL{{\cal L}}
\def\cM{{\cal M}} \def\cN{{\cal N}} \def\cO{{\cal O}} \def\cP{{\cal P}}
\def\cQ{{\cal Q}} \def\cR{{\cal R}} \def\cS{{\cal S}} \def\cT{{\cal T}}
\def\cU{{\cal U}} \def\cV{{\cal V}} \def\cW{{\cal W}} \def\cX{{\cal X}}
\def\cY{{\cal Y}} \def\cZ{{\cal Z}}

\def\ta{\tilde{\a}} \def\tb{\tilde{\b}} \def\tc{\tilde{\c}}
\def\td{\tilde{\d}} \def\te{\tilde{\e}} \def\tf{\tilde{\f}}
\def\tg{\tilde{\g}} \def\th{\tilde{\h}} \def\tib{\tilde{\i}}
\def\tj{\tilde{\j}} \def\tk{\tilde{\k}} \def\tl{\tilde{\l}}
\def\tm{\tilde{\m}} \def\tn{\tilde{\n}} \def\to{\tilde{\o}}
\def\tp{\tilde{\p}} \def\tq{\tilde{\q}} \def\tr{\tilde{\r}}
\def\tts{\tilde{\s}} \def\ttt{\tilde{\t}} \def\tu{\tilde{\u}}
\def\tv{\tilde{\v}} \def\tw{\tilde{\w}} \def\tx{\tilde{\x}}
\def\ty{\tilde{\y}} \def\tz{\tilde{\z}}

\def\tA{\tilde{\A}} \def\tB{\tilde{\B}} \def\tC{\tilde{\C}}
\def\tD{\tilde{\D}} \def\tE{\tilde{\E}} \def\tF{\tilde{\F}}
\def\tG{\tilde{\G}} \def\tH{\tilde{\H}} \def\tI{\tilde{\I}}
\def\tJ{\tilde{\J}} \def\tK{\tilde{\K}} \def\tL{\tilde{\L}}
\def\tM{\tilde{\M}} \def\tN{\tilde{\N}} \def\tO{\tilde{\O}}
\def\tP{\tilde{\P}} \def\tQ{\tilde{\Q}} \def\tR{\tilde{\R}}
\def\tS{\tilde{\S}} \def\tT{\tilde{\T}} \def\tU{\tilde{\U}}
\def\tV{\tilde{\V}} \def\tW{\tilde{\W}} \def\tX{\tilde{\X}}
\def\tY{\tilde{\Y}} \def\tZ{\tilde{\Z}}

\def\ooa{\overline{\a}} \def\oob{\overline{\b}}
\def\ooc{\overline{\c}} \def\ood{\overline{\d}}
\def\ooe{\overline{\e}} \def\oof{\overline{\f}}
\def\oog{\overline{\g}} \def\ooh{\overline{\h}}
\def\ooib{\overline{\ib}} \def\ooj{\overline{\j}}
\def\ook{\overline{\k}} \def\ool{\overline{\l}}
\def\oom{\overline{\m}} \def\oon{\overline{\n}}
\def\ooo{\overline{\o}} \def\oop{\overline{\p}}
\def\ooq{\overline{\q}} \def\oor{\overline{\r}}
\def\oos{\overline{\s}} \def\oot{\overline{\t}}
\def\oou{\overline{\u}} \def\oov{\overline{\v}}
\def\oow{\overline{\w}} \def\oox{\overline{\x}}
\def\ooy{\overline{\y}} \def\ooz{\overline{\z}}

\def\ooA{\overline{\A}} \def\ooB{\overline{\B}}
\def\ooC{\overline{\C}} \def\ooD{\overline{\D}}
\def\ooE{\overline{\E}} \def\ooF{\overline{\F}}
\def\ooG{\overline{\G}} \def\ooH{\overline{\H}}
\def\ooI{\overline{\I}} \def\ooJ{\overline{\J}}
\def\ooK{\overline{\K}} \def\ooL{\overline{\L}}
\def\ooM{\overline{\M}} \def\ooN{\overline{\N}}
\def\ooO{\overline{\O}} \def\ooP{\overline{\P}}
\def\ooQ{\overline{\Q}} \def\ooR{\overline{\R}}
\def\ooS{\overline{\S}} \def\ooT{\overline{\T}}
\def\ooU{\overline{\U}} \def\ooV{\overline{\V}}
\def\ooW{\overline{\W}} \def\ooX{\overline{\X}}
\def\ooY{\overline{\Y}} \def\ooZ{\overline{\Z}}

\def\va{\vec{\a}} \def\vb{\vec{\b}} \def\vc{\vec{\c}}
\def\vd{\vec{\d}} \def\ve{\vec{\e}} \def\vf{\vec{\f}}
\def\vg{\vec{\g}} \def\vh{\vec{\h}} \def\vib{\vec{\i}}
\def\vj{\vec{\j}} \def\vk{\vec{\k}} \def\vl{\vec{\l}}
\def\vm{\vec{\m}} \def\vn{\vec{\n}} \def\vo{\vec{\o}}
\def\vp{\vec{\p}} \def\vq{\vec{\q}} \def\vr{\vec{\r}}
\def\vts{\vec{\s}} \def\vtt{\vec{\t}} \def\vu{\vec{\u}}
\def\vv{\vec{\v}} \def\vw{\vec{\w}} \def\vx{\vec{\x}}
\def\vy{\vec{\y}} \def\vz{\vec{\z}}

\def\vA{\vec{\A}} \def\vB{\vec{\B}} \def\vC{\vec{\C}}
\def\vD{\vec{\D}} \def\vE{\vec{\E}} \def\vF{\vec{\F}}
\def\vG{\vec{\G}} \def\vH{\vec{\H}} \def\vI{\vec{\I}}
\def\vJ{\vec{\J}} \def\vK{\vec{\K}} \def\vL{\vec{\L}}
\def\vM{\vec{\M}} \def\vN{\vec{\N}} \def\vO{\vec{\O}}
\def\vP{\vec{\P}} \def\vQ{\vec{\Q}} \def\vR{\vec{\R}}
\def\vS{\vec{\S}} \def\vT{\vec{\T}} \def\vU{\vec{\U}}
\def\vV{\vec{\V}} \def\vW{\vec{\W}} \def\vX{\vec{\X}}
\def\vY{\vec{\Y}} \def\vZ{\vec{\Z}}

\def\ha{\hat{\a}} \def\hb{\hat{\b}} \def\hc{\hat{\c}}
\def\hd{\hat{\d}} \def\he{\hat{\e}} \def\hf{\hat{\f}}
\def\hg{\hat{\g}} \def\hh{\hat{\h}} \def\hib{\hat{\i}}
\def\hj{\hat{\j}} \def\hk{\hat{\k}} \def\hl{\hat{\l}}
\def\hm{\hat{\m}} \def\hn{\hat{\n}} \def\ho{\hat{\o}}
\def\hp{\hat{\p}} \def\hq{\hat{\q}} \def\hr{\hat{\r}}
\def\hts{\hat{\s}} \def\htt{\hat{\t}} \def\hu{\hat{\u}}
\def\hv{\hat{\v}} \def\hw{\hat{\w}} \def\hx{\hat{\x}}
\def\hy{\hat{\y}} \def\hz{\hat{\z}}

\def\hA{\hat{\A}} \def\hB{\hat{\B}} \def\hC{\hat{\C}}
\def\hD{\hat{\D}} \def\hE{\hat{\E}} \def\hF{\hat{\F}}
\def\hG{\hat{\G}} \def\hH{\hat{\H}} \def\hI{\hat{\I}}
\def\hJ{\hat{\J}} \def\hK{\hat{\K}} \def\hL{\hat{\L}}
\def\hM{\hat{\M}} \def\hN{\hat{\N}} \def\hO{\hat{\O}}
\def\hP{\hat{\P}} \def\hQ{\hat{\Q}} \def\hR{\hat{\R}}
\def\hS{\hat{\S}} \def\hT{\hat{\T}} \def\hU{\hat{\U}}
\def\hV{\hat{\V}} \def\hW{\hat{\W}} \def\hX{\hat{\X}}
\def\hY{\hat{\Y}} \def\hZ{\hat{\Z}}

\graphicspath{{./img}}

\newcommand{\ShortFig}{}

 \newtheorem{Remar}{\bf Remark}

 \newtheorem{Optimi}{\bf Optimization Problem}

\title{\LARGE \bf
On the Analysis of Stability, Sensitivity and Transparency in Variable Admittance Control for pHRI Enhanced by Virtual Fixtures
}

\author{Davide Tebaldi\orcidlink{0000-0003-1432-0489}, Dario Onfiani\orcidlink{0000-0001-6734-1094}, and Luigi Biagiotti\orcidlink{0000-0002-2343-6929}
\thanks{The authors are with the Department of Engineering ``Enzo Ferrari'', University of Modena and Reggio Emilia, via Pietro Vivarelli 10, 41125 Modena, Italy, e-mail: \{davide.tebaldi, dario.onfiani, luigi.biagiotti\}@unimore.it.}
\thanks{The work was partly supported by the University of Modena and Reggio Emilia
through the action FARD (Finanziamento Ateneo Ricerca Dipartimentale) 2022/2023 and 2023/2024, and funded under the National Recovery and Resilience Plan (NRRP), Mission 04 Component 2 Investment 1.5 – NextGenerationEU, Call for tender n. 3277 dated 30/12/2021 Award Number:  0001052 dated 23/06/2022.}
\thanks{This work has been submitted to the IEEE for possible publication. Copyright may be transferred without notice, after which this version may no longer be accessible.}
}

\begin{document}

\maketitle
\thispagestyle{empty}
\pagestyle{empty}

\begin{abstract}
The interest in Physical Human-Robot Interaction (pHRI) has significantly increased over the last two decades thanks to the availability of collaborative robots that guarantee user safety during force exchanges. For this reason, stability concerns have been addressed extensively in the literature while proposing new control schemes for pHRI applications. Because of the nonlinear nature of robots, stability analyses generally leverage passivity concepts. On the other hand, the proposed algorithms generally consider ideal models of robot manipulators.
For this reason, the primary objective of this paper is to conduct a detailed analysis of the sources of instability for a class of pHRI control schemes, namely proxy-based constrained admittance controllers, by considering parasitic effects such as transmission elasticity, motor velocity saturation, and actuation delay. Next, a sensitivity analysis supported by experimental results is carried out, in order to identify how the control parameters affect the stability of the overall system. Finally, an adaptation technique for the proxy parameters is proposed with the goal of maximizing transparency in pHRI. The proposed adaptation method is validated through both simulations and experimental tests.
\end{abstract}

\section{Introduction}

The increasing focus on physical human–robot interaction (pHRI) and collaborative robotics has 
enabled applications such as
assisted industrial manipulation, collaborative assembly, domestic tasks, rehabilitation, and medical interventions.

To achieve reliable pHRI for human safety, control strategies that ensure robot compliance have been extensively developed \cite{hogan1984impedance}. Among these, admittance control \cite{keemink2018admittance} has been particularly appealing in many practical scenarios.
Despite its advantages, the implementation of admittance control presents critical challenges regarding system stability~\cite{peer2008robust}.

In this paper, we consider admittance control for pHRI enhanced by virtual fixtures \cite{rosenberg1993use}, which constrain the user's motion along a specific path \cite{selvaggio2018passive}, thereby reducing both the cognitive and physical efforts.
\begin{figure}[t]
\centering
\psfrag{f}[c][c][1][0]{ $\black{\F_{h}}$}
\psfrag{F}[c][c][1][0]{ $\black{\F_{\bot}}$}
\psfrag{P}[c][c][1][0]{ $\boldphi(l)$}
\psfrag{S}[t][t][1.25][0]{$m, b$}
\psfrag{T}[c][c][1][0]{ $\black{F_\tau}$}
\psfrag{p}[c][c][0.8][0]{ $\K_{P}$}
\psfrag{d}[c][c][1][0.8]{ $\K_{D}$}
    {\includegraphics[width=0.635\columnwidth]{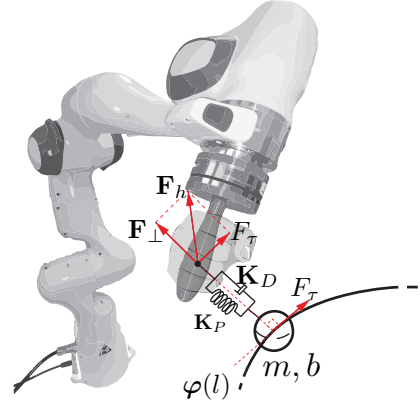}}
\caption{ 
Equivalent mechanical model of a
Proxy-based virtual fixture  in pHRI using a Franka-Emika collaborative robot.}
\label{fig:FrankaTrj2}
\vspace{-2.6mm}
\end{figure}
 In particular, a proxy-based virtual fixture definition is taken into account, as shown in Fig.~\ref{fig:FrankaTrj2}. The human operator, connected to the flange of the robot (in the picture via a simple handle), interacts with the proxy dynamics, which typically follows that of a mass-damper system whose motion is constrained allong a desired curve $\varphi(\cdot)$. 
The proxy position is then used as a reference signal for the robot, which is typically controlled under inverse dynamics control \cite{SicilianoBook}. Notably, many recent papers exploit this control paradigm to guide human users in various activities, ranging from limb rehabilitation to the precise operation of a tool along a predetermined route \cite{9562059,braglia2025phaseindependentdynamicmovementprimitives,onfiani2024optimizingdesigncontrolmethods,shahriari2024path}. Despite rigorous stability proofs, these types of control schemes also suffer from the same stability issues as standard admittance control.
These phenomena manifest as deviations from the position defined by the proxy and high-frequency vibrations of the robot end-effector, compromising both user safety and system performance \cite{Surdilovic}.
The causes are improper tuning of the admittance parameters \cite{peer2008robust} - the mass $m$ and the damping coefficient $b$ of the proxy in this case - and system delays due to control discretizazion and human reactions \cite{Tsumugiwa2002}.
Despite the several works available in the literature, discussed in Sec.~\ref{related_works_sect}, we identified some research gaps that, to the best of our knowledge, have not been thoroughly explored, as discussed in Sec.~\ref{sec:meth_contr}. 

The remainder of this paper is organized as follows. The related works and the contributions are discussed in Sec.~\ref{related_works_sect} and Sec.~\ref{sec:meth_contr}, while the robotic system is modeled in Sec.~\ref{Experimental_Setup_sect}. The stability analysis is performed in  Sec.~\ref{sec:Section_One}, while the admittance parameters adaptation approach is proposed in Sec.~\ref{sec:Section_STransparency}. Finally, the conclusions are given in Sec.~\ref{sec:Conclusions}.


\section{Related Works}\label{related_works_sect}

The stability problem in pHRI has been extensively studied in the literature, and various approaches for preventing or counteracting instability phenomena (generally based on the adaptation of admittance parameters) have been proposed.
In the following subsections, we consider the stability analysis and the approaches for stability recovery/guarantee.   

\subsection{Stability Analysis}\label{Stability_Analysis_sect}
The stability analysis of admittance-based pHRI frameworks has been conducted using various tools, which generally belong to linear control theory \cite{podobnik2007haptic,dimeas2016,gallagher2014improved,wang2023variable}. The typical approach consists of evaluating the position of the system poles, as in \cite{dimeas2016} where the root locus is used.
In this case, oscillatory phenomena are attributed to the resonant frequency of the dominant poles as the stiffness of the environment interacting with the robot increases. 
The closed-loop poles and zeros of the system are also analyzed in \cite{gallagher2014improved} by varying user stiffness, demonstrating the destabilizing effect of increased stiffness. 
A stability analysis was conducted in \cite{wang2023variable} using the Routh-Hurwitz criterion on a third-order human-robot interaction (HRI) model, taking into account human arm stiffness and hardware delay. The derived conditions indicate that increasing virtual damping enhances stability, while higher arm stiffness reduces the stable region.
In general, an overview of the relevant literature reveals that:

\noindent \mbox{
\pscircle[fillstyle=solid,fillcolor=black,linecolor=black](0.05,0.1){0.05}\hspace{0.68mm}} Nonidealities such as joint elasticity and system delays are often neglected and generally not considered together.

\noindent \mbox{
\pscircle[fillstyle=solid,fillcolor=black,linecolor=black](0.05,0.1){0.05}\hspace{0.68mm}} When considered, the delay is approximated by a rational transfer function, using Padé or similar approximations.

\noindent \mbox{
\pscircle[fillstyle=solid,fillcolor=black,linecolor=black](0.05,0.1){0.05}\hspace{0.68mm}} Nonlinear effects, such as actuator saturation, are not considered, as they cannot be addressed using linear control theory tools, especially if based on pole placement analysis.

When these parasitic effects are considered, the use of more complex tools, such as Lyapunov theory, is less common. In \cite{duchaine2008investigation}, the Lyapunov method is used to derive the critical damping and mass parameters analytically, but a linear approximation of the system is still considered. Although the results show a good match between the computed critical damping and experimental values, effects such as actuator saturation and joint elasticity are neglected. In \cite{shahriari2024path}, a passivity-based analysis is instead adopted to ensure overall system stability in a path-constrained haptic guidance scenario, introducing a virtual energy tank that limits potential passivity-violating conditions during parameter adaptation.

\subsection{Admittance Parameters Adaptation}\label{Admittance_Parameters_Adaptation_sect}

The methods used to prevent or mitigate unstable behaviors are primarily based on the adaptation of admittance parameters. Three main mechanisms can be identified in the literature: i) mass adaptation, ii) damping adaptation, and iii) joint mass/damping adaptation.
As shown in Table~\ref{tab:adaptation_classification}, only a limited number of the analyzed studies consider mass adaptation, while the majority focus on either damping adaptation or combined mass and damping adaptation.

The effects of the three different adaptation techniques were compared in \cite{dimeas2016}, where it was concluded that adjusting the mass while keeping the damping constant provides the best compromise between ensuring stability and minimizing human effort. A similar conclusion was reached in \cite{podobnik2007haptic}.

By considering the methods of class ii), the approach proposed in \cite{ryu2008frequency} increases damping proportionally to the detected level of instability. Since humans tend to increase muscle activation when performing precise tasks, \cite{grafakos2016variable} proposes a damping adaptation strategy to enhance accuracy. The variable admittance adaptation in \cite{duchaine2012stable} demonstrates that adjusting virtual damping based on human intention, while keeping virtual mass constant, provides the best compromise between ensuring stability and minimizing human effort compared to adjusting virtual mass. In \cite{erden2011assisting}, a robot-assisted welding application is proposed, where virtual damping is adjusted based on the task phase and speed while keeping virtual mass constant at a fixed value to prevent instability.

Among the works of  type iii), the simultaneous adaptation of both mass and damping is generally performed while maintaining a constant ratio \cite{dimeas2016, wang2023variable, lecours2012variable, ferraguti2019variable, topini2022variable}. In \cite{ferraguti2019variable}, this approach is motivated by the need to preserve a consistent system dynamics, resulting in a more intuitive behavior for the human user \cite{lecours2012variable}. In \cite{gallagher2014improved}, two different sets of parameters are applied based on the stiffness level of the user's arm, measured using EMG sensors: low mass and damping for low stiffness, and high mass and damping for high stiffness to counteract oscillations.
In \cite{okunev2012human}, a machine learning algorithm is employed to adapt the admittance parameters based on the intensity of oscillations, which is measured through FFT analysis.
A fuzzy logic controller is used in \cite{wang2023variable} to adjust mass and damping proportionally based on the force and velocity applied by the human, with the ultimate goal of enhancing stability and reducing user effort.
A similar approach is followed in \cite{topini2022variable}, where damping is decreased during acceleration and increased during deceleration, according to a heuristic model inspired by \cite{lecours2012variable}, while the virtual mass is updated proportionally.

What is clear from the aforementioned discussion is that, out of the three main methods for adapting admittance parameters used in the literature, there is no unified consensus on the best approach to ensure stability in pHRI applications.

\begin{table}[t]

    \caption{Related works classification based on admittance parameters adaptation strategy.}
    \label{tab:adaptation_classification}

    \begin{tabular}{ccc}
        \hline
        \textbf{Mass} & \textbf{Damping} & \textbf{Both Mass and} \\
              \textbf{Adaptation Only} & \textbf{Adaptation Only} & \textbf{Damping Adaptation} \\ \hline
        \multirow{2}{2.36cm}{\cite{podobnik2007haptic}, \cite{dimeas2016}}
        & \cite{podobnik2007haptic}, \cite{dimeas2016}, \cite{duchaine2008investigation}, \cite{ryu2008frequency},  & \cite{dimeas2016}, \cite{gallagher2014improved}, \cite{wang2023variable}, \cite{okunev2012human},      \\
        & $\!\!\!\!\!\!\!\!\!\!\!\!\!\!$\cite{grafakos2016variable}, \cite{duchaine2012stable}, \cite{erden2011assisting} & $\!\!\!\!\!\!\!\!\!\!\!\!\!\!$\cite{lecours2012variable}, \cite{ferraguti2019variable},  \cite{topini2022variable}\\
        \hline
    \end{tabular}
    \vspace{-2mm}
\end{table}


\section{Methodology and Contributions}\label{sec:meth_contr}

From the literature analysis, we have identified several aspects that deserve further investigation:\\
i) Nonidealities such as joint stiffness and system delays are not always included in system stability analysis, while other nonlinear effects, such as actuator saturation, are never considered. However, these effects are always present in robotic systems due to the physical limitations of the actuators.\\
ii) There is no unified consensus in the literature on the best strategy for adapting admittance parameters to ensure stability in admittance-based pHRI applications.\\
iii) All the works found in the literature focus on unconstrained user movements. To the best of our knowledge, no research paper addresses the stability of pHRI applications enhanced by virtual fixtures.

This paper advances the state of the art in stability analysis and control design for admittance-controlled pHRI systems through the following contributions:\\
i) The stability analysis of admittance-controlled pHRI systems enhanced by virtual fixtures is conducted.\\
ii) Nonidealities such as joint stiffness, system delays, and, most importantly, the nonlinearity associated with actuator saturation are introduced. In particular, the latter plays a crucial role in explaining the undesired oscillatory behavior leading to instability, as observed in experimental tests.\\
iii) A sensitivity analysis is performed to evaluate the influence of different admittance parameter variations on system stability, supported by a remarkable agreement between theoretical, simulation, and experimental results.\\
The tool that enabled this type of analysis was the Nyquist plot, which, together with the describing function method, allows for the efficient handling of static nonlinear functions (such as motor saturation) and system delays.\\
In addition,
iv) a novel strategy for admittance parameters adaptation is proposed, aiming to maximize transparency for the user while moving along the desired path.

\begin{figure}[t]
\hspace{-6.8mm}
\begin{minipage}{0.48\columnwidth}
    \centering
    \includegraphics[width=0.68002\linewidth]{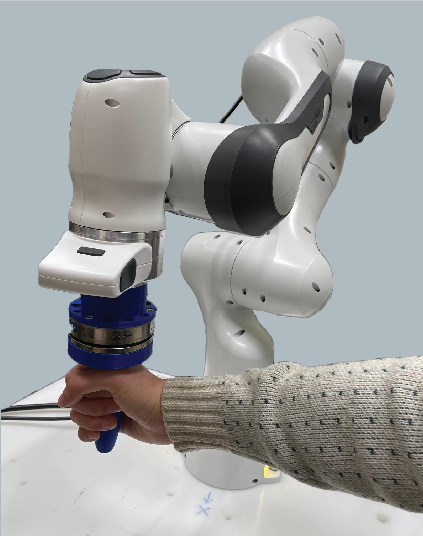}

\end{minipage}
\hspace{-6.8mm}
\begin{minipage}{0.52\columnwidth}
    \centering
    \includegraphics[clip,width=1.1402\linewidth]{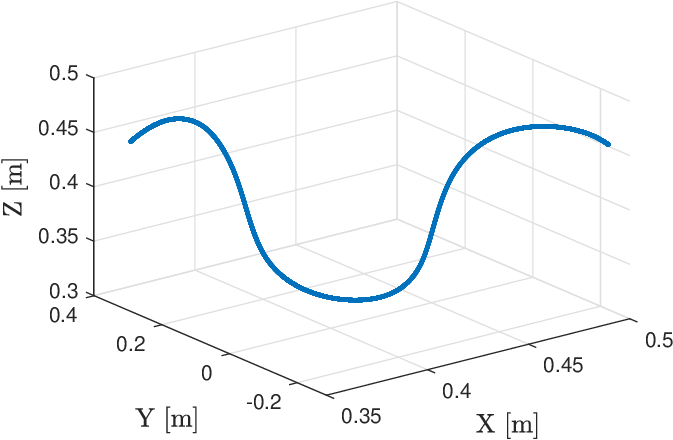}
\end{minipage}
\psset{unit=\unitlength} \SpecialCoor
       \rput(-178.68,-56.88){\footnotesize (a)}
   \rput(-59.98,-56.88){\footnotesize (b)}
\vspace{3.6mm}
\caption{Experimental robotic setup  (a) and reference path used in the experimental tests (b).}\label{fig:Path}
\vspace{-2mm}
 \end{figure}

\begin{figure}[t]
    \centering
    \includegraphics[width=0.9\linewidth]{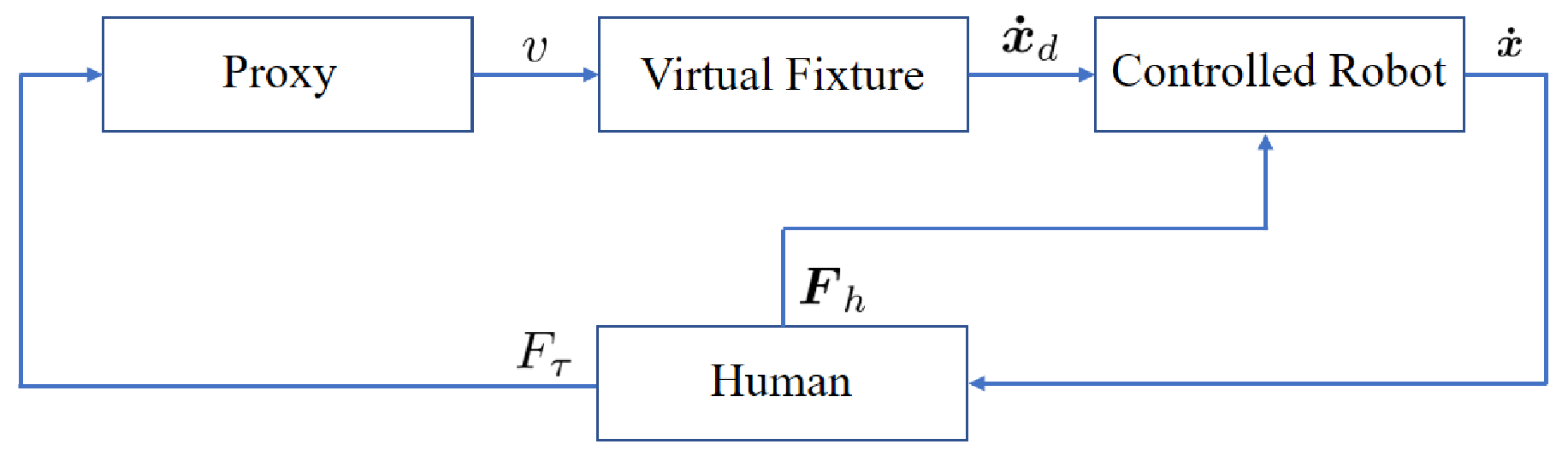}
    \vspace{-2mm}
    \caption{Schematic representation of the proxy-based pHRI framework enhanced by virtual fixture of Fig.~\ref{fig:FrankaTrj2}.} 
\label{fig:HR_scenario}
\vspace{-3mm}
\end{figure}
%


\section{Robotic System Description and Modeling  }\label{Experimental_Setup_sect}
The robotic setup is based on the 7-DoF Franka Emika Panda collaborative robot equipped with an Axia80-M20 force-torque sensor on its terminal flange, as shown in Fig.~\ref{fig:Path}(a). The user's limb is connected to the end-effector through a simple handle; however, different types of constraints can be designed depending on the application.

The user's motion is restricted along a generic path $\varphi(l)$, where $l$ is the arc length, 
without enforcing a specific traversal rate along the path,
using a virtual mass called the \textit{proxy} (see Fig.~\ref{fig:FrankaTrj2}) moving along the curve under the influence of the tangential component $F_\tau$ of the user force $\F_h$:
\begin{equation}\label{Proxy_eq}
 m\,\dot{v} + b\,{v} =F_{\tau}, \hspace{4.6mm} \mbox{where} \hspace{4.6mm} F_{\tau} = \frac{d\varphi(l)}{dl}^T \F_h
 \end{equation}
and $v = \dot l$ is the velocity along the curve.\\
As illustrated in Fig. \ref{fig:HR_scenario}, the position $\varphi(l(t))$along the guiding virtual fixture is used as the reference position for the robot, under a Cartesian inverse dynamics control law: 
\[
\boldsymbol{\tau} = \M(\q) \y + \C(\q, \dot{\q})\dot{\q} +\boldsymbol{g}(\q)
\]
with $
\y=\J^{-1}(\q)\left( \ddot \x_d -\boldsymbol{\dot{J}}(\q)\dot \q +\K_D'\dot{\tilde{\x}} +\F_{el}'(\tilde{\x})\right)$ being the auxiliary input, where $\tilde{\x} = \x-\x_d$ represents the Cartesian error with respect to the desired pose $\x_d(t) = \varphi(l(t))$.
The elastic force $\F_{el}(\tilde{x})$ is generally defined by considering different expressions along the longitudinal and orthogonal directions to the curve. This ensures the precise tracking of the virtual mass position along the path, while allowing the user the flexibility to deviate more or less from the prescribed position depending on the application. For more details on possible implementations, refer to \cite{9562059,braglia2025phaseindependentdynamicmovementprimitives,onfiani2024optimizingdesigncontrolmethods,shahriari2024path}. Note that the focus of this work concerns the stability of this type of control scheme when perturbed conditions are taken into account, rather than the stability of the control scheme in nominal conditions.
To this end, linear analysis based on Nyquist plot has been adopted, by considering a linearized model the controlled robot and the virtual fixture.
Therefore, the scheme of Fig \ref{fig:HR_scenario_bis}(a) has been deduced, where: 
\begin{figure}[t]
    \centering
    \begin{pspicture}(0,1.5)(20,9.56)
   \rput(4.25,7.425){
    \includegraphics[width=0.8525\linewidth]{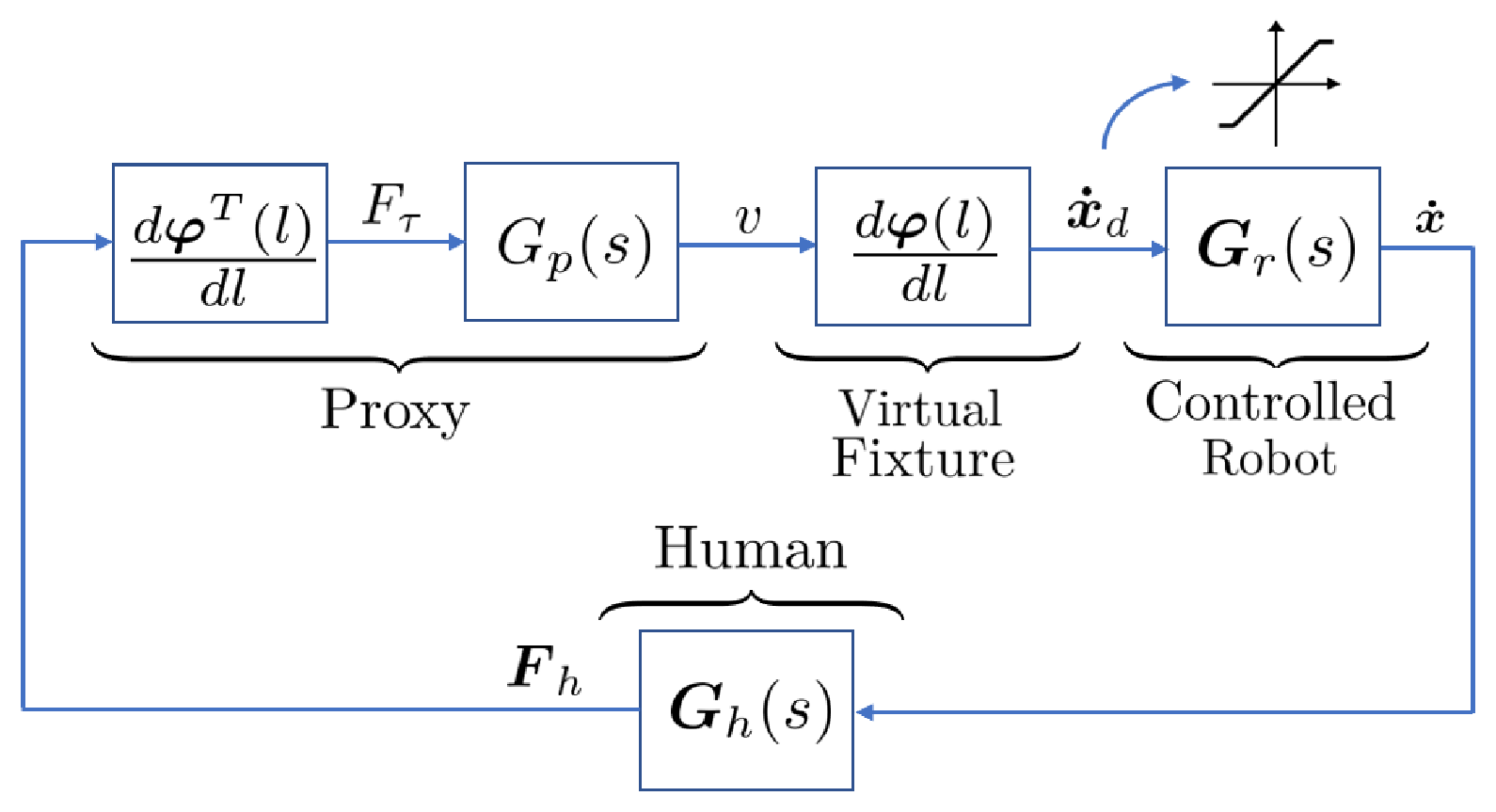}}
    \rput(4.25,3.9){\includegraphics[width=0.84\linewidth]{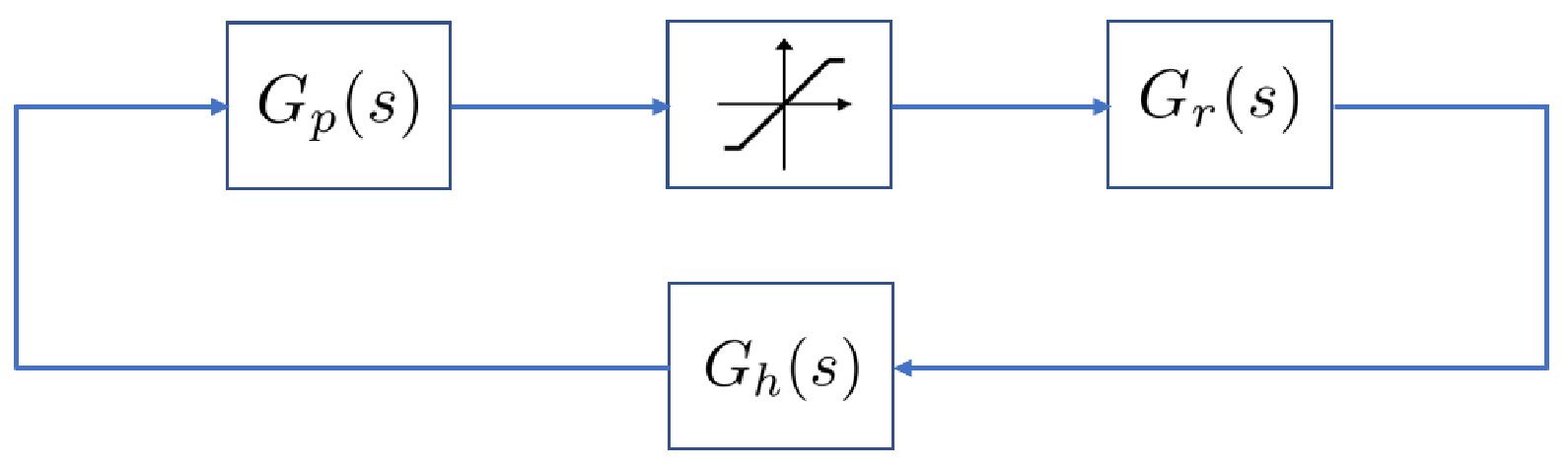}}
     \centering \scriptsize \setlength{\unitlength}{3.55mm}
     \end{pspicture}
\psset{unit=\unitlength} \SpecialCoor
       \rput(-0.82,118.5){\footnotesize (a)}
   \rput(-0.56,44.88){\footnotesize (b)}
    \vspace{-16mm}
    \caption{(a) Linearized block scheme of the proxy-based pHRI framework enhanced by virtual fixture of Fig.~\ref{fig:HR_scenario}. (b) Simplified scalar representation used for stability analysis.} 
\label{fig:HR_scenario_bis}
\vspace{-2mm}
\end{figure}

\noindent \mbox{
\pscircle[fillstyle=solid,fillcolor=black,linecolor=black](0.05,0.1){0.05}\hspace{0.68mm}} $G_p(s)$ is the transfer function of the proxy dynamics \eqref{Proxy_eq}:
\vspace{-2mm}
\begin{equation}\label{F_h_eq_bis}
G_p(s)=\frac{V(s)}{F_\tau(s)} =\dfrac{1}{m s+b }, 
\end{equation}
where $V(s) = \mathcal{L}\{v(t)\}$, $F_\tau(s) = \mathcal{L}\{F_\tau(t)\}$.

\noindent \mbox{
\pscircle[fillstyle=solid,fillcolor=black,linecolor=black](0.05,0.1){0.05}\hspace{0.68mm}} The human is modeled as a (multi-dimensional) spring-damper system:
\begin{equation}\label{human_tf}
 \G_h=   \frac{\F_h(s)}{\dot{\X}(s)} = \frac{\b_{h}s\!+\!\K_{h}}{s}, 
\end{equation}
where $\F_h(s) = \mathcal{L}\{\F_h(t)\}$, $\dot{\X}(s) = \mathcal{L}\{\dot x(t)\}$, $\b_h$ and $\K_h$ are the damping coefficient and stiffness of the user, respectively.

\noindent \mbox{
\pscircle[fillstyle=solid,fillcolor=black,linecolor=black](0.05,0.1){0.05}\hspace{0.68mm}} The transfer matrix $\G_r(s) = \frac{X(s)}{X_d(s)} = \frac{\dot X(s)}{\dot X_d(s)}$, whose expression will be detailed later, models the behavior of the (position) controlled robot linearized around the operating point $(\x,\dot{\x}) = (\varphi(l^\star),\,\0)$, which depends on the position $l^\star$ of the proxy along the desired path.

\noindent \mbox{
\pscircle[fillstyle=solid,fillcolor=black,linecolor=black](0.05,0.1){0.05}\hspace{0.68mm}} The virtual fixture shape is modeled through the relation between the proxy velocity along the curve and the velocity required to the robot:
\[
\dot{\x}_d(t) = \frac{d\varphi(l)}{d l} v(t).
\]
 \begin{figure}[tbp]
\centering
\includegraphics[clip,width=0.859\columnwidth]{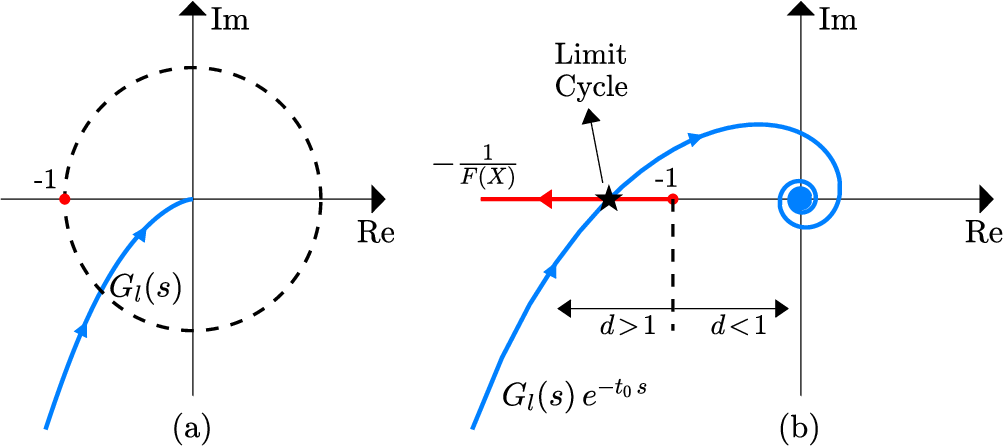}
\caption{Qualitative Nyquist plot of the loop gain function $G_l(s)$ with (a) and without (b) delay $t_0$ affecting the system.}\label{Nyquist_Diagrams_NEW}
 \end{figure}
Note that, when considering a specific operating point $l^\star$, the terms $\frac{d\varphi(l)}{d l}$ and $\frac{d\varphi^T(l)}{d l}$ reduce to constant (vectorial) gains. The gain $\frac{d\varphi(l)}{d l}$ maps the scalar proxy dynamics along the curve to the 3D space where the robot and human work, with a lifting operation on the velocity, while the gain $\frac{d\varphi^T(l)}{d l}$ projects the dynamics in the 3D space onto the curve where the proxy is constrained, based on the exchanged force. With the objective of studying the influence of the proxy parameters on the loop stability, the overall model is reduced to the scalar case, as shown in Fig. \ref{fig:HR_scenario_bis}(b), where $G_r(s)$ and $G_h(s)$ denote the dynamics of the robot and the human projected onto the path constraint and where the saturation function pointed out in Fig. \ref{fig:HR_scenario_bis}(a) is inserted into the loop, highlighting the velocity limits of the actuation system. The transfer function $G_h(s)$ is a scalar version of $\G_h(s)$ in \eqref{human_tf}, while $G_r$ is obtained by considering a one-DOF robot with an elastic joint, endowed with an inverse dynamics position controller. According to  \cite{ROCCO19972041}, the transfer function between the input torque $\tau_m$ and position $x_m$ of the link actuator is:
\begin{equation}\label{load_joint_actuator_model}
G_m(s)\!=\!\dfrac{X_m(s)}{T_m(s)}\!=\!\dfrac{J_{lr}\,s^2+D_{el}\,s+K_{el}}{
s[a_1\,s^3+a_2\,s^2+a_3\,s+d]}, 
\end{equation}
where $a_1=J_{lr}\,J_m$,  $a_2=J_{lr} D_m+J_t D_{el}$, $a_3=J_t K_{el}+D_m D_{el}$ and $a_4=D_m K_{el}$. $J_t=J_{lr}+J_{m}$ is the total inertia, $J_{lr}$ is the link inertia reduced by $n^2$, and $J_m$ is the motor inertia.
Parameters $D_m$ and $D_{el}$  are the damping coefficients of the motor and the elastic transmission, respectively, and $K_{el}$ is the transmission stiffness.
Assuming the position sensor to be located on the actuator, as typically done in industrial robots, the inverse dynamics control \cite{SicilianoBook} 
$\tau_m=J_t(\ddot{x}_d+K_D\tilde{\dot{x}}+K_P\tilde{x})$, where $\tilde{x}=x_d-x_m$, can be applied to model \eqref{load_joint_actuator_model}, leading to the following transfer function $G_{c}(s)$ between the actuator speed $x_m$ and the reference speed $x_d$ (the same relation applies to positions):
\begin{equation}\label{load_joint_actuator_model_1}
G_{c}(s)\!=\!\dfrac{\dot{X}_m(s)}{\dot{X}_d(s)}\!=\!\dfrac{b_1's^4+b_2's^3+b_3's^2+b_4's+b_5'}{a_1's^4+a_2's^3+a_3's^2+a_4's+a_5'},
\end{equation}
where the coefficients of $G_{c}(s)$ are defined in the appendix. 
The actuator and robot velocities, $\dot{x}_m$ and $\dot{x}$, are related by:
\begin{equation}\label{load_joint_actuator_model_1_bis}
G_{mr}(s)\!=\!\dfrac{\dot{X}(s)}{\dot{X}_m(s)}\!=\!\dfrac{D_{el}s+K_{el}}{n(J_{lr}s^2+D_{el}s+K_{el})}.
\end{equation}
From \eqref{load_joint_actuator_model_1} and \eqref{load_joint_actuator_model_1_bis}, the transfer function $G_r(s)$ of the controlled robot can be derived:
$G_{r}(s)\!=\!\frac{\dot{X}(s)}{\dot{X}_d(s)}\!=\!G_c(s)\,G_{mr}(s)$.
\begin{figure}[t]
\centering
\includegraphics[clip,width=0.859\columnwidth]{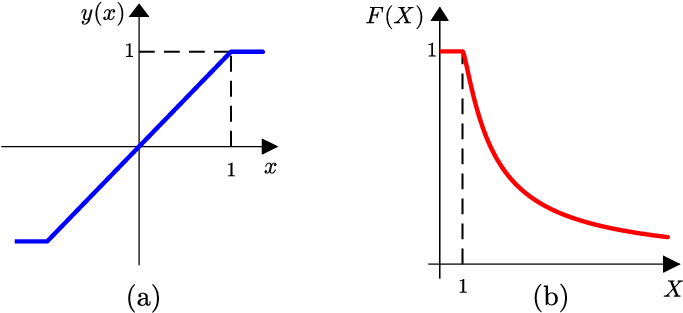}
\caption{(a) Saturation and (b) its describing function $F(X)$.}\label{funz_descrittiva_sat}
\vspace{-2mm}
 \end{figure}
 \section{Stability Analysis of the pHRI Framework}\label{sec:Section_One}
Using \eqref{F_h_eq_bis}, \eqref{human_tf}, \eqref{load_joint_actuator_model_1} and \eqref{load_joint_actuator_model_1_bis},
the loop gain function $G_l(s)=G_h(s)G_p(s)G_r(s)$ of the scheme of Fig.~\ref{fig:HR_scenario_bis}(b) can be found:
\[G_{l}(s)\!=\!\dfrac{\beta_1s^6\!\!+\!\beta_2s^5\!\!+\!\beta_3s^4\!\!+\!\beta_4s^3\!\!+\!\beta_5s^2\!\!+\!\beta_6s\!\!+\!\beta_7}{s(\alpha_1s^7\!\!+\!\!\alpha_2s^6\!\!+\!\alpha_3s^5\!\!+\!\alpha_4s^4\!\!+\!\alpha_5s^3\!\!+\!\alpha_6s^2\!\!+\!\alpha_7s\!\!+\!\alpha_8)}\!, 
\]
where the coefficients of $G_{l}(s)$ are defined in the appendix.\\
The qualitative Nyquist diagram of function $G_l(s)$ is reported in Fig.~\ref{Nyquist_Diagrams_NEW}(a), showing that the system is always asymptotically stable by using the Nyquist criterion, as no intersection with the negative real semiaxis occurs. However, a more realistic representation of the loop gain function incudes the following two nonidealities: 1) the system delays,
such as the discretization delay and the delay due to the finite bandwidth of the position control; 2) the velocity saturation of the actuators shown in the loop of 
Fig. \ref{fig:HR_scenario_bis}(b).
Letting $t_0$ represent the overeall delay affecting the system, the resulting loop gain function $G_{l0}(s)$ is given by:
\begin{equation}\label{lg_fcn_1}
G_{l0}(s)=G_{l}(s)\,e^{-t_0\,s}.
\end{equation}
Because of the delay $t_0$, the qualitative Nyquist plot of the loop gain function $G_l(s)$ shown in Fig.~\ref{Nyquist_Diagrams_NEW}(a) deforms as shown in Fig.~\ref{Nyquist_Diagrams_NEW}(b). The characteristic of the saturation nonlinearity $y(x)$ is shown in Fig.~\ref{funz_descrittiva_sat}(a). When the saturation is subject to a sinusoidal excitation $x=X \sin{(\omega t)}$, its response is given by its describing function $F(X)=\frac{Y_1(X)}{X}\,e^{j \varphi_1(X)}$~\cite{franklin2002feedback},
where $Y_1(X)$ and $\varphi_1(X)$ are the amplitude and the phase shift of the approximated sinusoidal signal $y(t) \simeq Y_1(X) \sin{(\omega t + \varphi_1(X))}$ after the saturation using a first-order truncated Taylor series expansion. The describing function $F(X)$ of the saturation is shown in shown in Fig.~\ref{funz_descrittiva_sat}(b). 
Using the Nyquist stability theory~\cite{1701151} and the describing function method, the sinusoidal signal $x=X \sin{(\omega t)}$ is persistent in the system if the so-called self-sustaining equation
$F(X)\,G_{l0}(s)=-1$ admits solutions. Given that the describing function $F(X)$ of the saturation only assumes positive real values, as shown in Fig.~\ref{funz_descrittiva_sat}(b), the self-sustaining equation is typically solved graphically on the Nyquist plane, by searching for an intersection between function $-1/F(X)$ and the Nyquist plot of the loop gain function $G_{l0}(s)$, as shown in Fig.~\ref{Nyquist_Diagrams_NEW}(b). If an intersection exists, the persistent sinusoidal excitation is called limit cycle, and determines the typical undesired oscillatory phenomena in the pHRI framework.
Fig.~\ref{Nyquist_Diagrams_NEW}(b) shows that the combined effect of the delay $t_0$ and a sufficiently large modulus of the loop gain function $G_{l0}(s)$ likely causes the existence of a limit cycle in the pHRI framework.
The intersection of the Nyquist plot of $G_{l0}(s)$ with the negative real semi-axis occurs at point $1/M_a$, indicated by the star in Fig.~\ref{Nyquist_Diagrams_NEW}(b), where $M_a$ is the gain margin of the system expressed in a linear scale. Therefore, the following distance $d$ can be defined, quantifying the distance from the critical point $-1$ in Fig.~\ref{Nyquist_Diagrams_NEW}(b): 
\begin{equation}\label{distance_definition}
\begin{array}{r@{\hspace{4.46mm}}l}
d=\dfrac{1}{M_a},  & 
\left\{\begin{array}{@{}r@{\,}c@{\,}l@{\hspace{2.68mm} \rightarrow \hspace{2.68mm}}c}
d & > & 1 & \mbox{limit cycle},  \\[1mm]
d & < & 1 & \mbox{no limit cycle}.  
\end{array}
\right. 
\end{array}
\end{equation}
 \begin{figure}[tbp]
\centering
\includegraphics[clip,width=0.9\columnwidth]{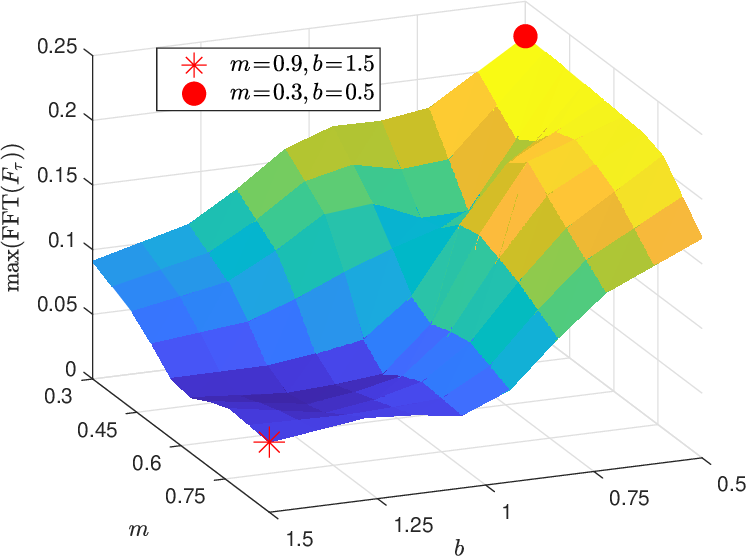}
      \setlength{\unitlength}{5.0mm}
 \psset{unit=\unitlength}
  %
\caption{Spectral analysis of the tangential human force $F_\tau$: peak value $P_k$ of the high frequency components of FFT($F_\tau$).}\label{Plotter_Davide_10_14_01_2024_non_live_FFT_1}
\vspace{-1.4mm}
 \end{figure}
  \begin{figure}[tbp]
\centering
\includegraphics[clip,width=1\columnwidth]{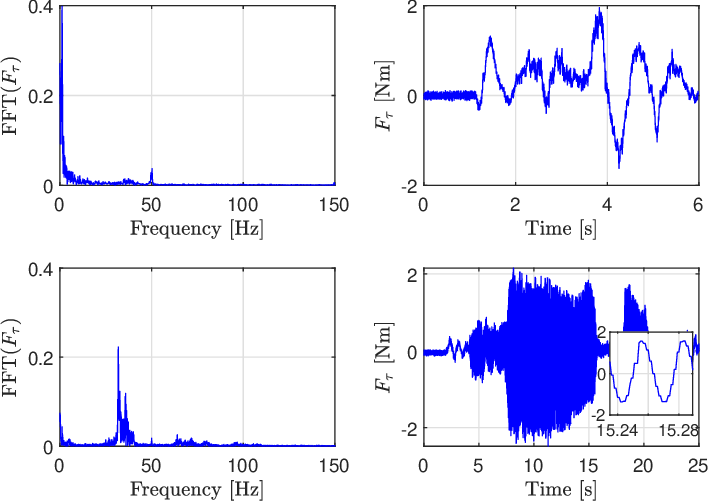}
      \setlength{\unitlength}{5.0mm}
 \psset{unit=\unitlength}
  \rput(-1.5,-0.5){
\rput(-2.5,1.056){\footnotesize (c)}
\rput(6.5,7.46){\footnotesize (b)}
\rput(-2.5,7.46){\footnotesize (a)}
\rput(6.5,1.029){\footnotesize (d)}
\rput(8.4,3.779){\footnotesize (e)}
}
\vspace{-2.8mm}
\caption{Amplitude spectrum (a, c) and time-plot (b, d) of the tangential human force $F_\tau$ when: (a, b) $m\!=\!0.9$ and $b\!=\!1.5$ (no oscillation occurring) and (c, d) $m\!=\!0.3$ and $b\!=\!0.5$ (oscillation occurring).
(e): Zoom-in of subplot (d).}\label{Plotter_Davide_10_14_01_2024_non_live_FFT_2}
\vspace{-3.6mm}
 \end{figure}
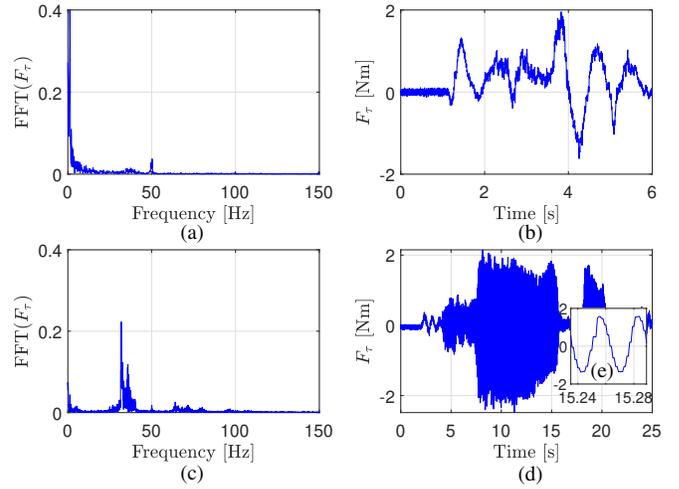
From \eqref{distance_definition}, it follows that a persistent oscillation occurs if $d>1$, 
namely if there is an intersection between the inverse $-1/F(X)$ of the describing function $F(X)$ and the Nyquist plot of $G_{l0}(s)$. Conversely, no oscillation occurs if $d<1$. 

The existence of a persistent oscillation when $m$ and $b$ are sufficiently low has been verified experimentally 
through the proposed experimental setup described in Sec.~\ref{Experimental_Setup_sect}. The trajectory of Fig.~\ref{fig:Path}(b) has been executed for $m \in \cN_m=[0.3,\;\; 0.9]$ and $b \in \cN_b=[0.5,\;\; 1.5]$.
For each execution, the tangential component $F_\tau$ of the human force has been acquired using the force/torque sensor described in Sec.~\ref{Experimental_Setup_sect}. The tangential component $F_\tau$ is analyzed because of its influence in the proxy dynamics \eqref{Proxy_eq}, directly affecting the evolution of the robot position.  
In particular, given the constraint condition enforced through the virtual fixture to assist the user during the task execution, the motion in directions orthogonal to the reference is restricted. Consequently, the corresponding force components do not contribute to motion generation.

Considering that the human arm bandwidth in voluntary motions is typically below $2$ Hz~\cite{dimeas2016}, a Fast-Fourier Transform (FFT) has been applied to all acquisitions of $F_\tau$. The peak value of the high frequency (i.e. $>2$ Hz) spectral components of $F_\tau$ is reported in Fig.~\ref{Plotter_Davide_10_14_01_2024_non_live_FFT_1}, showing that the oscillations intensity decreases as $m$ and, mainly, $b$ increase.

Furthermore, Fig.~\ref{Plotter_Davide_10_14_01_2024_non_live_FFT_2}(a) 
and Fig.~\ref{Plotter_Davide_10_14_01_2024_non_live_FFT_2}(c) show the amplitude spectra of the tangential force $F_{\tau}$ in the two cases highlighted in red in Fig.~\ref{Plotter_Davide_10_14_01_2024_non_live_FFT_1}, while Fig.~\ref{Plotter_Davide_10_14_01_2024_non_live_FFT_2}(b) 
and Fig.~\ref{Plotter_Davide_10_14_01_2024_non_live_FFT_2}(d) show the corresponding time behavior of the tangential force $F_{\tau}$. From the zoom-in of Fig.~\ref{Plotter_Davide_10_14_01_2024_non_live_FFT_2}(e), the presence of a limit cycle $F_\tau(t)=X_{cl}\sin{(2\pi f_{cl}t)}$ can be clearly seen, at a frequency  $f_{cl}\simeq 30$ Hz. The proposed analysis using the describing function method allows to provide a clear justification on why the proxy-based pHRI framework enhanced by virtual fixture is affected by oscillatory behaviors deteriorating the stability of the pHRI interaction.

\subsection{Sensitivity Analysis to Admittance Parameters Variation}\label{sec:Section_Sensitivity}
\begin{figure}[t]
\centering
\includegraphics[clip,width=0.8\columnwidth]{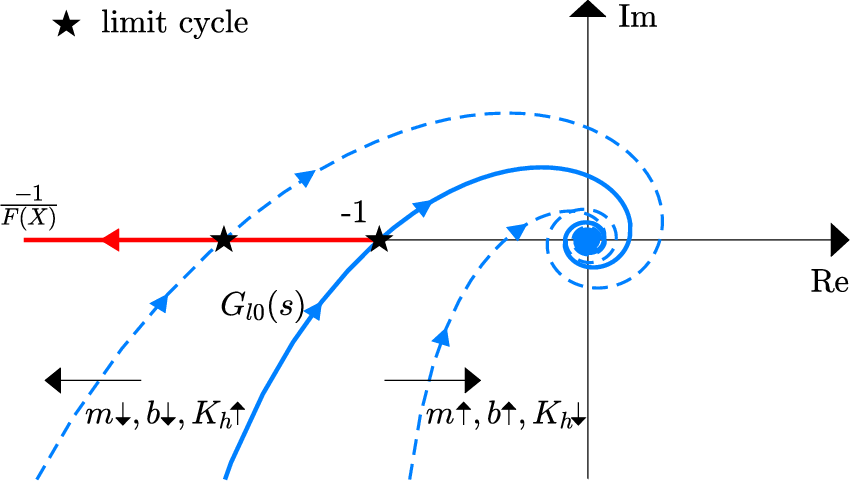}
\vspace{-1.22mm}
\caption{Loop gain function: qualitative Nyquist plot as a function of the admittance parameters and human stiffness.}\label{prova_new_new_bis_sensibility}
\vspace{-1mm}
 \end{figure}

Consider the Nyquist diagram of the loop gain function $G_{l0}(s)$ in \eqref{lg_fcn_1} shown  in Fig.~\ref{prova_new_new_bis_sensibility} for a given set of parameters. This represents the boundary condition for the presence of a limit cycle in the system due to the intersection between $G_{l0}(s)$ and $-1/F(X)$. 
It is widely known~\cite{dimeas2016} that larger values  of the human stiffness $K_h$ and/or smaller values of the admittance parameters $m$, $b$ cause instability of the pHRI framework, and viceversa.
This is consistent with the structure of the loop gain function $G_{l0}(s)$, suggesting that larger values of  $K_h$  increase the modulus of $G_{l0}(s)$, while larger values of $m$ and $b$ reduce it, see Fig.~\ref{prova_new_new_bis_sensibility}.  While we have no direct control on the stiffness $K_h$, the admittance parameters $m$ and $b$ can be adapted to prevent undesired oscillatory phenomena in the pHRI. This is indeed the widely adopted approach in the literature, as indicated in Table~\ref{tab:adaptation_classification}.

However, the literature review in Sec.~\ref{Admittance_Parameters_Adaptation_sect} has shown that there is no unified consensus on the best admittance parameters adaptation approach to ensure stability in the considered pHRI system.
Therefore, we propose a sensitivity analysis in order to determine which of the three most commonly used approaches in the literature - namely, i) adapting the virtual mass $m$ only, ii) adapting the virtual friction $b$ only, iii) adapting both $m$ and $b$ while keeping a constant ratio - has the strongest impact on system stability. 

\begin{table}[t]
  \caption{Sensitivity analysis: robotic system parameters.} \label{Gearbox_metrics_table}
  \vspace{-2mm}
  \begin{center}
\begin{tabular}{|@{\;}c@{\;}|}
\hline 
$J_l\!=\!0.66$ km $\mbox{m}^2\!$, \;$K_{el}\!=\!100$ Nm/rad, \; $D_{el}\!=\!0.01$ Nm s/rad, \; $n\!=\!50$, \\ \hline 
$\!J_m\!\!=\!0.10$ km $\mbox{m}^2\!$, \!\!\! $D_{m}\!\!=\!0.11$ Nm s/rad, \!\!\! $K_{h}\!\!=\!150$ N/m, \!\!\! $b_{h}\!\!=\!0.68$ N s/m \\ \hline 
  \end{tabular}
    \end{center}
    \vspace{-2.6mm}
\end{table}

\begin{Remar}\label{remark_sensitivity}
The sensitivity analysis of the distance $d$ in \eqref{distance_definition}
with respect to the admittance parameters indicates the effectiveness of these parameters in preventing the proxy-based pHRI framework enhanced by virtual fixture from exhibiting a limit cycle causing undesired oscillation.
\end{Remar}

\begin{figure}[t]
\begin{minipage}{0.28\columnwidth}
\psfrag{db su b0}[][][0.9][0]{$\Delta b/b_0$}
\psfrag{dM su M0}[b][b][0.9][0]{$\Delta {d}/{d_{0}}$}
 \centering \scriptsize \setlength{\unitlength}{3.55mm}
\psset{unit=\unitlength} \SpecialCoor
\begin{pspicture}(0,4)(20,9.5)
   \rput(3.95,7.05){\includegraphics[clip,width=1.14\columnwidth]{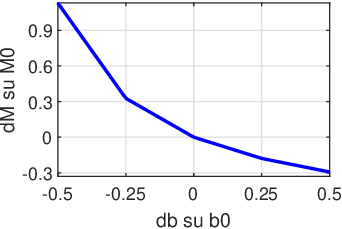}}
   \rput(4.4,3.56){(a)}
\end{pspicture}
\end{minipage}
\hspace{2.6mm}
\begin{minipage}{0.28\columnwidth}
\psfrag{dm su m0}[][][0.9][0]{$\Delta m/m_0$}
\psfrag{dM su M0}[b][b][0.9][0]{$\Delta_{d}/{d_{0}}$}
 \centering 
 \scriptsize \setlength{\unitlength}{3.55 mm}
\psset{unit=\unitlength} \SpecialCoor
\begin{pspicture}(0,4)(20,9.5)
   \rput(4,7){
   \includegraphics[clip,width=1.14\columnwidth]{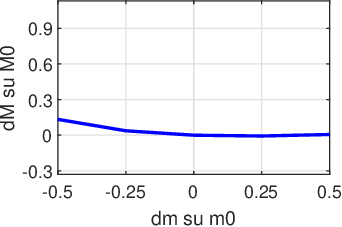}}
   \rput(4.56,3.56){(b)}

\end{pspicture}
\end{minipage}
\hspace{2.8mm}
\begin{minipage}{0.28\columnwidth}
\psfrag{dKv su Kv0}[][][0.9][0]{$\Delta r/r_0$}
\psfrag{dM su M0}[b][b][0.9][0]{$\Delta {d}/{d_{0}}$}
 \centering \scriptsize \setlength{\unitlength}{3.55 mm}
\psset{unit=\unitlength} \SpecialCoor
\begin{pspicture}(0,4)(20,9.5)
   \rput(4,7){\includegraphics[clip,width=1.14\columnwidth]{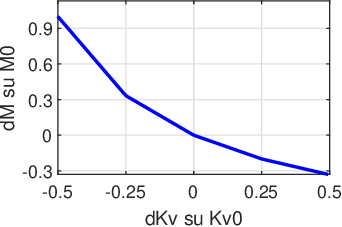}}
   \rput(4.56,3.56){(c)}
\end{pspicture}
\end{minipage}
\vspace{2mm}
\caption{Sensitivity functions $S^{d}_b$ (a), $S^{d}_m$ (b) and $S^{d}_r$ (c).}\label{Nyquist_Plots_08_01_2024_m_e_b_variabile_con_const_ratio}
 \end{figure}

The sensitivity functions $S^{d}_b$, $S^{d}_m$, $S^{d}_r$ of the distance $d$ with respect to variations of the admittance parameters $m$, $b$, and of the distance $r=\sqrt{m^2+b^2}$ from the origin of the plane $(m,b)=(0,0)$ while keeping a constant ratio $m/b$, can be defined as follows~\cite{6429294}: 
\begin{equation}\label{sensitiv_fcn_th}
S^{d}_b=\dfrac{\Delta d/d_{0}}{\Delta b/b_0}, \hspace{4mm}
S^{d}_m=\dfrac{\Delta d/d_{0}}{\Delta m/m_0}, \hspace{4mm}
S^{d}_r=\dfrac{\Delta d/d_{0}}{\Delta r/r_0},
\end{equation}
where $\Delta d$ is the variation of the distance $d$ from the nominal value $d_0$ when the parameters $b$, $m$ and $r$ are subject to variations $\Delta b$, $\Delta m$ and $\Delta r$ with respect to the nominal values $b_0$, $m_0$ and $r_0$. The three sensitivity functions are shown in Fig. \ref{Nyquist_Plots_08_01_2024_m_e_b_variabile_con_const_ratio}.
As expected, when the parameters have negative variations, $\Delta d$ is positive, meaning that the intersection with the negative real axis of the Nyquist plot tends to move to the left, where instability arises (see Fig. \ref{prova_new_new_bis_sensibility}). On the contrary, increasing the parameters values leads to a reduction of $\Delta d$, resulting in a stable behavior. Interestingly, the maximum variation of $\Delta d$ occurs when $b$ is modified.

From an experimental point of view, the existance of a limit cycle can be detected through the peak value $P_k$ of the high frequency (i.e. $>2$ Hz) spectral components of FFT($F_\tau$), as discussed in Sec.~\ref{sec:Section_One}. 
The sensitivity functions $S^{P_k}_b$, $S^{P_k}_m$, $S^{P_k}_r$ of the peak value $P_k$ with respect to variations of the admittance parameters $m$, $b$, and with respect to the aforementioned distance $r$,
can be defined as follows~\cite{6429294}: 
\begin{equation}\label{sensitiv_fcn_exper}
\!\!\!\!S^{P_k}_b\!=\!\dfrac{\Delta P_k/P_{k_0}}{\Delta b/b_0}\!, \hspace{0.52mm}
S^{P_k}_m\!=\!\dfrac{\Delta P_k/P_{k_0}}{\Delta m/m_0}\!, \hspace{0.52mm}
S^{P_k}_r\!=\!\dfrac{\Delta P_k/P_{k_0}}{\Delta r/r_0}\!,
\end{equation}
where $\!\Delta P_k$ is the variation of $P_k$ from its nominal value $\!P_{k_0}$.$\!$

Assuming typical values \cite{1570781,new1} for the parameters of the robotic system and human model as in Table~\ref{Gearbox_metrics_table}, the nominal values $m_0=0.6$ kg and $b_0=1 \mbox{N s/m}$ 
for the admittance parameters, a delay $t_0=0.005$ s, and the control parameters $K_D=89.44$ and $K_P=2000$, the sensitivity functions $S^{d}_b$, $S^{d}_m$, $S^{d}_r$ in \eqref{sensitiv_fcn_th} can be computed for $b \in b_0 \pm 50 \% (b_0)$, $m \in m_0 \pm 50 \% (m_0)$ and $r \in r_0 \pm 50 \% (r_0)$. Using the same ranges for the admittance parameters, the sensitivity functions $S^{P_k}_b$, $S^{P_k}_m$, $S^{P_k}_r$ in \eqref{sensitiv_fcn_exper} can also be computed from the exerimental acquisitions of Fig.~\ref{Plotter_Davide_10_14_01_2024_non_live_FFT_1}.  
\begin{figure}[t]
\begin{minipage}{0.28\columnwidth}
\psfrag{db su b0}[][][0.9][0]{$\Delta b/b_0$}
\psfrag{dPk su Pk0}[b][b][0.9][0]{$\Delta {P_{k}}/{P_{k_0}}$}
 \centering \scriptsize \setlength{\unitlength}{3.55mm}
\psset{unit=\unitlength} \SpecialCoor
\begin{pspicture}(0,4)(20,9.5)
   \rput(3.95,7.05){\includegraphics[clip,width=1.14\columnwidth]{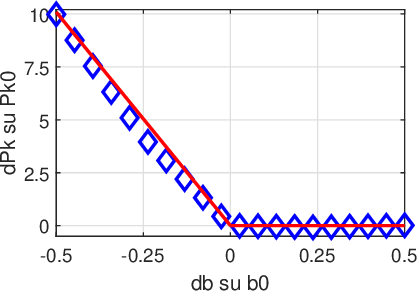}}
    \rput(4.4,3.56){(a)}
\end{pspicture}
\end{minipage}
\hspace{2.6mm}
\begin{minipage}{0.28\columnwidth}
\psfrag{dm su m0}[][][0.9][0]{$\Delta m/m_0$}
\psfrag{dPk su Pk0}[b][b][0.9][0]{$\Delta {P_{k}}/{P_{k_0}}$}
 \centering 
 \scriptsize \setlength{\unitlength}{3.55 mm}
\psset{unit=\unitlength} \SpecialCoor
\begin{pspicture}(0,4)(20,9.5)
   \rput(4,7){
   \includegraphics[clip,width=1.14\columnwidth]{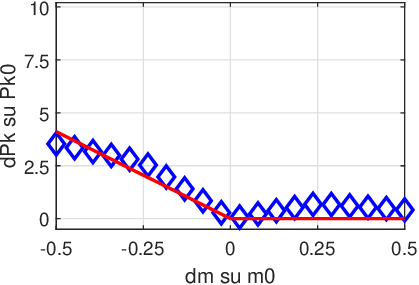}}
   \rput(4.4,3.56){(b)}
\end{pspicture}
\end{minipage}
\hspace{2.8mm}
\begin{minipage}{0.28\columnwidth}
\psfrag{dKv su Kv0}[][][0.9][0]{$\Delta r/r_0$}
\psfrag{dPk su Pk0}[b][b][0.9][0]{$\Delta {P_{k}}/{P_{k_0}}$}
 \centering \scriptsize \setlength{\unitlength}{3.55 mm}
\psset{unit=\unitlength} \SpecialCoor
\begin{pspicture}(0,4)(20,9.5)
   \rput(4,7){\includegraphics[clip,width=1.14\columnwidth]{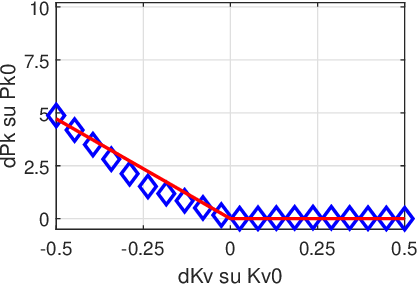}}
    \rput(4.4,3.56){(c)}
\end{pspicture}
\end{minipage}
\vspace{2mm}
\caption{Sensitivity functions $S^{P_k}_b$ (a), $S^{P_k}_m$ (b) and $S^{P_k}_r$ (c).}\label{Plotter_Davide_10_08_01_2024_non_live_m_e_b_variabile}
    \vspace{-3.6mm}
 \end{figure}

 Fig.~\ref{Plotter_Davide_10_08_01_2024_non_live_m_e_b_variabile}(a) shows that $S^{P_k}_b$ from experimental data is the highest sensitivity, in agreement with the theoretical sensitivity $S^{d}_b$ being the highest one in Fig.~\ref{Nyquist_Plots_08_01_2024_m_e_b_variabile_con_const_ratio}(a), meaning that the admittance parameter $b$ is the one mainly influencing the pHRI stability. 
Fig.~\ref{Plotter_Davide_10_08_01_2024_non_live_m_e_b_variabile}(b) shows that $S^{P_k}_m$ from experimental data is the lowest sensitivity, in agreement with the theoretical sensitivity $S^{d}_m$ being the lowest one in Fig.~\ref{Nyquist_Plots_08_01_2024_m_e_b_variabile_con_const_ratio}(b), meaning that the admittance parameter $m$ is the one least influencing the pHRI stability.
Fig.~\ref{Plotter_Davide_10_08_01_2024_non_live_m_e_b_variabile}(c) and Fig.~\ref{Nyquist_Plots_08_01_2024_m_e_b_variabile_con_const_ratio}(c) show that the theoretical analysis and experimental results agree in showing that increasing both admittance parameters $m$ and $b$ with a constant ratio provides intermediate performances between cases (a) and (b)
in reducing the pHRI oscillations.

\section{Adaptation of Admittance Parameters to Ensure Transparency and Stability}\label{sec:Section_STransparency}
\noindent
Based on the considerations of the previous section, an adaptation strategy for the admittance parameters is proposed, in order to
maximize transparency in pHRI while ensuring stability under all conditions. In human-robot collaborative tasks, the user effort is strongly influenced by the admittance parameters \cite{kim2021variable}, that shape the perceived interaction.
\begin{figure}[t]
    \centering
    \includegraphics[width=0.636\linewidth]{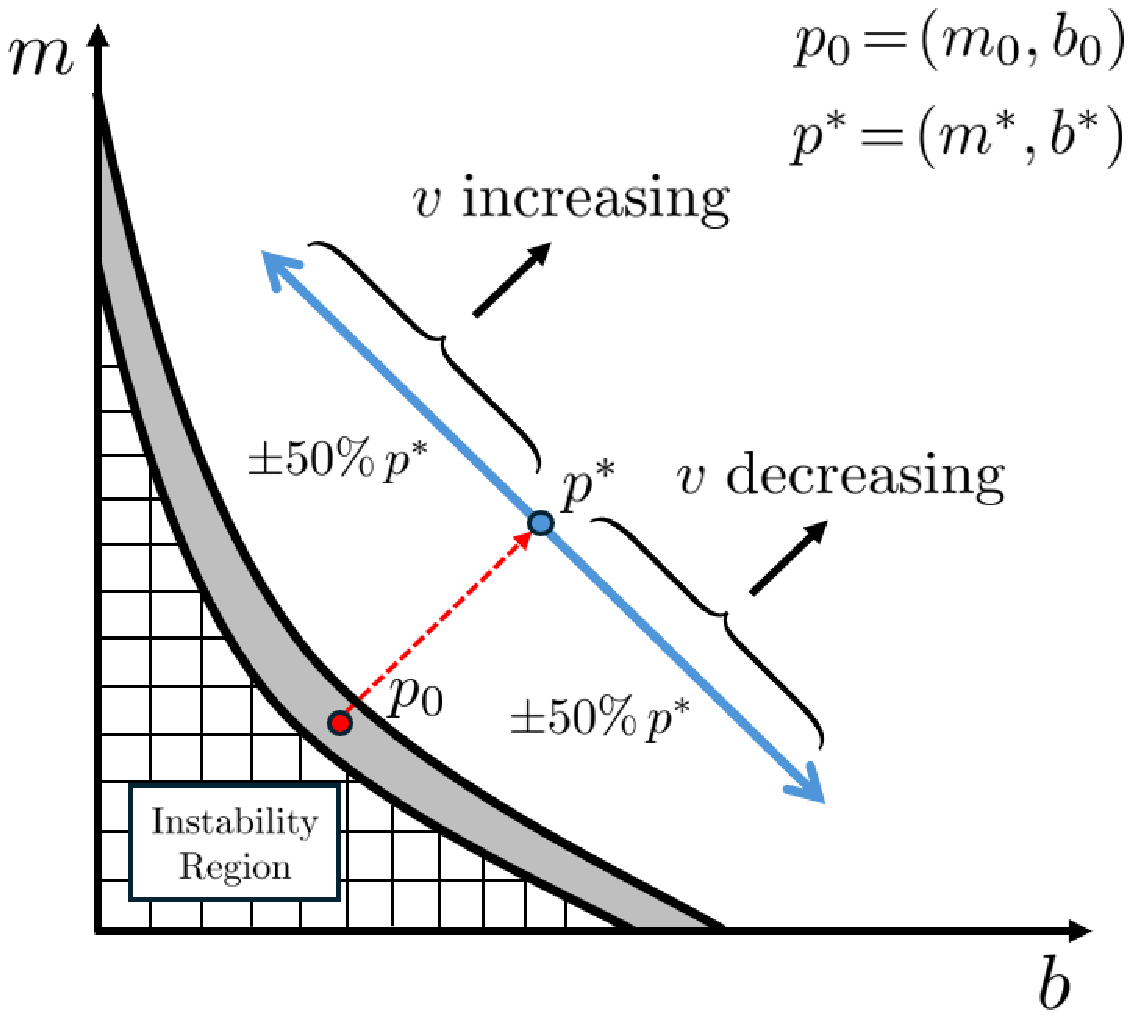}
    \vspace{-2mm}
    \caption{Adaptation strategy for the admittance parameters, onrder to maximize transparency within the stability region.}
    \label{fig:map}
    \vspace{-2.6mm}
\end{figure}

Transparency in pHRI refers to the system's ability to minimize the user's perception of resistive forces during movement, caused, e.g., by real or virtual inertia, friction, etc.
In physical control applications, such as robotic rehabilitation or teleoperation, a transparent robot allows the user to perform smooth and natural movements without perceiving artificial constraints (other than the virtual fixture). This is achieved through impedance control strategies, compensation for the robot's dynamics (such as gravity and Coriolis forces), and the reduction of apparent inertia and friction. 
Transparency can be quantitatively evaluated by measuring the average force exerted by the user while performing a task. A lower applied force indicates greater system transparency, i.e. the robot imposes minimal resistance to the user’s movement.
In general, maintaining low values for the parameters $m$ and $b$ reduces the required user effort. However, as shown in the previous section, this condition is highly detrimental to the system stability.

Based on these considerations, the following simulative experiment has been performed. Letting $v_d$ be the velocity profile imposed by the human, the effort required by the human to perform the trajectory can be quantified as $\bar{F}_\tau=\mbox{avg}(|F_{\tau}|)$, where $F_{\tau}$ can be computed from \eqref{Proxy_eq}.
The nominal value $\bar{F}^0_{\tau}$ of the metric $\bar{F}_\tau$ can be computed by replacing the admittance parameters $m$ and $b$ in \eqref{Proxy_eq} with the nominal values $m_0=0.6$ kg and $b_0=1$ N s/m highlighted in Fig.~\ref{fig:map}, and represents the maximum acceptable user effort to execute the considered task. The admittance parameters are varied, as a function of the proxy velocity $v$, using the following logic, as depicted in Fig.~\ref{fig:map}: 
\vspace{-1mm}
\begin{equation}\label{proxy_param_algorithm}
\begin{array}{r@{\,}c@{\,}l}
m(v) & = & m_{min}\!+\!(m_{max}-m_{min}) \dfrac{\mbox{sat}(|v|)}{v_{0}}, \\[3.68mm]
b(v) & = & b_{max}\!+\!(b_{min}-b_{max}) \dfrac{\mbox{sat}(|v|)}{v_{0}},
\end{array}
\end{equation}
where $v_0$ is the nominal velocity for the task (=max($v_d$)) and  sat($\cdot$)  is a saturation function that limits the output to the range $[0,v_0]$. The idea behind the adoption of this mechanism, which linearly changes the physical parameters from  $(m_{min},b_{max})$ to $(m_{max},b_{min})$ as $v$ increases,  is based on some basic considerations: i) instability occurs when both $m$ and $b$ are characterized by small values; ii) from the dynamical equation of the proxy in \eqref{Proxy_eq}, it clearly emerges that the inertial term is dominant for small velocity values while, when the velocity $v$ approaches $v_0$, it is no longer necessary to accelerate, and the user force is spent against the linear friction. Therefore, small values of $m$ are desirable for low velocity, while small values of $b$ help the user at high velocities; iii) with this approach, a lower level of human force is required, 
while stability is guaranteed without feedback action that requires oscillations detection (which is probably the most challenging problem in stability recovery).\\
To guarantee strong stability properties, the central values $p_0 = (m_0,b_0)$ of Fig.~\ref{fig:map} have been modified with respect to the initial setting, with the goal of maintaining the same user effort, through the following optimization problem:
\begin{Optimi}\label{opt:optim_pstar}
Let $\bar{F}_{\tau_v}$ be the user effort when the admittance parameters are varied as in Fig.~\ref{fig:map} around a generic point $p$. 
The objective is to find the optimal value $p^*$ of point $p=(m,b)$, around which the admittance parameters can be varied using \eqref{proxy_param_algorithm}, such that 
\vspace{-1mm}
\[
\displaystyle \min_{p} J(p), \;\;\;\;\;\; \mbox{where} \;\;\;\;\;
J(p)\!=\!|\bar{F}_{{\tau }_v}-\bar{F}^0_{{\tau}}|
\]
 \end{Optimi}
and $m_{min}=0.5 \,m_*$, $m_{max}=1.5 \,m_*$, $b_{min}=0.5 \,b_*$, $b_{max}=1.5 \,b_*$. The results of the optimization problem, solved by iteration implementing \eqref{Proxy_eq} in the Simulink environment, are reported in Table~\ref{transparency_maximization_resul},
\renewcommand{\arraystretch}{1.26}
\begin{table}[t]
  \caption{Results of the admittance parameters optimization for maximizing transparency: $p^*=1.5 p^0$.} \label{transparency_maximization_resul}
  \vspace{-2.8mm}
  \begin{center}
\begin{tabular}{|cccc|}
\hline 
& $p^0\!=\!(m_0,b_0)$ & $p^*\!=\!1.5\,
p^0$\, & $p^*\!=\!1.5\,
p^0$ \mbox{varying as} \eqref{proxy_param_algorithm}   \\ \hline 
$\bar{F}_{\tau}$ & $\bar{F}^0_{{\tau}}=0.35$ & $\bar{F}^*_{{\tau}}=0.53$ & $\bar{F}^*_{{\tau}_v}=0.34$ 
\\ \hline 
  \end{tabular}
    \end{center}
    \vspace{-4.8mm}
\end{table}
\begin{table}[t]
  \caption{Results of the application of the transparency optimal results $p^*=1.5 p_0$ 
  to the experimental setup.} \label{transparency_maximization_resul_exper}
  \vspace{-2.8mm}
  \begin{center}
\begin{tabular}{|cccc|}
\hline
& $p^0\!=\!(m_0,b_0)$ & $p^*\!=\!1.5\,
p^0$\, & $p^*\!=\!1.5\,
p^0$ \mbox{varying as} \eqref{proxy_param_algorithm}   \\ \hline
$\bar{F}_{\tau}$ & $\bar{F}^0_{{\tau}}=0.28$ & $\bar{F}^*_{{\tau}}=0.38$ & $\bar{F}^*_{{\tau}_v}=0.28$ 
\\ \hline 
  \end{tabular}
    \end{center}
        \vspace{-2.8mm}
\end{table}
showing that $\bar{F}^0_{{\tau}} \simeq \bar{F}^*_{{\tau}_v}$. This means that the effort required by the user when parameters $m$ and $b$ are varied as in \eqref{proxy_param_algorithm} starting from point $p^*=1.5\,p^0$ - see Fig.~\ref{fig:map} - is approximately the same as the effort required by the user when parameters $m$ and $b$ are constant in point $p_0$ at the edge of the instability region.
Notably, if the admittance parameters $m$ and $b$ are kept constant at point $p^*$, the user effort $\bar{F}^*_{{\tau}}$ is significantly higher instead (by $35.85\%$), as shown in Table~\ref{transparency_maximization_resul}. 
Furthermore, the aforementioned benefit in terms of transparency comes together with stability, as the blue path in Fig.~\ref{fig:map} lies outside of the instability region.

These results have been verified experimentally using the setup described in Sec.~\ref{Experimental_Setup_sect}. 
The results are shown in Table~\ref{transparency_maximization_resul_exper}. In this case, if the admittance parameters are kept constant at point $p^*$, the user effort $\bar{F}^*_{{\tau}}$ results to be higher by $26.32\%$, showing that the experimental results well agree with the results of the simulative optimization shown in Table~\ref{transparency_maximization_resul}.

 \section{Conclusions} 
\label{sec:Conclusions}
This paper has addressed the stability, sensitivity and transparency analyses in pHRI using variable admittance control. While all the analyzed related works refer to free co-manipulation, the present work has considered pHRI enhanced by virtual fixtures for the first time. The performed stability analysis includes nonlinear effects that enabled to gain a deep unserstanding of the nature of undesired oscillations causing the unstable behavior. Given the missing consensus in terms of best performances between the different admittance parameters adaptation techniques proposed in the analyzed related works, we have then performed a sensitivity analysis to discriminate the admittance parameter having the greatest influence in pHRI. The obtained simulative results have been confirmed by experimental validation. Lastly, an optimization problem equipped with a novel admittance parameters adaptation technique have been proposed in order to maximize transparency in the pHRI. The very good results of the optimization problem in terms of transparency have also been verified by the experimental results.

\appendix

\section*{A: Coefficients of Functions $G_c(s)$ and $G_l(s)$}
The coefficients of the transfer function $G_c(s)$ are:
\vspace{-1.46mm}
\[
\footnotesize
\begin{array}{l}
b_1'\!=\!J_{t}J_{lr},\, b_2'\!=\!J_{t} (D_{el}\!+\!K_D J_{lr}),\,  b_3'\!=\!J_t(K_{el}\!+\!K_D D_{el}\!+\!K_P J_{lr}), \\[1.4mm] b_4'\!\!=\!J_t(K_D K_{el}\!\!+\!\!K_P D_{el}), 
b_5'\!\!=\!J_t K_P K_{el},  a_1'\!\!=\!J_{lr}J_{m}, a_2'\!\!=\!J_{lr}(D_{m}\!+\\[1.4mm]
+\!J_t K_D)\!+\!D_{el}J_t, \;\; a_3'\!=\!J_t(J_{lr}K_P\!+\!D_{el}K_D)\!+\!J_t K_{el}\!\!+\!\!D_m D_{el}, \\[1.4mm]
a_4'=K_{el}(D_m+J_t K_D)+D_{el}J_t K_P, \;\;\;\;a_5'=K_{el} J_t K_P.
\end{array}
\]
The coefficients of the loop gain function $G_l(s)$ in are: 
\vspace{-1.46mm}
\[
\footnotesize
\begin{array}{l}
\beta_1\!=\!D_{el} b_1' b_h, \beta_2\!=\!D_{el}(K_h b_1'\!+\!b_2' b_h)\!+\!K_{el} b_1 b_h,
\beta_3\!=\!D_{el}(K_h b_2'\!+\\[1.4mm]
+b_h b_3')\!+\!K_{el}(K_h b_1'\!+\!b_2' b_h),
\beta_4\!=\!D_{el}(K_h b_3'\!+\!b_h b_4')\!+\!K_{el}(K_h b_2'\!+\!\\[1.4mm]
\!+\!b_h b_3'),\beta_5\!=\!D_{el}(K_h b_4'\!+\!b_h b_5')\!+\!K_{el}(K_h b_3'\!+\!b_h b_4'),%
\beta_6\!=\!D_{el} K_h b_5'\!+\\[1.4mm]
+\!K_{el}(K_h b_4'\!+\!b_h b_5'),\;\;\beta_7\!=\!K_{el} K_h b_5',
 \alpha_1\!=\!\!J_{lr} a'_1 m n,\alpha_2\!=\!\!J_{lr}n(a_1' b\!+\\[1.4mm]
+\!a_2' m  )\!+\!D_{el} a_1' m n,  \alpha_3\!=\!J_{lr}n(a_2' b+a_3' m  )\!+\!D_{el}n (a_1' b+a_2' m)\!+,  \\[1.4mm]
+\!K_{el} a_1' m n,\alpha_4\!=\!J_{lr}n(a_3' b\!+\!a_4' m  )\!+\!D_{el}n (a_2' b\!+\!a_3' m)\!+\!K_{el}n (a_1'b \!+  \\[1.4mm]
+\! a_2' m), \alpha_5\!=\!J_{lr}n(a_4' b\!+\!a_5' m  )\!\!+\!\!D_{el}n (a_3' b\!+\!a_4' m)\!\!+\!\!K_{el}n (a_2'b \!+\! a_3' m),  \\[1.4mm]
\alpha_6\!=\!J_{lr}n a_5' b+D_{el}n (a_4' b+a_5' m)\!+\!K_{el}n (a_3'b + a_4' m),  \\[1.4mm]
\alpha_7\!=\!D_{el}n a_5' b\!+\!K_{el}n (a_4'b + a_5' m),\;\;\;\;  \alpha_8\!=\!K_{el} a'_5 b n.
\end{array}
\]

\bibliographystyle{IEEEtran}
\bibliography{bibliography}

\end{document}